\pdfoutput=1
\documentclass{article}

\usepackage{fix-cm}
\usepackage[T1]{fontenc}
\usepackage[utf8]{inputenc}
\usepackage{microtype}
\usepackage{graphicx}
\usepackage{booktabs}
\PassOptionsToPackage{dvipsnames,table}{xcolor}
\PassOptionsToPackage{hyphens}{url}
\usepackage{hyperref}

\usepackage[preprint]{icml2027}
\makeatletter
\renewcommand{\Notice@String}{\textit{Preprint.}}
\makeatother
\usepackage{amsmath,amssymb,mathtools,amsthm}
\usepackage{multirow}
\usepackage{tabularx}
\usepackage{array}
\usepackage{xspace}
\usepackage{enumitem}

\usepackage{tikz}
\usetikzlibrary{arrows.meta,positioning,fit,backgrounds,shapes.geometric}
\usepackage{tcolorbox}
\tcbuselibrary{skins,breakable}
\usepackage{amsmath,amsfonts,bm}

\def\eqref#1{equation~\ref{#1}}
\def\1{\bm{1}}

\DeclareMathAlphabet{\mathsfit}{\encodingdefault}{\sfdefault}{m}{sl}
\SetMathAlphabet{\mathsfit}{bold}{\encodingdefault}{\sfdefault}{bx}{n}

\definecolor{evoblue}{HTML}{2C5AA0}
\definecolor{evogreen}{HTML}{3A7D44}
\definecolor{evoorange}{HTML}{C46A1A}
\definecolor{evogray}{HTML}{F1F3F5}
\definecolor{qfInk}{HTML}{162B49}
\definecolor{qfMuted}{HTML}{4A607A}
\definecolor{qfRule}{HTML}{D8E2EC}
\definecolor{qfTeal}{HTML}{1E7B64}
\definecolor{qfTealBg}{HTML}{EAF6F1}
\definecolor{qfTealBd}{HTML}{93C8B8}
\definecolor{qfBronze}{HTML}{A66814}
\definecolor{qfBronzeBg}{HTML}{FDF5E6}
\definecolor{qfBronzeBd}{HTML}{DFBC83}
\definecolor{qfCoral}{HTML}{B84730}
\definecolor{qfCoralBg}{HTML}{FDEEEB}
\definecolor{qfCoralBd}{HTML}{E3A396}
\definecolor{qfCrimson}{HTML}{991B2E}
\definecolor{qfCrimsonBg}{HTML}{FCEBEC}
\definecolor{qfCrimsonBd}{HTML}{DF9AA4}
\definecolor{qfBlue}{HTML}{256396}
\definecolor{qfBlueBg}{HTML}{EEF5FB}
\definecolor{qfBlueBd}{HTML}{A3C3DC}
\definecolor{qfGray}{HTML}{52667A}
\definecolor{qfGrayBg}{HTML}{F1F5F9}
\definecolor{qfGrayBd}{HTML}{CBD5E1}

\newcommand{\vbadge}[4]{%
  \tikz[baseline=(X.base)]\node[inner xsep=3.2pt,inner ysep=1.3pt,rounded corners=2.3pt,fill=#1,draw=#2,line width=0.45pt,font=\sffamily\bfseries\fontsize{6.3}{7.2}\selectfont,text=#3](X){#4};}
\newcommand{\vSup}{\vbadge{qfTealBg}{qfTealBd}{qfTeal}{Supported}}
\newcommand{\vMed}{\vbadge{qfBronzeBg}{qfBronzeBd}{qfBronze}{Mediator-only}}
\newcommand{\vTar}{\vbadge{qfCoralBg}{qfCoralBd}{qfCoral}{Target-only}}
\newcommand{\vUns}{\vbadge{qfCrimsonBg}{qfCrimsonBd}{qfCrimson}{Unsafe}}
\newcommand{\vUnr}{\vbadge{qfBlueBg}{qfBlueBd}{qfBlue}{Unresolved}}
\newcommand{\vNef}{\vbadge{qfGrayBg}{qfGrayBd}{qfGray}{No effect}}
\newcommand{\tAcc}{\vbadge{qfTealBg}{qfTealBd}{qfTeal}{Transfer: accepted}}
\newcommand{\tRej}{\vbadge{qfCoralBg}{qfCoralBd}{qfCoral}{Transfer: rejected}}
\newcommand{\tNone}{\vbadge{qfGrayBg}{qfGrayBd}{qfGray}{Transfer: --}}
\newcommand{\focaltag}{\vbadge{qfBlueBg}{qfBlueBd}{qfBlue}{focal}}
\newcommand{\fbadge}[1]{\vbadge{qfGrayBg}{qfGrayBd}{qfInk}{#1}}

\newcommand{\stagebanner}[4]{%
  \par\smallskip\noindent
  \begin{tcolorbox}[enhanced,frame hidden,boxrule=0pt,left=7pt,right=7pt,top=3pt,bottom=3pt,
    colback=#1,borderline west={3pt}{0pt}{#2},arc=3pt,before skip=5pt,after skip=3pt]
    {\sffamily\bfseries\small\color{#2}#3}\par\vspace{1pt}{\scriptsize\itshape\color{qfMuted}#4}
  \end{tcolorbox}\nopagebreak
}

\newenvironment{roundcard}[7]{%
  \begin{tcolorbox}[
    enhanced,
    colback=white,colframe=#2,boxrule=0.6pt,
    borderline west={2.6pt}{0pt}{#3},
    arc=3pt,
    colbacktitle=#1,coltitle=qfInk,
    titlerule=0.5pt,titlerule style={#2},
    toptitle=2.2pt,bottomtitle=2.2pt,lefttitle=6pt,righttitle=6pt,
    top=2.8pt,bottom=2.8pt,left=6pt,right=6pt,
    before skip=3.5pt,after skip=3.5pt,
    title={\begin{tabular}{@{} >{\raggedright\arraybackslash}p{0.50\linewidth} @{\hspace{4pt}} >{\raggedleft\arraybackslash}p{0.48\linewidth} @{}}
      {\footnotesize\textbf{\sffamily\textcolor{#3}{#4}}\enspace{\scriptsize\textcolor{qfMuted}{\mbox{#5}}}} &
      {#6\,#7}
    \end{tabular}}
  ]
  \renewcommand{\arraystretch}{1.04}%
  \scriptsize
  \begin{tabular}{@{} p{0.125\linewidth} @{\hspace{6pt}} p{0.845\linewidth} @{}}
}{%
  \end{tabular}\arrayrulecolor{black}
  \end{tcolorbox}
}
\newcommand{\rcrow}[2]{{\sffamily\bfseries\scriptsize\textcolor{qfMuted}{#1}} & #2 \\[1pt]}
\newcommand{\rcrowsep}[2]{\arrayrulecolor{qfRule}\hline\noalign{\vskip 1.5pt}{\sffamily\bfseries\scriptsize\textcolor{qfMuted}{#1}} & #2 \\[1pt]}
\newcommand{\QuantForge}{\textup{QuantForge}\xspace}
\newcommand{\HiRes}{\textup{HiRes}\xspace}
\newcommand{\revcolor}{}
\hypersetup{
  pdftitle={QuantForge: Discovering Residual Decompositions for MXFP4 Post-Training Quantization},
  pdfauthor={Qiulin Shang, Zhoutong Wu, Jie Hu, Kun Yuan},
  pdfsubject={Preprint},
  colorlinks=true,linkcolor=evoblue,citecolor=evoblue,urlcolor=evoblue
}

\icmltitlerunning{QuantForge: Discovering Residual Decompositions for MXFP4 Post-Training Quantization}

\begin{document}

\twocolumn[
  \icmltitle{QuantForge: Discovering Residual Decompositions for MXFP4 Post-Training Quantization}

  \begin{icmlauthorlist}
    \icmlauthor{Qiulin Shang}{pku}
    \icmlauthor{Zhoutong Wu}{pku}
    \icmlauthor{Jie Hu}{pku}
    \icmlauthor{Kun Yuan}{pku}
  \end{icmlauthorlist}
  \icmlaffiliation{pku}{Peking University, Beijing, China}
  \icmlcorrespondingauthor{Qiulin Shang}{qiulin.shang@stu.pku.edu.cn}

  \icmlkeywords{Post-Training Quantization, MXFP4, Large Language Models, Automated Algorithm Discovery}

  \vskip 0.3in
]

\printAffiliationsAndNotice{}

\begin{abstract}
Designing post-training quantization (PTQ) algorithms for strict 4-bit MXFP4 (W4A4) is difficult because errors interact across weights, activations, and consecutive stages: improving one step often shifts what later steps must correct. LLM program search can explore such multi-stage quantizers as code, yet scoring candidates only by perplexity hides \emph{why} an edit helped or hurt. Unlike black-box search, however, PTQ always has the original BF16 model running layer-for-layer alongside the quantized model on the same calibration data. We introduce \QuantForge, which uses this paired BF16--MXFP4 twin to guide discovery: it runs local counterfactual controls to separate competing explanations, verifies with runtime checks that the next code edit acts on each finding, and remeasures the remaining error after each stage. Seeded from GPTQ, \QuantForge discovers \HiRes, uniting geometric shaping, installed-weight-aware activation coding, and sequential attention-to-SwiGLU recovery. Across seven models (3B--32B) and nine baselines, \HiRes achieves the lowest C4 perplexity and mean degradation on every model ($-41\%$ on Qwen3-32B vs.\ WUSH); on matched-budget runs, \QuantForge hits the held-out target twice as often as reflection memory ($6/8$ vs.\ $3/8$ seeds) and wins $15/16$ paired runs across three proposers while cutting search compute $35.7\%$. Code is available at \url{https://github.com/qiulin13145/QuantForge}.
\end{abstract}

\section{Introduction}
\label{sec:introduction}

\begin{figure*}[t]
  \centering
  \includegraphics[width=0.93\textwidth]{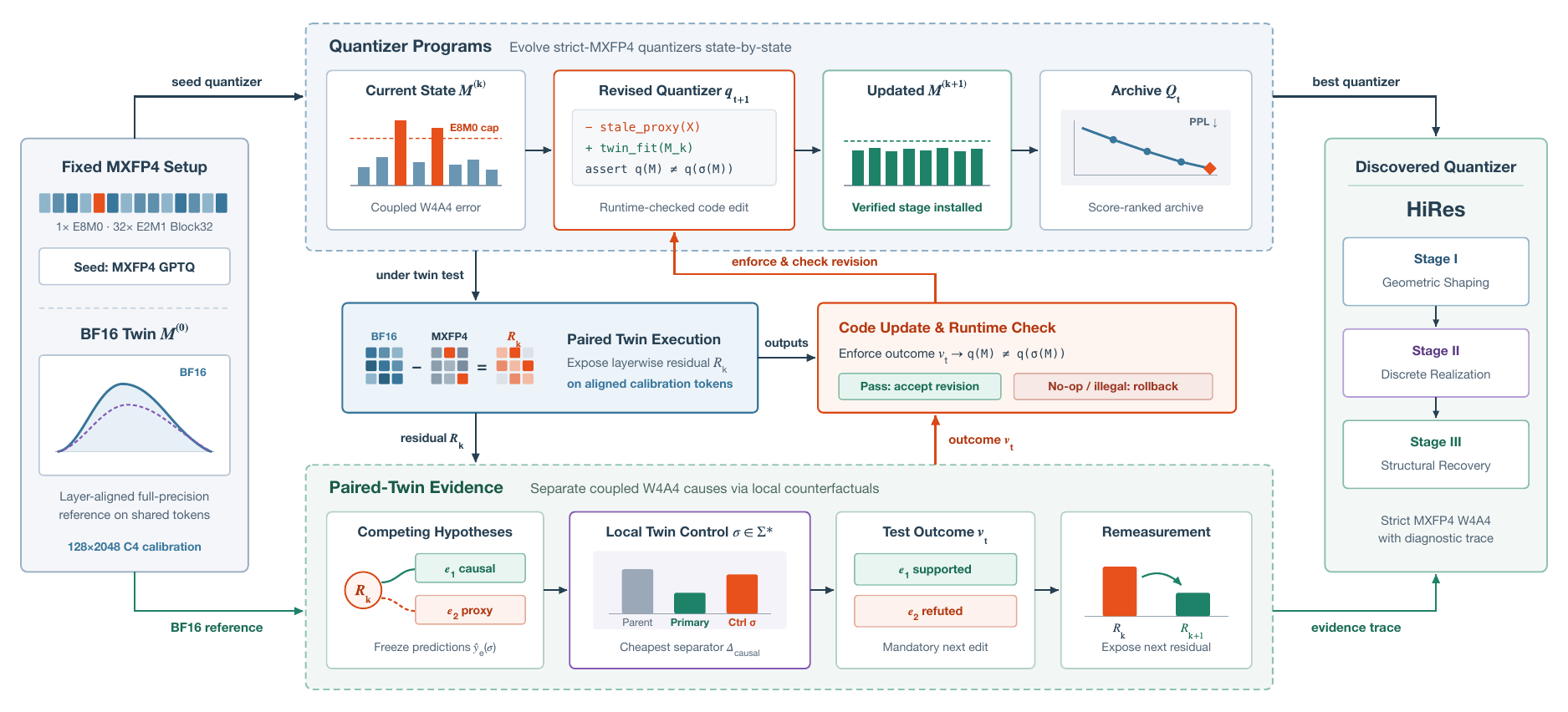}
  \caption{\textbf{Overview of \QuantForge.} Under a fixed strict-MXFP4 W4A4 setup (E2M1/E8M0, Block32), \QuantForge co-evolves quantizer programs (\textbf{top}) and paired-twin causal evidence (\textbf{bottom}) from strict-MXFP4 GPTQ. Each state $M^{(k)}$ runs alongside its layer-aligned BF16 twin to expose residual $R_k$; competing hypotheses ($e_1,e_2$) are separated by a minimum-cost local control $\sigma$, enforcing the test outcome as a runtime-checked edit while strong quantizers stay archived in $\mathcal{Q}_t$. Installing verified stages and remeasuring fresh residuals $R_{k+1}$ yields \HiRes.}
  \label{fig:errata-framework}
\end{figure*}

Large language models power modern text generation, question answering, and reasoning \citep{grattafiori2024llama,yang2025qwen3}, making memory and compute reduction essential at scale. Post-training quantization (PTQ) reduces these costs without retraining \citep{frantar2023gptq}, and four-bit weights and activations (W4A4) cut nominal precision fourfold relative to 16-bit formats. MXFP4 provides a standardized representation: 32 values share a power-of-two scale, letting short blocks adapt to local magnitudes with little metadata \citep{rouhani2023microscaling,ocp2023mxspec}. This compact format also couples rounding precision within each block: a scale set by large values can leave smaller entries poorly resolved.

A natural response is to transform coordinates to spread large values across a block \citep{xiao2023smoothquant,ashkboos2024quarot,chen2026wush}. In MXFP4, transformed values must still fit a sparse four-bit grid with a shared power-of-two scale \citep{cook2025mrgptq,shao2025brq}, so the coordinate choice affects scale selection and code assignment and can diverge from smooth second-moment proxies. It also alters the input correlations used by second-order reconstruction to compensate for weight-rounding error \citep{frantar2023gptq}. Once quantized operators are installed, their errors propagate through attention and interact through gated MLP products \citep{shazeer2020glu}, shifting the trajectory on which downstream recovery must be fitted \citep{li2025gptaq,yan2026d2quant}. Recent hand-crafted methods address individual couplings---weight--activation balancing \citep{shao2024omniquant,sun2025flatquant,chen2026wush}, block rotation with GPTQ \citep{cook2025mrgptq,shao2025brq}, or upstream error compensation \citep{li2025gptaq}---yet composing them into a full W4A4 pipeline remains difficult: improving one step changes what the next step must correct.

LLM-driven program evolution can explore these dependencies by rewriting quantization operations as executable code \citep{romeraparedes2024funsearch,novikov2025alphaevolve,liu2024eoh}. Performance is an effective selection signal, but an incomplete design signal: a transform may help by spreading outliers or by improving shared-exponent selection, pointing to a stronger transform in one case and revised block encoding in the other. Even when the editable interface expands \citep{li2026optscientist} or search uses textual reflection \citep{ye2024reevo,gepa2026,liu2026oragent}, aggregate scores and free-form diagnostics rarely separate rival error mechanisms or ensure that the next program implements the conclusion.

We study PTQ algorithm discovery when improving one component changes the errors later components must address. Unlike black-box search, PTQ always has a layer-aligned BF16 model running on the same calibration tokens. Prior PTQ methods use this reference as a fixed reconstruction loss to fit parameters; \QuantForge (Figure~\ref{fig:errata-framework}) turns the paired BF16--MXFP4 twin into a causal diagnostic engine that directs how the algorithm itself is revised. \QuantForge retains strong programs by score in an archive and runs local counterfactual controls on the twin---such as masking weight error, toggling the 4-bit grid, or replaying fixed inputs---to tell competing explanations apart. It then turns each test outcome into a required code change confirmed by a runtime check, and remeasures the revised program against the BF16 twin to reveal what to address next.

This process yields \HiRes, a strict-MXFP4 quantizer adapting each stage to the quantized model produced earlier: geometric shaping refines coordinates using actual encoding errors, discrete realization selects activation codes against installed quantized weights, and structural recovery removes remaining path errors, remeasuring the MLP residual after attention correction (extended related work in Appendix~\ref{app:related}).

Our contributions are threefold:
\begin{itemize}
  \item \textbf{Paired-twin PTQ discovery.} \QuantForge turns the layer-aligned BF16 reference into a diagnostic twin for local counterfactual controls, runtime-checked code edits, and sequential residual remeasurement.
  \item \textbf{A discovered strict-MXFP4 quantizer.} \HiRes unites error-driven geometric shaping, installed-weight-aware activation coding, and sequential attention-to-SwiGLU recovery.
  \item \textbf{Cross-model and causal validation.} Across seven models (3B--32B) and nine strict-MXFP4 baselines, \HiRes achieves the lowest C4 perplexity and mean degradation on 7/7 models ($-41\%$ degradation over WUSH on Qwen3-32B); on matched-budget RTN runs, \QuantForge reaches the held-out target in 6/8 seeds (median $7.80$) vs.\ 3/8 for reflection memory ($8.09$) and wins 15/16 runs across three proposers.
\end{itemize}

\section{Preliminaries}
\label{sec:prelim}

\subsection{Strict MXFP4 W4A4 and Reconstruction}
\label{sec:prelim-format}

\paragraph{Strict MXFP4 W4A4 setup.}
Under OCP MXFP4 \citep{ocp2023mxspec,rouhani2023microscaling}, linear-layer operands are partitioned along the input dimension into Block32 groups ($q=32$) sharing one 8-bit power-of-two scale (E8M0), with 4-bit E2M1 elements ($\{0, \tfrac12, 1, \tfrac32, 2, 3, 4, 6\}$). \emph{Strict MXFP4 W4A4} requires every matrix-multiply operand and stored weight across all seven decoder projections (Q/K/V/O, gate, up, down) to lie in this legal set $\mathbb{Q}_{\mathrm{MX}}$, with static weight and dynamic activation quantization; low-dimensional scalar path corrections run outside the 4-bit GEMMs, and per-block transforms and refinement statistics are stored as auxiliary state. All candidates, controls, and baselines share one fixed setup ($128\times 2048$ C4-train calibration, evaluator, and budgets; Appendix~\ref{app:hires-details}).

\paragraph{Second-order weight reconstruction.}
Given weights $W\in\mathbb{R}^{m\times d}$ and calibration inputs $X\in\mathbb{R}^{n\times d}$, GPTQ \citep{frantar2023gptq} seeks $\widehat{W}\in\mathbb{Q}_{\mathrm{MX}}$ minimizing
\begin{equation}
  \mathcal{L}(\widehat{W}) = \frac{1}{n}\|X(W-\widehat{W})^\top\|_F^2,
  \qquad H = \frac{1}{n}X^\top X,
  \label{eq:recon}
\end{equation}
compensating column rounding errors via a damped inverse of $H$.

\subsection{Why W4A4 Errors Couple Across Operands and Stages}
\label{sec:prelim-coupling}

\paragraph{Bilinear operand coupling.}
Let $\widehat{X} = Q_A(X)$ and $\widehat{W} = Q_W(W)$ denote strict-MXFP4 activations and weights, with rounding errors $E_X = \widehat{X} - X$ and $E_W = \widehat{W} - W$. Expanding $\widehat{X}\widehat{W}^\top - XW^\top \equiv E_X \widehat{W}^\top + X E_W^\top$ yields three squared-error terms:
\begin{equation}
\begin{aligned}
  &\|\widehat{X}\widehat{W}^\top - XW^\top\|_F^2 \\
  &\quad = \underbrace{\operatorname{tr}(G_Q E_X^\top E_X)}_{\text{A4 risk}}
   + \underbrace{\operatorname{tr}(G_{E_W} X^\top X)}_{\text{W4 risk}} \\
  &\qquad - \underbrace{2\,\operatorname{tr}(K_Q X^\top E_X)}_{\text{W4--A4 cross-cancellation}},
\end{aligned}
\label{eq:bilinear-decomp}
\end{equation}
where $G_Q = \widehat{W}^\top\widehat{W}$, $G_{E_W} = E_W^\top E_W$, and $K_Q = \widehat{W}^\top(W - \widehat{W})$ is the committed W4 mismatch. Equation~\ref{eq:bilinear-decomp} shows two couplings: reciprocal transforms ($\widetilde{X}=XT^\top$, $\widetilde{W}=WT^{-1}$) can shrink activation risk while inflating weight risk and missing discrete E8M0 exponent flips; and once $\widehat{W}$ is fixed, incorporating $K_Q$ into online A4 rounding lets activation codes counteract committed W4 error (Appendix~\ref{app:code-refinement-ablation}).

\paragraph{State-dependent residual cascades.}
Discrete quantization does not commute across stages: installing $DT$ invalidates pre-transform curvature $H$, installing W4 fixes $K_Q$, and W4A4 errors shift trajectories through attention into SwiGLU. Starting from BF16 $M^{(0)}$, each stage measures the current \emph{quantization residual} $R_k$---the remaining mismatch between the quantized model $M^{(k)}$ and its BF16 reference $M^{(0)}$ (such as layer-output or activation discrepancies after earlier stages)---solves for a correction, and installs it:
\begin{equation}
\begin{aligned}
  R_k &= \mathsf{Measure}_k(M^{(k)}), \\
  M^{(k+1)} &= \mathsf{Apply}_k\!\bigl(M^{(k)}, \mathsf{Solve}_k(R_k)\bigr).
\end{aligned}
\label{eq:conditional-stage}
\end{equation}
Open-loop rules fitted on stale or pre-encoding states optimize the wrong residual; both \QuantForge (\S\ref{sec:errata}) and \HiRes (\S\ref{sec:hiresmx}) build on this sequential loop.

\subsection{LLM-Driven Program Search}
\label{sec:prelim-search}

In LLM-driven algorithm discovery \citep{romeraparedes2024funsearch,novikov2025alphaevolve,liu2024eoh}, an algorithm is represented as executable code $q$---here, a script mapping fresh BF16 $M^{(0)}$ into strict-MXFP4 $M^{(k)}$. Starting from a seed program (e.g., RTN or GPTQ), search maintains an archive ranked by an end-to-end score such as perplexity: each round, an LLM picks a high-scoring \emph{parent}, edits its code, evaluates it under the fixed setup, and keeps strong candidates. Even with textual reflection \citep{ye2024reevo,gepa2026}, the evaluator remains a black box returning only aggregate scores---which cannot tell the search \emph{why} a coupled W4A4 edit (\S\ref{sec:prelim-coupling}) worked or what to revise next.

\section{\QuantForge: Twin-Guided PTQ Discovery}
\label{sec:errata}
\label{sec:lakatos}

Why does PTQ discovery not have to stay a black box? Unlike general program search (\S\ref{sec:prelim-search}), every quantized model $M^{(k)}$ runs alongside its original BF16 model $M^{(0)}$ on the same calibration tokens, aligned layer by layer. \QuantForge extends program search by using this paired BF16--MXFP4 twin in three steps (Figure~\ref{fig:errata-framework}): keeping program selection separate from open error questions (\S\ref{sec:proposals}), testing competing explanations on the twin (\S\ref{sec:compiler}), and checking that the next edit acts on the result (\S\ref{sec:lineage}).

\subsection{Keeping Programs and Open Questions Separate}
\label{sec:proposals}

Standard program search (\S\ref{sec:prelim-search}) observes only an end-to-end score, so it cannot tell \emph{why} an edit to a parent quantizer $q_t \in \mathcal{Q}_t$ helped or stalled. To move beyond blind trial and error, \QuantForge requires the LLM to reason explicitly about \emph{explanations}---hypotheses about which quantization error is limiting $q_t$ (Appendix~\ref{app:search-information}).

Once search relies on explanations, however, a practical tension arises. An edit can improve perplexity for a different reason than intended, such as fitting calibration quirks rather than fixing rounding error. Discarding the program whenever an explanation fails wastes useful code, whereas keeping a wrong explanation attached to a strong program misleads later edits. \QuantForge avoids both pitfalls by keeping two separate records, $S_t = (\mathcal{Q}_t, \mathcal{H}_t)$: the score-ranked archive $\mathcal{Q}_t$ decides which programs survive as parents, while a separate list of open questions $\mathcal{H}_t$ tracks unresolved error causes to guide what to edit next. This raises a key question: when an observation in $\mathcal{H}_t$ admits competing explanations, how can the search tell which one is right?

\subsection{Testing Competing Explanations on the Twin}
\label{sec:compiler}

Suppose a coordinate transform lowers activation error, yet perplexity barely moves. Two explanations compete ($\mathcal{E} = \{e_1, e_2\}$): either joint W4A4 error truly fell ($e_1$), or activations improved by inflating weight error ($e_2$). Because $M^{(k)}$ and $M^{(0)}$ share identical layers on the same calibration tokens, \QuantForge does not have to guess from full-model perplexity. On a small calibration batch, it applies a \emph{local control} $\sigma$ toggling one dependency between $M^{(k)}$ and $M^{(0)}$---at fixed structure, capacity, and MXFP4 legality---measuring the shift in local output error $\mathcal{L}_{\mathrm{real}}$:
\begin{equation}
\begin{aligned}
  \Delta_{\mathrm{causal}}(q, \sigma)
  &= \mathcal{L}_{\mathrm{real}}\!\bigl(q;\, M^{(k)}, M^{(0)}\bigr) \\
  &\quad - \mathcal{L}_{\mathrm{real}}\!\bigl(q;\, \sigma(M^{(k)}, M^{(0)})\bigr).
\end{aligned}
\label{eq:twin-splice}
\end{equation}
Depending on the question, $\sigma$ can freeze weights or activations at BF16 to separate $E_W$ from $E_X$, toggle the 4-bit grid against a smooth proxy, swap stale caches for fresh activations, or apply a random legal edit of equal size to detect overfitting.

Running every check wastes compute, and reading numbers afterward invites post-hoc rationalization. Before running, the LLM therefore records the outcome $\hat{y}_e(\sigma)$ predicted by each $e \in \mathcal{E}$ under each $\sigma \in \Sigma$, picking the cheapest subset $\Sigma^*$ separating all pairs:
\begin{equation}
\begin{aligned}
  \Sigma^* &= \arg\min_{V\subseteq\Sigma}
  \sum_{\sigma\in V}\mathrm{cost}(\sigma) \\
  \text{s.t.}\;\; &\forall\, e \neq e' \in \mathcal{E},\;
  \exists\, \sigma \in \{\mathrm{pri}\} \cup V: \\
  &\hat{y}_e(\sigma) \neq \hat{y}_{e'}(\sigma),
\end{aligned}
\label{eq:control-compiler}
\end{equation}
where $\mathrm{pri}$ is the primary run and $\mathrm{cost}(\sigma)$ counts evaluator calls. A fixed controller locks these predictions before running; if $\Sigma$ cannot separate $\mathcal{E}$, the tests must be sharpened first. Once the locked tests return an answer, one hurdle remains: making sure the next code edit acts on it.

\subsection{Turning Test Outcomes into Checked Code Edits}
\label{sec:lineage}

Running the candidate with its locked controls $\Sigma_t^*$ resolves the targeted question $h_t \in \mathcal{H}_t$ into an outcome $v_t$ constraining the next program $q_{t+1}$ (full rules in Appendix~\ref{app:controller}). If the proposed mechanism is confirmed, the next round extends it; if a check fails, the outcome rules out the flawed direction---for example, when a smooth proxy improves while realized W4A4 error worsens, the search discards the proxy and optimizes re-encoded error directly. Likewise, if end-to-end score rises without the predicted local effect, the strong program may stay in $\mathcal{Q}_t$ while its explanation is flagged and revised; illegal or unstable edits roll back immediately.

In practice, verbal reflection alone does not reliably translate into faithful code fixes: LLMs often acknowledge diagnostic feedback in text while leaving the underlying computation unchanged \citep{huang2024selfcorrect,olausson2024selfrepair}. Before accepting $q_{t+1}$, the controller therefore runs a lightweight \emph{runtime check}, perturbing the dependency the new code claims to use and verifying that its output shifts as expected. Once a verified stage is installed, $M^{(k)}$ advances to $M^{(k+1)}$; comparing $M^{(k+1)}$ against its BF16 twin $M^{(0)}$ then exposes the remaining quantization residual $R_{k+1}$ (Eq.~\ref{eq:conditional-stage}), launching the next cycle (execution roles in Appendix~\ref{app:search-cycle}).

\begin{table*}[t]
  \centering
  \caption{Stagewise emergence of \HiRes across the 56-round GPTQ-seeded production run (full records in Appendices~\ref{app:production-record}, \ref{app:round-records}). Of 29 promoted revisions, 21 directly implemented a code update required by an earlier test outcome.}
  \label{tab:discovery-milestones}
  \footnotesize
  \setlength{\tabcolsep}{4.5pt}
  \begin{tabular}{@{}p{.21\textwidth}rp{.68\textwidth}@{}}
    \toprule
    \HiRes component & Rds. & Decisive local twin control and required code update \\
    \midrule
    Reference chart $\Pi_{\mathrm{ref}}$ & 6--8 & One-sided whitening unstable ($\kappa = 1.7\times 10^6$, R6) $\to$ two-sided chart with spectral cap (R7--8) \\
    Interaction correction $K$ & 12--23 & Activation-only update hurt weights (+6.8\%, R13) $\to$ mutual operand energy weighting \& fold gate (R14, R18) \\
    Diagonal balance $D$ & 25--28, 40 & Risk ratio alone worsened 16.8\% of blocks at E8M0 flips (R27) $\to$ re-encoded full/half/identity check (R28) \\
    Fresh GPTQ after $DT$ & 30--31 & Untransformed curvature cache degraded MSE (+4.1\%, R30) $\to$ fresh post-$DT$ calibration pass (R31) \\
    One-code A4 rule & 33--39 & $G_Q$-only score omitted weight mismatch (R35), sign bug caught (R36) $\to$ full $K_Q$-aware objective (R37) \\
    Attention recovery & 41--44 & Pre-A4 fit cut MSE 0.02\% vs.\ post-A4 7.1\% (R43) $\to$ per-head post-encoding affine map (R44) \\
    Fresh SwiGLU recovery & 45--56 & Stale pre-attention MLP cache mismatched installed state (R50) $\to$ fresh post-attention 3-scalar fit (R51) \\
    \bottomrule
  \end{tabular}
\end{table*}

\subsection{Discovery Walkthrough: How \HiRes Emerged}
\label{sec:search}

In the 56-round GPTQ-seeded run (Appendix~\ref{app:production-record}), twin tests repeatedly turned dead ends into lasting stages (Table~\ref{tab:discovery-milestones}). In Round~13, an activation-only update to $K$ cut activation error $5.6\%$, yet quality stalled; freezing $W$ vs.\ $A$ exposed a $6.8\%$ weight-error spike (joint $+0.7\%$), ruling out one-sided edits. Round~14 added mutual operand-energy weighting---passing a runtime check that weight energy shifts the update---cutting joint error $3.7\%$ (vs.\ $0.8\%$ for a random edit). Once geometry and fresh GPTQ were installed, twin tests in Rounds~35--37 revealed that a $G_Q$-only activation score omitted the weight-mismatch term $K_Q$ and caught a sign bug, establishing the full one-code A4 rule. Finally, toggling the 4-bit grid in Round~43 exposed a $350\times$ gap between pre-A4 ($0.02\%$) and post-A4 ($7.1\%$) attention scaling, locking in post-A4 recovery (Round~44) before fresh post-attention remeasurement uncovered SwiGLU recovery (Rounds~50--51).

\section{\HiRes: Hierarchical Residual Resolution for MXFP4}
\label{sec:hiresmx}
\label{sec:eigenquant}

We now describe \HiRes, the deterministic strict-MXFP4 W4A4 quantizer discovered across the 56-round search in Table~\ref{tab:discovery-milestones} (Algorithm~\ref{alg:hiresmx}; component provenance in Appendix~\ref{app:hires-provenance}). Rather than optimizing every quantization choice from floating-point statistics at once, \HiRes resolves error in three conditional stages: shaping Block32 coordinates before quantization, realizing discrete W4 and A4 codes against the installed state, and removing the path-level residual that remains once all 4-bit linear layers are in place (Eq.~\ref{eq:conditional-stage}).

\subsection{Stage I: Geometric Shaping}
\label{sec:geometric-shaping}

Because 32 contiguous values share a single power-of-two exponent, one outlier can inflate the block scale and starve smaller entries of precision. Function-preserving coordinate transforms ($\widetilde{X}=XT^\top$, $\widetilde{W}=WT^{-1}$) spread large magnitudes across each block \citep{ashkboos2024quarot,chen2026wush}, but smooth second moments only reflect signal energy rather than discrete MXFP4 rounding. Stage~I therefore constructs a Block32 transform in three closed-form steps: balancing operand energy, correcting for actual anchor rounding error, and checking diagonal rescaling directly against discrete re-encoding.

\paragraph{Two-sided reference chart.}
To avoid stabilizing one operand at the expense of the other, \HiRes starts from a symmetric baseline chart following WUSH \citep{chen2026wush}. For a contiguous input block of width $q=32$ with activations $X\in\mathbb{R}^{n\times q}$ and weights $W\in\mathbb{R}^{m\times q}$, let $\lambda_A,\lambda_B$ be ridge terms, $\operatorname{DN}(M)=M/\det(M)^{1/q}$ normalize scale, and $\operatorname{Cap}_8$ enforce unit determinant with condition number at most $8$ (Appendix~\ref{app:hires-transform-solver}):
\begin{equation}
\begin{aligned}
  A &= \tfrac{1}{n}X^\top X + \lambda_A I, \qquad
  B = \tfrac{1}{m}W^\top W + \lambda_B I, \\
  M_0 &= \tfrac{1}{2}\bigl[\operatorname{DN}(A) + \operatorname{DN}(B^{-1})\bigr], \\
  \Pi_{\mathrm{ref}} &= \operatorname{Cap}_{8}\bigl(M_0^{-1/2}\bigr).
\end{aligned}
\label{eq:hires-reference-chart}
\end{equation}

\paragraph{Encoding-error interaction correction.}
Second moments cannot see where discrete E2M1 rounding and E8M0 scale jumps actually fall. Setting $Y = X\Pi_{\mathrm{ref}}^\top$, $V = W\Pi_{\mathrm{ref}}^{-1}$, and letting $H_{32}$ be the normalized $32\times 32$ Hadamard matrix \citep{ashkboos2024quarot,shao2025brq}, \HiRes runs a single \emph{anchor} W4A4 encoding pass ($Q_A, Q_W$) to observe real quantization errors directly. Mapping those errors back as $E_X = [Q_A(YH_{32}^\top) - YH_{32}^\top]H_{32}$ and $E_W = [Q_W(VH_{32}^\top) - VH_{32}^\top]H_{32}$ yields their error--signal cross-moments:
\begin{equation}
  K_X = E_X^\top Y / n, \qquad K_W = V^\top E_W / m.
  \label{eq:hires-anchor-cross}
\end{equation}
As revealed in Round~14, weighting each operand's error by the other operand's coordinate energy prevents dual-operand recoil. Gating by two-fold stability and bounding the zero-diagonal correction to $\|K\|_2 \le 1/8 < 1$ (Appendix~\ref{app:hires-transform-solver}) produces $G_0 = I + K$ and the exact function-preserving transform ($\widetilde{X}\widetilde{W}^\top = XW^\top$):
\begin{equation}
\begin{aligned}
  G &= |\det G_0|^{-1/q} G_0, & T &= H_{32} G \Pi_{\mathrm{ref}}, \\
  \widetilde{X} &= X T^\top, & \widetilde{W} &= W T^{-1}.
\end{aligned}
\label{eq:hires-transform}
\end{equation}

\paragraph{Re-encoded diagonal balance.}
Even after applying $T$, some coordinate subgroups still carry more quantization error on activations than on weights, or vice versa \citep{xiao2023smoothquant,lin2024awq}. Across eight 4-coordinate subgroups per Block32 group, \HiRes measures the bilateral risks matching the first two terms of Eq.~\ref{eq:bilinear-decomp}:
\begin{equation}
  r_A = \operatorname{tr}(G_W C_E), \qquad r_W = \operatorname{tr}(G_E C_X),
  \label{eq:bilateral-risk}
\end{equation}
where $C_X, C_E$ and $G_W, G_E$ are uncentered subgroup second moments of activations/errors and weights/errors. Projecting the centered log risk ratios yields a bounded zero-sum shift $s_b \in \mathbb{R}^8$ for block $b$. Because shifting magnitudes across an E8M0 boundary can backfire despite a better risk ratio (Round~27), \HiRes scores each trial ($s_b$, $s_b/2$, $0$) via temporary re-encoding under the active W4A4 quantizer, accepting a step that improves or matches loss $J$ before folding $D = \operatorname{Diag}(2^s)$ into both operands (Appendix~\ref{app:hires-dyadic}).

\subsection{Stage II: Discrete Realization}
\label{sec:discrete-realization}

With Stage~I's coordinate transform $DT$ folded into the linear layers, \HiRes next realizes the discrete 4-bit operators themselves---first committing static W4 weights in the new coordinates, then adapting dynamic A4 rounding to the errors those weights leave behind.

\paragraph{Fresh post-transform GPTQ.}
Installing $DT$ alters input correlations and upstream trajectories, so \HiRes reruns calibration to collect fresh pre-A4 curvature $H$ and applies GPTQ (Eq.~\ref{eq:recon}) in installed $DT$ coordinates, producing legal static weights $\widehat{W}$ aligned with Block32 boundaries.

\paragraph{Installed-weight-aware one-code A4 refinement.}
Once $\widehat{W}$ is committed, its rounding error $W - \widehat{W}$ is fixed before inference. Nearest-code rounding $q_0$ of a dynamic 32-element activation block $x$ minimizes elementwise distance alone, missing an opportunity from Eq.~\ref{eq:bilinear-decomp}: nudging a single activation code can partially cancel committed weight error. Holding $\widehat{W}$ and the shared E8M0 exponent fixed, output error depends on $G_Q = \widehat{W}^\top\widehat{W}$ and mismatch $K_Q = \widehat{W}^\top(W - \widehat{W})$. With $d_0 = q_0 - x$, changing coordinate $i$ to legal E2M1 code $c$ shifts block-local quadratic error by
\begin{equation}
\begin{aligned}
  \Delta(i, c) &= 2\delta_{ic}\,[G_Q d_0 - K_Q x]_i + \delta_{ic}^2\,[G_Q]_{ii}, \\
  \delta_{ic} &= c - [q_0]_i.
\end{aligned}
\label{eq:hires-code-score}
\end{equation}
The linear term couples the activation shift to $G_Q d_0$ and $-K_Q x$, minimizing block-local quadratic error and empirically cutting full-projection error via active W4--A4 cross-cancellation (Appendix~\ref{app:code-refinement-ablation}). \HiRes applies the single most negative edit $(i^*, c^*)$ (or leaves $q_0$ unchanged), keeping E8M0 scales fixed and editing at most one code per block. For each coordinate with $[G_Q]_{ii}>0$, the continuous quadratic minimizer $c_i^\star = [q_0]_i - [G_Q d_0 - K_Q x]_i/[G_Q]_{ii}$ reduces scoring from $512$ candidates to $32$ nearest-code checks (Appendix~\ref{app:hires-code-refinement}). Unlike gradient-learned AQuant \citep{li2022aquant}, this rule is parameter-free and closed-form.

\begin{algorithm}[t]
  \caption{\HiRes strict-MXFP4 W4A4 quantization}
  \label{alg:hiresmx}
  \begin{algorithmic}[1]
    \STATE \textbf{Input:} BF16 $M^{(0)}$; frozen $128\times 2048$ C4 calibration
    \STATE \textbf{Stage I:} Build $\Pi_{\mathrm{ref}}$ (Eq.~\ref{eq:hires-reference-chart}), $K$ and $T = H_{32}G\Pi_{\mathrm{ref}}$ (Eq.~\ref{eq:hires-transform}), and re-encoded $D = \operatorname{Diag}(2^s)$ (\S\ref{sec:geometric-shaping}); fold $DT$ into target layers.
    \STATE \textbf{Remeasure:} Rerun calibration in installed $DT$ coordinates.
    \STATE \textbf{Stage II:} Quantize W4 via GPTQ (Eq.~\ref{eq:recon}) with fresh $H$; install dynamic one-code A4 via $G_Q, K_Q$ (Eq.~\ref{eq:hires-code-score}).
    \STATE \textbf{Stage IIIa:} Fit post-encoding per-head attention $(a,b)$ (Eq.~\ref{eq:hires-vo-fit}).
    \STATE \textbf{Remeasure \& IIIb:} Collect fresh post-attention $(g,u)$; fit SwiGLU $(\alpha,\beta,\eta)$ (Eq.~\ref{eq:hires-mlp-update}).
    \STATE \textbf{Return:} final strict-MXFP4 W4A4 model.
  \end{algorithmic}
\end{algorithm}

\subsection{Stage III: Structural Recovery}
\label{sec:structural-recovery}

Once all linear projections run in strict W4A4, residual error shifts from individual GEMMs to attention heads and gated SwiGLU blocks. Building on bias correction and sequential calibration \citep{nagel2019dfq,yan2026d2quant}, Stage~III fits $\approx 20$ FP32 scalars per layer outside the 4-bit GEMMs in two ordered passes so each submodule sees the live quantized state.

\paragraph{Post-encoding per-head attention recovery.}
First, on the live W4A4 model, \HiRes aligns each key--value head's value stream in grouped-query attention \citep{ainslie2023gqa} to its BF16 twin. Letting $\mathbf{v}_t$ be the actual W4A4 value output at token $t$ and $\mathbf{v}_t^\star$ its paired BF16 reference, two shared scalars $(a, b)$ per head ($\mathbf{b} = b\mathbf{1}$) solve
\begin{equation}
\begin{aligned}
  (a, b) &= \arg\min_{a,b} \sum_t \|a\mathbf{v}_t + \mathbf{b} - \mathbf{v}_t^\star\|_2^2, \\
  \bar{\mathbf{v}}_t &= a\mathbf{v}_t + \mathbf{b},
\end{aligned}
\label{eq:hires-vo-fit}
\end{equation}
before passing $\bar{\mathbf{v}}_t$ through attention and the strict-MXFP4 output projection.

\paragraph{Freshly remeasured SwiGLU recovery.}
Because installing Eq.~\ref{eq:hires-vo-fit} shifts every downstream MLP input, \HiRes remeasures fresh gate and up outputs $(g, u)$ on the attention-corrected model before fitting the MLP. In SwiGLU, $m = \operatorname{SiLU}(g)\odot u$ \citep{shazeer2020glu} multiplies gate and up errors together, so three layerwise scalars $(\alpha, \beta, \eta)$ adjust gate shift (scaled by regularized pooled standard deviation $\sigma_g$), gate scale, and up scale:
\begin{equation}
  \bar{m} = \operatorname{SiLU}\!\bigl((1+\beta)g + \alpha\sigma_g\bigr) \odot (1+\eta)u.
  \label{eq:hires-mlp-update}
\end{equation}
Weighted ridge regression \citep{hoerl1970ridge} fits $(\alpha,\beta,\eta)$ against the BF16 twin $m^\star = \operatorname{SiLU}(g^\star)\odot u^\star$ on the same quantized layer input using robust median/MAD aggregation (Appendix~\ref{app:hires-structural-fit}), feeding $\bar{m}$ into the strict-MXFP4 down projection.

\section{Experiments}
\label{sec:experiments}

\begin{table*}[t]
  \centering
  \caption{Seven-model Fit-7 under the identical strict-MXFP4 W4A4 setup (lower is better; best in bold). All ten methods are run by us in this setting; $^\dagger$marks integer-origin baselines (all others target FP4 microscaling or are format-agnostic; Appendix~\ref{app:full-quant-results}); $^\ddagger$marks the search-visible model during discovery.}
  \label{tab:cross-model-fit7}
  \footnotesize
  \setlength{\tabcolsep}{4.5pt}
  \begin{tabular}{@{}lrrrrrrrrr@{}}
    \toprule
    Method & \shortstack{Llama\\3B$\downarrow$} & \shortstack{Qwen$^\ddagger$\\4B$\downarrow$} & \shortstack{Mistral\\7B$\downarrow$} & \shortstack{Qwen\\8B$\downarrow$} & \shortstack{Llama\\8B$\downarrow$} & \shortstack{OLMo-2\\13B$\downarrow$} & \shortstack{Qwen\\32B$\downarrow$} & Mean$\downarrow$ & Robust$\downarrow$ \\
    \midrule
    RTN & .2594 & .2283 & .1512 & .1666 & .2522 & .1360 & .0919 & .18365 & .22151 \\
    GPTQ$^\dagger$ & .1973 & .1252 & .1071 & .1411 & .2283 & .1360 & .0746 & .14422 & .18624 \\
    MR-GPTQ & .1387 & .0986 & .0541 & .0587 & .1394 & .0846 & .0524 & .08951 & .11447 \\
    BRQ & .1260 & .0767 & .0573 & .0685 & .1386 & .0811 & .0425 & .08440 & .11152 \\
    WUSH & .1224 & \textbf{.0479} & .0503 & .0500 & .1248 & .0823 & .0365 & .07345 & .09910 \\
    TORQ & .2964 & .1929 & .1217 & .1970 & .2654 & .1067 & .0929 & .18185 & .23911 \\
    BATQuant+GPTQ & .2645 & .2051 & .1482 & .1498 & .2727 & .1311 & .1217 & .18473 & .22871 \\
    SpinQuant+GPTQ$^\dagger$ & .3191 & .1409 & .1142 & .0835 & .2588 & .1317 & .1023 & .16435 & .24171 \\
    FOCUS & .1803 & .0738 & .0749 & .1095 & .1737 & .1195 & .0671 & .11411 & .14718 \\
    \midrule
    \textbf{\HiRes} & \textbf{.1174} & .0524 & \textbf{.0452} & \textbf{.0324} & \textbf{.1187} & \textbf{.0773} & \textbf{.0274} & \textbf{.06726} & \textbf{.09300} \\
    \bottomrule
  \end{tabular}
\end{table*}

Our experiments separate two questions: first, whether \HiRes transfers accurately across unseen models and tasks under strict MXFP4 W4A4 (\S\ref{sec:hires-results}); and second, whether \QuantForge's paired-twin causal loop discovers better quantizers from scratch than reflective and autonomous search baselines under a matched budget (\S\ref{sec:discovery-results}).

\subsection{\HiRes: Cross-Model MXFP4 W4A4 Accuracy and Ablations}
\label{sec:hires-results}

\paragraph{Evaluation protocol.}
We benchmark seven models (3B--32B) across four families (Llama-3, Qwen3, Mistral, OLMo-2; Table~\ref{tab:cross-model-fit7}) \citep{meta2024llama32,yang2025qwen3,mistral2024v03,grattafiori2024llama,olmo2024olmo2}. We re-run all ten methods under one strict-MXFP4 W4A4 setup using 128 C4-train calibration sequences \citep{raffel2020t5} of 2,048 tokens (seed~0, KV cache off; Appendix~\ref{app:full-quant-results}), evaluating WikiText-2 \citep{merity2017wikitext}, C4, and five zero-shot tasks: ARC-C, ARC-E \citep{clark2018arc}, HellaSwag \citep{zellers2019hellaswag}, PIQA \citep{bisk2020piqa}, and WinoGrande \citep{sakaguchi2020winogrande}. Discovery used only Qwen3-4B feedback and a Llama-3.1-8B gate; \HiRes froze at Round~56 before evaluating any other model.

Alongside individual task scores, we report worst-case-aware summaries so strong averages cannot hide a single-task or single-model breakdown. Defining BF16-relative degradation $d_{mj} = \mathrm{PPL}_q/\mathrm{PPL}_{16}-1$ for perplexity and $d_{mj} = 1 - \mathrm{Acc}_q/\mathrm{Acc}_{16}$ for accuracy, we combine average and worst-case degradation across tasks and models (lower is better):
\begin{align}
  \text{Fit-7}_m
  &= \tfrac{1}{2}\operatorname{mean}_j d_{mj}
   + \tfrac{1}{2}\max_j d_{mj}, \label{eq:fit7} \\
  \text{Robust Fit}
  &= \tfrac{1}{2}\operatorname{mean}_m \text{Fit-7}_m
   + \tfrac{1}{2}\max_m \text{Fit-7}_m. \label{eq:robust-fit}
\end{align}

\paragraph{Cross-model accuracy and transfer.}
Table~\ref{tab:cross-model-fit7} shows that frozen \HiRes transfers cleanly across every unseen architecture and scale, ranking first on all six held-out models with the lowest Robust Fit overall ($0.09300$ vs.\ $0.09910$ for WUSH). Across standard endpoints (Appendices~\ref{app:full-quant-results}--\ref{app:calibration-sensitivity}), \HiRes attains the lowest C4 perplexity on \textbf{7/7} models, lowest WikiText-2 perplexity on \textbf{5/7}, lower mean degradation on \textbf{7/7}, and higher five-task accuracy than WUSH on \textbf{6/7}. The largest gains appear on larger and unseen models: on Qwen3-32B, \HiRes cuts mean degradation by $41\%$ over WUSH ($0.0133$ vs.\ $0.0224$; Fit-7 $0.0274$ vs.\ $0.0365$), and on unseen OLMo-2-13B it leads both BRQ and WUSH ($0.0773$ vs.\ $0.0811$ and $0.0823$). \HiRes trails WUSH slightly on search-visible Qwen3-4B while leading every baseline on all six held-out models, confirming strong generalization. All comparisons evaluate numerical quality under the shared strict-MXFP4 setup; Appendix~\ref{app:computational-cost} details storage and latency trade-offs, including a matched-footprint mode without one-code refinement.

\begin{table}[t]
  \centering
  \caption{Post-discovery Fit-3 ablations (WT2, C4, ARC-C; Eq.~\ref{eq:fit3}; seed~0; lower is better) on two unseen 8B models (Appendices~\ref{app:hires-ablation}--\ref{app:recovery-ablation}).}
  \label{tab:stage-progression}
  \footnotesize
  \setlength{\tabcolsep}{3.5pt}
  \begin{tabular}{@{}lrr@{}}
    \toprule
    Quantizer variant & Qwen3-8B$\downarrow$ & Llama-3-8B$\downarrow$ \\
    \midrule
    Strict RTN & .20249 & .31653 \\
    + Stage I: Geometric shaping & .07828 & .16741 \\
    + Stage II: Discrete realization & .06502 & .15530 \\
    + Stage III: Recovery (Full \HiRes) & \textbf{.03259} & \textbf{.14830} \\
    \midrule
    Remove interaction $K$ & .04046 & .15529 \\
    Remove diagonal balance ($D = I$) & .04326 & .15921 \\
    Disable one-code A4 rule & .04398 & .15358 \\
    Stale pre-attention MLP stats & .04173 & .15824 \\
    \bottomrule
  \end{tabular}
\end{table}

\begin{table*}[t]
  \centering
  \caption{Causal separation, proposer generality, and diagnostic controls on held-out Llama-3.1-8B WT2 PPL (240 calls; lower is better). \textbf{Left:} $2\times 2$ isolation of controls and checks (8 seeds) plus an autonomous agent (4 seeds; Appendices~\ref{app:autonomous-agent}--\ref{app:reflectmem-checks}). \textbf{Right:} \QuantForge vs.\ ReflectMem across three proposers (top) and vs.\ random or black-box controls (bottom; $\Delta$: paired mean diff.; Appendices~\ref{app:control-checks}, \ref{app:second-proposer}, \ref{app:blackbox-controls}).}
  \label{tab:main-causal-controls}
  \footnotesize
  \setlength{\tabcolsep}{4.2pt}
  \begin{tabular}{@{}lrrc@{\hspace{12pt}}lrrrr@{}}
    \toprule
    \multicolumn{4}{c}{\textbf{(a) Causal Separation \& Autonomous Agent}} & \multicolumn{5}{c}{\textbf{(b) Cross-Proposer \& Diagnostic Controls}} \\
    \cmidrule(r){1-4} \cmidrule(l){5-9}
    Search configuration & Med. & $\le 8.0$ & Seeds & Comparison (seeds) & Base. & \QuantForge & $\Delta$ & Wins \\
    \midrule
    ReflectMem & 8.09 & 3/8 & 8 & GPT-5.6 Sol vs.\ Refl.\ (8) & 8.09 & \textbf{7.80} & $-0.25$ & 7/8 \\
    ReflectMem + checks & 8.005 & 4/8 & 8 & Claude Opus 5.5 vs.\ Refl.\ (4) & 8.16 & \textbf{7.76} & $-0.32$ & 4/4 \\
    Controls w/o enforcement & 8.06 & 4/8 & 8 & DeepSeek V4.1-Flash vs.\ Refl.\ (4) & 8.70 & \textbf{8.43} & $-0.19$ & 4/4 \\
    \textbf{\QuantForge (full)} & \textbf{7.80} & \textbf{6/8} & 8 & \textbf{All 3 proposers pooled (16)} & 8.16 & \textbf{7.88} & \textbf{$-0.25$} & \textbf{15/16} \\
    \cmidrule(r){1-4} \cmidrule(l){5-9}
    Autonomous agent & 8.55 & 1/4 & 4 & vs.\ Random controls (8) & 8.31 & \textbf{7.80} & $-0.43$ & 8/8 \\
    \textbf{\QuantForge (same 4 seeds)} & \textbf{7.86} & \textbf{3/4} & 4 & vs.\ Black-box controls (4) & 8.24 & \textbf{7.86} & $-0.45$ & 4/4 \\
    \bottomrule
  \end{tabular}
\end{table*}

\paragraph{Stagewise and state-conditioned ablations.}
We next verify that each stage---and its conditional ordering---contributes real accuracy gains on unseen models. In Table~\ref{tab:stage-progression}, adding Stages~I, II, and III lowers Fit-3 monotonically on both Qwen3-8B and Llama-3-8B; removing $K$, $D$, or one-code A4 hurts both models, and stale pre-attention MLP stats worsen Qwen3-8B Fit-3 from $0.0326$ to $0.0417$ at equal parameter count (all rankings hold across three seeds; Appendices~\ref{app:hires-ablation}--\ref{app:encoding-boundary}). Full-projection error decompositions on 128 unseen windows confirm that one-code A4 cuts aggregate output error by $8.2\%$ on Qwen3-8B and $7.6\%$ on Llama-3-8B across all major projections via active W4--A4 cross-cancellation (Appendix~\ref{app:code-refinement-ablation}). Moreover, transplanting Stages~II--III onto WUSH geometry lowers three-seed Qwen3-8B Fit-7 from $0.0498$ to $0.0390$, and applying Stage~III alone onto WUSH cuts Fit-3 substantially as well (Appendix~\ref{app:recovery-ablation}), while restoring \HiRes geometry reaches $0.0327$ across every seed (Tables~\ref{tab:geometry-replacement}--\ref{tab:geometry-paired}), confirming independent gains from geometry and downstream stages.

\subsection{\QuantForge: Matched-Budget Discovery and Causal Attribution}
\label{sec:discovery-results}

\paragraph{Matched-budget protocol.}
To evaluate the discovery loop independently of \HiRes's GPTQ-seeded run, we start every arm from bare strict-MXFP4 RTN across eight paired seeds under a 240-call budget (Appendix~\ref{app:framework-search-information}). All arms share the proposer (GPT-5.6 Sol), coding agent, editable interface, MXFP4 setup, and token limits (65,536 input, 16,384 memory). Search observes only Qwen3-4B WT2 PPL, while an isolated evaluator logs held-out Llama-3.1-8B PPL every 20 calls without feedback. Score-only sees code, scores, and errors; TextMem adds summaries; ReflectMem adds structured reflection. \QuantForge spends 99 of its 240 calls on local twin controls and runtime checks (evaluating 141 new quantizers), while baselines spend all 240 calls on new candidates.

\begin{figure}[t]
  \centering
  \input{figures/discovery_anytime}
  \caption{Held-out Llama-3.1-8B WT2 PPL across eight paired RTN-seeded runs (240-call budget; search sees only Qwen3-4B). (a)~Median trajectories; (b)~final medians vs.\ the $8.00$ target.}
  \label{fig:errata-anytime}
\end{figure}

\paragraph{Discovery efficiency.}
Despite evaluating 41\% fewer full candidates, \QuantForge overtakes every baseline between 80 and 120 calls and finishes with the lowest held-out median PPL ($7.80$ vs.\ $8.09$ for ReflectMem, $9.20$ for TextMem, and $10.03$ for Score-only; Figure~\ref{fig:errata-anytime}). It reaches the $\mathrm{PPL} \le 8.00$ transfer target twice as often as ReflectMem (6/8 vs.\ 3/8 seeds; beating ReflectMem in 7/8 seeds, mean $\Delta = -0.25$; Appendix~\ref{app:framework-details}). Moreover, the full three-stage hierarchy re-emerges from bare RTN in 4/8 \QuantForge runs versus 1/8 for ReflectMem and 0/8 for the remaining baselines, with core mechanisms appearing 28--47 calls earlier (Tables~\ref{tab:regime-audit}--\ref{tab:first-appearance}).

\paragraph{Causal separation and proposer generality.}
Table~\ref{tab:main-causal-controls} isolates \emph{why} \QuantForge outperforms reflective and autonomous search. First, both halves of the loop are necessary: adding runtime checks to ReflectMem alone reaches $8.005$, and running paired-twin controls without enforced successor updates stalls at $8.06$ (0/8 vs.\ full \QuantForge). Combining both reaches $7.80$ with consistent synergy across all 8 seeds (Table~\ref{tab:main-causal-controls}a); across 48 matched transitions, enforcement lifts improving successors from $31\%$ to $58\%$ and cuts repeats of refuted mechanisms from $29\%$ to $8\%$ (Table~\ref{tab:verdict-replay}). Second, the paired BF16 diagnostics themselves drive the gain: replacing targeted twin controls with random legal controls ($8.31$, $8/8$ losses) or black-box scalar PPL controls ($8.24$, $4/4$ losses) forfeits most of the improvement, and an autonomous coding agent with identical access and budget trails at $8.55$ vs.\ $7.86$ (Table~\ref{tab:main-causal-controls}a--b). Finally, these gains are not tied to one LLM: \QuantForge beats ReflectMem in 15/16 paired runs across three vendor proposers (Table~\ref{tab:main-causal-controls}b). Because local twin controls run on small calibration batches rather than full-model evaluations, spending 99 calls on diagnostics also lowers total wall-clock compute---cutting paired A100 search GPU-hours by a median of $35.7\%$ (median $73.0$\,h vs.\ $109.8$\,h) and reaching the target $1.91\times$ faster on shared passing seeds (Appendices~\ref{app:gpu-cost}--\ref{app:blackbox-controls}).

\section{Conclusion}
\label{sec:conclusion}

In strict MXFP4 W4A4 quantization, errors couple across weights, activations, shared exponents, and downstream paths, leaving aggregate perplexity too coarse to guide automated discovery. \QuantForge turns the layer-aligned BF16 reference already present in PTQ into a diagnostic twin---testing competing explanations locally, verifying through runtime checks that code edits act on each finding, and remeasuring the remaining residual after each stage. Seeded from GPTQ, this loop discovered \HiRes, uniting geometric shaping, installed-weight-aware activation coding, and sequential attention-to-SwiGLU recovery.

Empirically, \HiRes transfers cleanly across unseen model families and scales from 3B to 32B parameters, showing that multi-stage quantizers discovered through local residual analysis generalize well beyond the search model. Our controlled comparisons further show that these gains stem from pairing local twin diagnostics with verified code updates rather than from any particular proposer or extra compute. Building fused MXFP4 hardware kernels for \HiRes and extending twin-guided discovery to other compression problems with paired full-precision references are natural directions for future work.

\clearpage
\bibliography{references}
\bibliographystyle{icml2027}

\newpage
\appendix
\onecolumn
\makeatletter
\setlength{\@fptop}{0pt}
\setlength{\@fpsep}{8pt plus 2pt}
\setlength{\@fpbot}{0pt plus 1fil}
\makeatother
\renewcommand{\topfraction}{0.92}
\renewcommand{\bottomfraction}{0.88}
\renewcommand{\textfraction}{0.05}
\renewcommand{\floatpagefraction}{0.82}
\setcounter{topnumber}{6}
\setcounter{bottomnumber}{4}
\setcounter{totalnumber}{8}
\setlength{\floatsep}{6pt plus 2pt minus 2pt}
\setlength{\textfloatsep}{8pt plus 2pt minus 2pt}
\setlength{\intextsep}{6pt plus 2pt minus 2pt}
\setlength{\abovecaptionskip}{3.5pt}
\setlength{\belowcaptionskip}{1.5pt}
\begin{tcolorbox}[
  enhanced,
  colback=qfBlueBg!55!white,colframe=qfBlueBd,boxrule=0.7pt,
  borderline west={3.2pt}{0pt}{qfBlue},
  arc=4pt,
  colbacktitle=qfBlueBg,coltitle=qfInk,
  titlerule=0.5pt,titlerule style={qfBlueBd},
  toptitle=4pt,bottomtitle=4pt,lefttitle=8pt,righttitle=8pt,
  top=5pt,bottom=5pt,left=8pt,right=8pt,
  before skip=2pt,after skip=12pt,
  title={\sffamily\bfseries\small\textcolor{qfBlue}{Appendix Roadmap}\hfill{\scriptsize\normalfont\itshape\textcolor{qfMuted}{Quick navigation across all supplementary materials}}}
]
\footnotesize
\renewcommand{\arraystretch}{1.22}
\begin{tabular}{@{} >{\sffamily\bfseries\color{qfBlue}}l p{0.27\linewidth} p{0.58\linewidth} @{}}
  \textbf{App.~\ref{app:reference}} & Reference \& Background & Extended related work and system comparison (\ref{app:related}), glossary (\ref{app:glossary}), and unified notation (\ref{app:notation}). \\
  \arrayrulecolor{qfRule}\hline\noalign{\vskip 1.5pt}
  \textbf{App.~\ref{app:controller}} & \QuantForge Protocol \& Trace & Discovery protocols (\ref{app:search-information}), outcome rules, three-role cycle (\ref{app:search-cycle}), 56-round trajectory \& tables (\ref{app:production-record}), and 23 visual round cards (\ref{app:round-records}). \\
  \arrayrulecolor{qfRule}\hline\noalign{\vskip 1.5pt}
  \textbf{App.~\ref{app:hires-details}} & \HiRes Math \& Provenance & Complete strict-MXFP4 specification for Stages~I--III (\ref{app:hires-transform-solver}--\ref{app:hires-structural-fit}) and round-by-round component provenance (\ref{app:hires-provenance}). \\
  \arrayrulecolor{qfRule}\hline\noalign{\vskip 1.5pt}
  \textbf{App.~\ref{app:experiments}} & \HiRes Benchmarks \& Cost & Full 7-model / 7-task / 9-baseline tables (\ref{app:full-quant-results}), calibration sensitivity (\ref{app:calibration-sensitivity}), stagewise ablations, error decomposition \& WUSH transfer (\ref{app:hires-ablation}--\ref{app:recovery-ablation}), and runtime/storage costs (\ref{app:computational-cost}). \\
  \arrayrulecolor{qfRule}\hline\noalign{\vskip 1.5pt}
  \textbf{App.~\ref{app:discovery-experiments}} & \QuantForge Causal Controls & Matched-budget RTN runs (\ref{app:framework-details}), outcome replay (\ref{app:errata-ablations}), runtime checks (\ref{app:control-checks}), autonomous agent (\ref{app:autonomous-agent}), $2\times2$ separation (\ref{app:reflectmem-checks}), 3 vendor proposers (\ref{app:second-proposer}), GPU-hour cost (\ref{app:gpu-cost}), and black-box controls (\ref{app:blackbox-controls}). \\
\end{tabular}\arrayrulecolor{black}
\end{tcolorbox}

\section{Extended Related Work, Glossary, and Notation}
\label{app:reference}

This section positions \QuantForge and \HiRes in detail against prior post-training quantization methods, LLM-driven program evolution frameworks, and automated scientific discovery agents (Table~\ref{tab:discovery-comparison}), followed by a concise glossary and unified notation table.

\subsection{Extended Related Work}
\label{sec:related}
\label{app:related}

\paragraph{Post-training quantization and strict MXFP4 W4A4.}
Post-training quantization reduces inference memory and compute without retraining via four main mechanisms. First, second-order weight reconstruction compensates rounding errors through layer curvature, as in GPTQ \citep{frantar2023gptq}, blockwise reconstruction (BRECQ; \citealp{li2021brecq}), and asymmetric-calibration GPTAQ \citep{li2025gptaq}. Second, coordinate transforms suppress outliers before quantization through diagonal scaling (SmoothQuant, AWQ, OmniQuant; \citealp{xiao2023smoothquant,lin2024awq,shao2024omniquant}), global or blockwise orthogonal rotations (QuaRot, SpinQuant, DuQuant, DuQuant++; \citealp{ashkboos2024quarot,liu2025spinquant,lin2024duquant,hsu2026duquantpp}), activation-error rotation refinement (DFRot; \citealp{xiang2025dfrot}), or learned affine maps (FlatQuant; \citealp{sun2025flatquant}). Under the OCP MXFP4 standard \citep{ocp2023mxspec,rouhani2023microscaling}, shared 32-element power-of-two exponents (E8M0) invalidate global smoothing assumptions; specialized MXFP4/NVFP4 methods therefore introduce block-aligned rotations, dual-operand second-moment charts, or search over block transforms (MR-GPTQ, BRQ, TORQ, BATQuant, WUSH, FOCUS; \citealp{cook2025mrgptq,shao2025brq,xu2026torq,li2026batquant,chen2026wush,yan2026focus}; see \citealp{zhang2026mxfpbenchmark} for a recent benchmark). Third, rounding-decision methods move beyond nearest-code projection via adaptive weight rounding (AdaRound, SignRoundV2; \citealp{nagel2020adaround,cheng2026signroundv2}), vector quantization (QTIP; \citealp{tseng2024qtip}), or gradient-learned online activation thresholds (AQuant; \citealp{li2022aquant}). Fourth, post-quantization recovery corrects residual bias or layer drift after quantized operators are installed (DFQ, Norm Tweaking, D$^2$Quant; \citealp{nagel2019dfq,li2024normtweaking,yan2026d2quant}). Earlier mixed-precision search methods (HAQ, HAWQ-V2; \citealp{wang2019haq,shen2020hawqv2}) allocate bitwidths across layers rather than synthesizing new operators inside a fixed W4A4 format. \HiRes operates strictly within legal Block32 E2M1/E8M0 W4A4 and unites closed-form dual-operand encoding-error correction ($K$), parameter-free installed-W4-aware one-code A4 refinement, and sequentially remeasured attention-to-SwiGLU recovery.

\paragraph{LLM-driven program evolution.}
FunSearch \citep{romeraparedes2024funsearch} and AlphaEvolve \citep{novikov2025alphaevolve} demonstrated that coupling an LLM proposer with an automated evaluator and an evolutionary program archive can discover mathematical constructions and systems algorithms beyond manual heuristics (extending earlier RL-based kernel discovery such as AlphaDev; \citealp{mankowitz2023alphadev}). Subsequent systems refine the evolutionary controller or prompt memory: Evolution of Heuristics (EoH; \citealp{liu2024eoh}) co-evolves natural-language thoughts with code; ReEvo \citep{ye2024reevo} and HSEvo \citep{dat2025hsevo} add verbal reflection and diversity control over elite programs; LLaMEA \citep{vanstein2025llamea} and MCTS-AHD \citep{zheng2025mctsahd} apply evolutionary strategies and Monte Carlo tree search; GEPA \citep{gepa2026} optimizes prompts through reflective feedback; AutoEP \citep{autoep2026} evolves metaheuristic hyperparameters; and self-improving coding agents such as the Darwin G\"odel Machine \citep{zhang2025dgm} and Huxley--G\"odel Machine \citep{huxley2026} evolve the agent harness itself. Across this entire family, however, the evaluator acts as a \emph{black box returning a scalar fitness score} (and optional textual logs). That loop excels when evaluations are fast, the score directly measures the ground-truth objective, and the search modifies one isolated routine inside a fixed skeleton. In strict MXFP4 W4A4 PTQ, all three assumptions break: each evaluation quantizes a full LLM, end-to-end perplexity aliases multiple coupled physical causes (weight--activation recoil, discrete E8M0 flips, stale calibration statistics, or calibration overfitting), and the right multi-stage decomposition is unknown beforehand. As our matched-budget comparisons confirm (Section~\ref{sec:discovery-results}), scalar evolution and reflective memory spend their limited budget repeatedly revisiting refuted shortcuts.

\paragraph{Hypothesis-driven and scientific discovery agents.}
A growing body of work brings explicit scientific reasoning to LLM agents. General automated research pipelines---including The AI Scientist \citep{lu2024aiscientist}, AI Scientist-v2 \citep{yamada2025aiscientistv2}, AI Co-Scientist \citep{gottweis2025aicoscientist}, Agent Laboratory \citep{schmidgall2025agentlaboratory}, and Kosmos \citep{mitchener2025kosmos}---propose ideas, run code, and synthesize written reports, while formal hypothesis-testing frameworks select informative experiments by Bayesian surprise (AutoDiscovery; \citealp{agarwal2025autodiscovery}) or sequential falsification tests (Popper; \citealp{huang2025popper}), and diagnostic-guided synthesis refines probabilistic programs from runtime statistics (RefineStat; \citealp{refinestat2026}). Closer to algorithm design, OR-Agent \citep{liu2026oragent} maintains a hypothesis tree with written reflections, CliffSearch \citep{mroueh2026cliffsearch} pairs candidate programs with evolving textual theories, and OPTScientist \citep{li2026optscientist} searches over typed optimizer expressions with reusable macros. \QuantForge shares the principle that search state should track \emph{why} a change worked, but closes two gaps left open by prior agents (Table~\ref{tab:discovery-comparison}). First, rather than testing hypotheses only through expensive end-to-end reruns or free-form prose, \QuantForge exploits PTQ's \emph{paired BF16--MXFP4 execution twin} to run minimum-cost local counterfactual controls on locked pre-registered predictions. Second, rather than storing conclusions as advisory text that an LLM may silently ignore in later rounds, \QuantForge enforces every test outcome as a required code update verified by a deterministic runtime check (rejecting no-op edits) while keeping strong quantizers in the archive independently of refuted explanations.

\paragraph{Comparison with surrogate evaluators and open-ended search.}
When end-to-end evaluation is costly, an alternative strategy is to accelerate search with learned or evolved surrogate evaluators. \QuantForge takes a structurally different route tailored to post-training quantization: it never approximates or relaxes the strict-MXFP4 benchmark. Instead, because the full-precision BF16 model is already present at zero extra modeling cost alongside every candidate quantized model, local twin controls supply exact, physically grounded causal diagnostics on small calibration fixtures while keeping the final evaluation target fixed and untouched.

\begin{table}[htbp]
\centering
\DeclareRobustCommand{\cyes}{\tikz[baseline=-0.55ex]\fill[qfTeal] (0,0) circle (2.7pt);}
\DeclareRobustCommand{\chalf}{\tikz[baseline=-0.55ex]{\draw[qfBronze, line width=0.7pt] (0,0) circle (2.4pt); \fill[qfBronze] (0,-2.4pt) arc (-90:90:2.4pt) -- cycle;}}
\DeclareRobustCommand{\cno}{\tikz[baseline=-0.55ex]\draw[qfGrayBd, line width=0.75pt] (0,0) circle (2.4pt);}
\caption{\textbf{Structural comparison of \QuantForge with LLM-driven program evolution and scientific discovery agents.} Columns contrast six architectural properties critical for coupled PTQ algorithm discovery: evolving executable code, exploiting a paired full-precision execution twin for layerwise residual diagnostics, selecting minimum-cost local counterfactual controls on pre-registered rival explanations, enforcing test outcomes via runtime checks on subsequent code, decoupling quantizer archive retention from hypothesis refutation, and sequentially remeasuring fresh residuals after each installed stage. \cyes~built in and enforced; \chalf~partial (advisory text reflection, fixed-skeleton search, or open-loop testing without enforced code checks); \cno~absent.}
\label{tab:discovery-comparison}
\scriptsize
\renewcommand{\arraystretch}{1.24}
\setlength{\tabcolsep}{2.8pt}
\begin{tabular}{@{}lccccccp{0.245\linewidth}@{}}
\toprule
\textbf{System} & \shortstack{Evolves\\code} & \shortstack{Paired\\BF16 twin} & \shortstack{Min-cost\\controls} & \shortstack{Runtime\\code check} & \shortstack{Decoupled\\archive} & \shortstack{Stagewise\\remeasure} & \textbf{State carried across rounds} \\
\midrule
\rowcolor{qfBlueBg}\multicolumn{8}{@{}l}{\sffamily\bfseries\scriptsize\color{qfBlue}LLM-Driven Program Evolution} \\
FunSearch \citep{romeraparedes2024funsearch} & \cyes & \cno & \cno & \cno & \cno & \cno & Scored programs in island archive \\
AlphaEvolve \citep{novikov2025alphaevolve} & \cyes & \cno & \cno & \cno & \cno & \cno & Scored programs \& prompt templates \\
EoH \citep{liu2024eoh} & \cyes & \cno & \cno & \cno & \cno & \cno & Programs paired with thought summaries \\
ReEvo \citep{ye2024reevo} & \cyes & \cno & \cno & \cno & \cno & \cno & Programs \& short/long verbal reflections \\
LLaMEA \citep{vanstein2025llamea} & \cyes & \cno & \cno & \cno & \cno & \cno & Programs, fitness scores, \& error traces \\
MCTS-AHD \citep{zheng2025mctsahd} & \cyes & \cno & \cno & \cno & \cno & \cno & MCTS tree of scored heuristic code \\
GEPA \citep{gepa2026} & \chalf & \cno & \cno & \cno & \cno & \cno & Reflective prompt/text lineages \\
\midrule
\rowcolor{qfBronzeBg}\multicolumn{8}{@{}l}{\sffamily\bfseries\scriptsize\color{qfBronze}Hypothesis-Driven \& Scientific Discovery Agents} \\
AI Scientist / v2 \citep{lu2024aiscientist,yamada2025aiscientistv2} & \cyes & \cno & \chalf & \cno & \cno & \cno & Experimental scripts, logs, \& manuscripts \\
AutoDiscovery \citep{agarwal2025autodiscovery} & \chalf & \cno & \chalf & \cno & \cno & \cno & Bayesian surprise hypothesis tree \\
Popper \citep{huang2025popper} & \chalf & \cno & \cyes & \cno & \cno & \cno & Sequential hypothesis falsification tests \\
OR-Agent \citep{liu2026oragent} & \cyes & \cno & \chalf & \cno & \cno & \cno & Hypothesis tree \& written reflections \\
CliffSearch \citep{mroueh2026cliffsearch} & \cyes & \cno & \chalf & \cno & \cno & \cno & Programs paired with written theories \\
OPTScientist \citep{li2026optscientist} & \cyes & \cno & \chalf & \cno & \cno & \cno & Typed optimizer programs, macros, notes \\
\midrule
\rowcolor{qfTealBg}\textbf{\QuantForge (Ours)} & \cyes & \cyes & \cyes & \cyes & \cyes & \cyes & \textbf{MXFP4 quantizer archive $\mathcal{Q}_t$, locked twin-control outcomes, \& runtime-checked edits} \\
\bottomrule
\end{tabular}
\end{table}

\subsection{Glossary}
\label{app:glossary}

Table~\ref{tab:glossary} summarizes the core concepts of \QuantForge in plain quantization terms and maps them to compact shorthand labels used in the detailed production-run tables of Appendix~\ref{app:controller}.

\begin{table}[htbp]
\centering\revcolor
\caption{Core terms used for \QuantForge, their quantization instances, and compact appendix shorthand where applicable.}
\label{tab:glossary}
\small\setlength{\tabcolsep}{4pt}
\begin{tabular}{@{}p{.22\linewidth}p{.43\linewidth}p{.31\linewidth}@{}}
\toprule
Term & Meaning & Instance in this paper \\
\midrule
Quantizer / Program & Executable code mapping a BF16 model to a strict-MXFP4 W4A4 model under the fixed evaluation setup & Strict-MXFP4 GPTQ seed (Figure~\ref{fig:errata-framework}); final \HiRes \\
Candidate \& Parent & New quantizer proposed in one round (rebuilt from fresh BF16 weights) and the archived parent it modifies & Round~44 post-encoding affine recovery modifying \#41 \\
Fixed setup (App.: \emph{contract}) & Shared format, target modules, calibration set, evaluator, and budgets held constant across all methods (\S\ref{sec:prelim-format}) & E2M1/E8M0 Block32 W4A4; C4 $128\times2048$ \\
BF16--MXFP4 twin & Layer-aligned pair $(M^{(k)}, M^{(0)})$ running on identical calibration tokens to expose exact residuals $R_k$ & BF16 value output $\mathbf{v}_t^\star$ vs.\ quantized $\mathbf{v}_t$ \\
Open question (App.: \emph{residual} $\rho_t$) & Specific error-attribution question attached to a quantizer whose rival explanations imply different edits (\S\ref{sec:lakatos}) & Must attention recovery be fitted after A4 encoding? \\
Local twin control ($\sigma$) & Counterfactual toggle on $(M^{(k)}, M^{(0)})$ changing one quantization dependency at equal capacity (\S\ref{sec:compiler}) & Same two per-head scalars fitted before vs.\ after A4 \\
Test outcome (App.: \emph{verdict} $v_t$) & Causal classification of a controlled run: \emph{supported}, \emph{proxy-only} (App.: \emph{mediator-only}), \emph{metric-only} (App.: \emph{target-only}), \emph{no effect}, \emph{unresolved}, or \emph{unsafe} & Proxy-only outcome in Round~43 barring pre-A4 fits \\
Outcome-guided update (App.: \emph{compilation}) & Translating a causal test outcome into a required code edit for the successor quantizer (\S\ref{sec:lineage}) & Collect value outputs on the installed W4A4 path \\
Runtime check (App.: \emph{execution probe}) & Targeted test verifying that toggling the declared dependency changes runtime output, rejecting no-op edits & With $E_X$ fixed, changing weight energy shifts the activation-side update (Round~14) \\
Archive / Hypothesis queue & Score-retained quantizers ($\mathcal{Q}_t$) / open attribution questions ($\mathcal{H}_t$), kept separate so strong code survives refuted hypotheses & Figure~\ref{fig:errata-framework} (top and bottom tracks) \\
Transfer gate & Production-run validation on Llama-3.1-8B returning only aggregate acceptance & All 6 metric-only shortcuts rejected (Table~\ref{tab:verdict-gate}) \\
Evaluator call & One recorded build/evaluation of a candidate, local control, runtime check, entry check, or transfer check & 240 calls per matched-budget run; 174 in production \\
Proposer / controller / coding agent & LLM making algorithmic judgments / deterministic code enforcing locks, legality, and checks / coding harness implementing code (App.~\ref{app:search-cycle}) & GPT-5.6 Sol / \QuantForge controller / Codex with GPT-5.6 Sol \\
\bottomrule
\end{tabular}
\end{table}

\subsection{Notation}
\label{app:notation}

Table~\ref{tab:notation} lists the primary symbols across the main text and appendices. Where detailed appendix derivations use compact shorthand from the production logs, the correspondence is listed inline: archive $\mathcal{Q}_t$ ($\mathcal{P}_t$ in App.~\ref{app:controller}, with $P_t, P_{t+1}$ matching $q_t, q_{t+1}$); hypothesis queue $\mathcal{H}_t$ ($\mathcal{R}_t$); local controls $\sigma \in \Sigma$ with predictions $\hat{y}_e(\sigma)$ ($c \in \mathcal{U}$, $P_e(c)$); and reference chart and Hadamard $\Pi_{\mathrm{ref}}, H_{32}$ ($P_{\mathrm{ref}}, R_H$ in App.~\ref{app:hires-details}). Shared letters are disambiguated locally: $D$ is diagonal balance in \S\ref{sec:hiresmx} (and a local design matrix in App.~\ref{app:hires-details}); $\mathcal{C}$ is the fixed setup while $C_X, C_E$ are second moments; $v_t$ is the round outcome in Sec.~\ref{sec:lineage} while $\mathbf{v}_t$ is an attention value vector in Eq.~\ref{eq:hires-vo-fit}; $u$ is a balance exponent or up-projection output; and $U=DT$ is the folded transform in Apps.~\ref{app:hires-details} and~\ref{app:computational-cost}. Production-run programs are labelled \#00--\#56.

\begin{table}[htbp]
\centering\revcolor
\caption{Unified notation across the main text and appendices, with the section or equation introducing each symbol.}
\label{tab:notation}
\small\setlength{\tabcolsep}{4pt}
\begin{tabular}{@{}p{.24\linewidth}p{.57\linewidth}p{.14\linewidth}@{}}
\toprule
Symbol & Meaning & Introduced \\
\midrule
\rowcolor{qfBlueBg}\multicolumn{3}{@{}l}{\sffamily\bfseries\scriptsize\color{qfBlue}Discovery (\QuantForge)} \\
$\mathcal{Q}_t$ (App.: $\mathcal{P}_t$) & Quantizer archive: score-retained programs at round $t$ & Sec.~\ref{sec:lakatos} \\
$q_t, q_{t+1}$ (App.: $P_t, P_{t+1}$) & Parent quantizer selected at round $t$ and its successor & Sec.~\ref{sec:proposals} \\
$\mathcal{H}_t$ (App.: $\mathcal{R}_t$); $h_t$ ($\rho_t$) & Hypothesis queue of open questions; question targeted at round $t$ & Sec.~\ref{sec:lakatos} \\
$S_t=(\mathcal{Q}_t,\mathcal{H}_t)$ & Full search state at round $t$ & Sec.~\ref{sec:lakatos} \\
$\mathcal{E}$; $e,e'$ & Registered competing explanations; two distinct hypotheses & Eq.~\ref{eq:control-compiler} \\
$\sigma \in \Sigma$ (App.: $c \in \mathcal{U}$) & Local counterfactual twin control; allowed control family & Eq.~\ref{eq:control-compiler} \\
$\hat{y}_e(\sigma)$ (App.: $P_e(c)$) & Outcome predicted by explanation $e$ under control $\sigma$ & Eq.~\ref{eq:control-compiler} \\
$\Sigma^\ast$, $\mathrm{cost}(\sigma)$ & Minimum-cost discriminating control set; evaluation cost & Eq.~\ref{eq:control-compiler} \\
$v_t$; $\mathcal{C}$ & Causal test outcome at round $t$; fixed strict-MXFP4 setup & Sec.~\ref{sec:lineage}, Alg.~\ref{alg:errata} \\
$\mathcal{S}_{t+1}$ & Next intervention space opened after stagewise remeasurement & Eq.~\ref{eq:progressive-factorization} \\
\rowcolor{qfTealBg}\multicolumn{3}{@{}l}{\sffamily\bfseries\scriptsize\color{qfTeal}Quantizer (\HiRes)} \\
$M^{(0)}, M^{(k)}; R_k$ & BF16 reference twin, quantized state at stage $k$, actual residual & Sec.~\ref{sec:prelim}, Eq.~\ref{eq:conditional-stage} \\
$X,W$; $q=32$ & Activations and weights of one input block; Block32 width & Sec.~\ref{sec:geometric-shaping} \\
$A,B$; $\mathrm{DN}$, $\mathrm{Cap}_8$ & Regularized second moments; determinant normalization; spectral cap & Eq.~\ref{eq:hires-reference-chart} \\
$\Pi_{\mathrm{ref}}$ (App.: $P_{\mathrm{ref}}$) & Dual-geometry reference coordinate chart & Eq.~\ref{eq:hires-reference-chart} \\
$H_{32}$ (App.: $R_H$) & Normalized $32\times32$ Hadamard matrix & Eq.~\ref{eq:hires-anchor-cross} \\
$E_X,E_W$; $K_X,K_W$ & Anchor MXFP4 encoding errors; error--signal cross-moments & Eq.~\ref{eq:hires-anchor-cross} \\
$K$, $G$, $T$ & Interaction correction; normalized $I+K$; $T=H_{32}G\Pi_{\mathrm{ref}}$ & Eq.~\ref{eq:hires-transform} \\
$r_A,r_W$; $D$ & Subgroup encoding risks; dyadic balance $\mathrm{Diag}(2^s)$ (proposal $u_b$) & Eq.~\ref{eq:bilateral-risk} \\
$H$; $\widehat{W}$ & Fresh transformed-state GPTQ curvature; decoded W4 weights & Eq.~\ref{eq:recon} \\
$G_Q,K_Q$; $q_0,d_0$ & $\widehat{W}^\top\widehat{W}$, $\widehat{W}^\top(W-\widehat{W})$; nearest A4 code and error & Eq.~\ref{eq:hires-code-score} \\
$a,b$ & Post-encoding per-head affine coefficients for attention values & Eq.~\ref{eq:hires-vo-fit} \\
$\alpha,\beta,\eta$; $\sigma_g$ & SwiGLU gate shift, gate scale, up scale; gate standard deviation & Eq.~\ref{eq:hires-mlp-update} \\
\bottomrule
\end{tabular}
\end{table}

\raggedbottom
\section{\QuantForge: Search Protocol and Production Discovery Trace}
\label{app:controller}

This section documents how \QuantForge operates and traces the complete 56-round production discovery run that yielded \HiRes. Subsections~\ref{app:search-information}--\ref{app:search-cycle} specify the two discovery protocols, verdict-compilation rules, and three-role execution loop; Subsection~\ref{app:production-record} visualizes and tabulates all 56 production rounds; and Subsection~\ref{app:round-records} provides structured diagnostic cards for 23 pivotal rounds.

\subsection{Discovery Protocols}
\label{app:search-information}

\paragraph{Algorithm discovery.}
Algorithm development starts from {\revcolor a strict-MXFP4 GPTQ program
\citep{frantar2023gptq}, a standard weight-reconstruction starting point with
no coordinate transform, activation-code rule, or recovery path,} and retains
successful descendants as starting points for later revisions.
{\revcolor The whole program is editable. The transform, discrete
reconstruction, and residual recovery are not fixed by the initial
implementation: the seed contains none of them, and every operation of the
final \HiRes beyond GPTQ was introduced and revised by the search
(Appendix~\ref{app:production-record}).}
Each candidate is rebuilt from BF16 weights using the same strict-MXFP4
contract and $128\times2048$ calibration protocol.

Qwen3-4B \citep{yang2025qwen3} is the search-visible model. Its task results
and diagnostic measurements guide subsequent proposals. Llama-3.1-8B
\citep{grattafiori2024llama} serves as the transfer-validation model.
A visible-model improvement is accepted as transferable only when this
validation model also improves against the same comparison program.
The proposer receives the aggregate transfer acceptance signal; detailed
task, layer, and module diagnostics remain with the evaluator.

Llama-3.2-3B, Mistral-7B-v0.3, Qwen3-8B, Meta-Llama-3-8B, OLMo-2-13B, and Qwen3-32B are excluded
from proposal generation and candidate selection and are evaluated only
after the final algorithm is fixed. Component and stage ablations on
Qwen3-8B and Meta-Llama-3-8B then measure the contribution of its parts.
Their results do not feed back into discovery or component selection.

The search score combines WikiText-2 and C4 perplexity with ARC-Challenge
length-normalized accuracy (\texttt{acc\_norm})
\citep{merity2017wikitext,raffel2020t5,clark2018arc}.
For each model $m$, damage is measured against its own BF16 reference:
\begin{align}
 d_{m,\mathrm{WT2}}&=\frac{\mathrm{PPL}_{q,m,\mathrm{WT2}}}
 {\mathrm{PPL}_{16,m,\mathrm{WT2}}}-1,
 \qquad d_{m,\mathrm{C4}}=\frac{\mathrm{PPL}_{q,m,\mathrm{C4}}}
 {\mathrm{PPL}_{16,m,\mathrm{C4}}}-1,\\
 d_{m,\mathrm{ARC}}&=\frac{\mathrm{Acc}^{\rm norm}_{16,m}-
 \mathrm{Acc}^{\rm norm}_{q,m}}{\mathrm{Acc}^{\rm norm}_{16,m}},\\
 \text{Fit-3}_m&=\frac{d_{m,\mathrm{WT2}}+d_{m,\mathrm{C4}}+
 d_{m,\mathrm{ARC}}}{6}
 +\frac12\max_{j\in\{\mathrm{WT2,C4,ARC}\}}d_{mj}.
 \label{eq:fit3}
\end{align}
Lower is better. Equal weight on mean and worst-task damage rewards broad
improvements while penalizing a large loss on any one task. Damage is not
clipped at zero, so gains over BF16 retain their negative sign.

\paragraph{Independent framework comparison.}
\label{app:framework-search-information}

The matched-budget study in Section~\ref{sec:discovery-results} starts each
arm from the same strict-MXFP4 round-to-nearest (RTN) program.
The proposer revises this code{\revcolor, through the coding agent,} using Qwen3-4B scores and execution feedback;
\QuantForge additionally carries forward unresolved questions and controlled
experimental conclusions. Each accepted revision supplies the parent state
for later interventions. The editable program can introduce new operations,
so the final three-stage quantizer is not part of the initial RTN program.

Qwen3-4B is search-visible; Llama-3.1-8B is held out for independent
transfer assessment. Every 20 evaluator calls, a separate evaluator tests
the current champion on Llama. Neither its score nor its per-task,
per-layer, or per-module diagnostics are returned to the proposer or used
to choose descendants. This evaluator separation follows the holdout
principle \citep{dwork2015reusableholdout}; the recorded outcomes measure
transfer under the same search budget.

This controlled study uses WikiText-2 perplexity as its visible objective
and held-out endpoint. Its $\mathrm{PPL}\leq8$ success gate and 240-call
results describe this independent framework comparison. The
algorithm-development search above uses Fit-3; the quantizer benchmark uses
the seven-task panel.

\subsection{Controlled Experiments and Program Updates}

Each proposal specifies an executable candidate $p$, its parent $\pi$,
and the residual $\rho$ it aims to reduce. It also records a focal
explanation and its competitors in $\mathcal E$, their predicted outcomes
$P_e(c)$ under each condition $c$, and the allowed controls $\mathcal U$.

Before execution, {\revcolor the proposer selects controls by
Eq.~\ref{eq:control-compiler}, and the controller registers and freezes them}
together with $p$, $\pi$, $\rho$, $\mathcal E$, {\revcolor the predictions $P_e(c)$}, and the evaluator identity.  The primary run and its controls share this
packet.  A completed condition can be reused only when the full execution
identity matches.
If the allowed controls cannot distinguish the registered explanations, the
proposal is revised before model evaluation. Every condition shares the model,
calibration data, quantizer format, evaluator, and resource ceiling; only the
declared intervention changes.

The observed results determine both the status of the explanation and the
required response in the next program:
\begin{center}
\small
\setlength{\tabcolsep}{6pt}
\renewcommand{\arraystretch}{1.22}
\begin{tabular}{lll}
  \toprule
  Verdict & Residual status & Required program response \\
  \midrule
  \vSup & closed under the focal account & extend or compose the direction \\
  \vTar & gain lacks the predicted signature & repair the explanation \\
  \vMed & signature lacks the target gain & repair transfer to the target \\
  \vNef & intervention is parent-equivalent & retire the intervention \\
  \vUnr & current controls do not separate accounts & add a discriminator \\
  \vUns & rollback or contract check fails & reject and restore the parent \\
  \bottomrule
\end{tabular}
\end{center}
Program retention is evaluated separately.  A target-only candidate, for
example, may remain useful as executable code even though its focal account
cannot guide a descendant.

For every response-bearing descendant, {\revcolor the proposer fixes the
probe target and its pass criterion, and the controller records pass or fail
from the measured probe output against that criterion}.  A source-level no-op
does not count as a residual closure.  The check is scoped to the computation
named by $\rho$; it does not require unrelated code to change.

\subsection{Search Cycle}
\label{app:search-cycle}

The search maintains executable programs and unresolved residuals as
separate frontiers. {\revcolor The proposer chooses parents from the
executable frontier, the controller schedules experiments, and residual
compilation, which maps each verdict to the required response of
Appendix~\ref{app:controller},} determines which conclusions can guide a program
update. Population size, mutation operators, and scheduling can vary while retaining this interface; the production run retains the full archive, whereas the matched-budget study fixes single-lineage updates across all arms (Appendices~\ref{app:framework-details}, \ref{app:reflectmem-checks}).

{\revcolor
\paragraph{Roles.} A round involves three roles. The \emph{proposer} model,
GPT-5.6 Sol in the production run and in every comparison, joined by two
further proposers in Appendix~\ref{app:second-proposer}, makes every judgment within the search: it proposes algorithms, formulas, competing
explanations, and controls, fixes what the successor must implement and what
its probe must show, interprets the results, and issues the verdict. The
\emph{controller}, part of the \QuantForge framework code and separate from the coding agent, applies fixed rules: it maintains the experiment state and
budget, registers and freezes each packet the proposer submits, schedules the
evaluations, records probe pass or fail from measured outputs against the
proposer's criterion, and applies the acceptance, promotion, and rollback
rules. A \emph{coding agent} implements the proposer's specifications as
candidate code and probes, runs them, and collects the outputs; any change that
touches a formula or an experimental decision goes back to the proposer. The
coding agent can be any existing harness, such as Codex or Claude Code, and
the protocol refers to no particular one; our runs use Codex with GPT-5.6 Sol, and no
algorithm code was written by hand. All arms in
Appendix~\ref{app:discovery-experiments}, including the baselines and the
autonomous agent, share the same coding agent; evaluator budgets count
evaluator calls, and the coding agent's model calls are accounted separately. Because each packet,
with its predictions, is frozen before any of its conditions runs, predictions
cannot be fitted to that experiment's results. In the matched-budget
comparisons, held-out results never return to the proposer; in the production
run, the proposer receives only the aggregate transfer signal
(Appendix~\ref{app:search-information}). In the main text, phrases such as
``\QuantForge selects'' refer to the proposer acting under this protocol.}

After compilation, the revised program is remeasured and exposes the next
intervention set:
\begin{equation}
  \mathcal R(P_{t+1})=\mathsf{Measure}(P_{t+1}),
  \qquad
  \mathcal S_{t+1}=\mathsf{Open}(P_{t+1},\mathcal R(P_{t+1})).
  \label{eq:progressive-factorization}
\end{equation}
Here $\mathcal R(P_{t+1})$ denotes the residuals remeasured on this program,
not a replacement of the full residual frontier. Unresolved questions
attached to other retained programs remain available.
$\mathcal S_{t+1}$ contains interventions made actionable by residuals
measured on $P_{t+1}$. It can include existing operations or newly introduced
components. The hierarchy records both the operation and the earlier program
state that made it relevant.

The history stores completed experiments, Pareto donor programs, duplicate
proposals, and stagnation information alongside the two frontiers. These
records let the proposer reuse successful computations and distinguish a
new condition from a repeated intervention.

\begin{algorithm}[H]
  \caption{One \QuantForge round}
  \label{alg:errata}
  \begin{algorithmic}[1]
    \STATE \textbf{input:} executable frontier $\mathcal P_t$, residual
           frontier $\mathcal R_t$, frozen contract $\mathcal C$
    \STATE select a program and an actionable residual
    \STATE propose executable interventions with competing explanations
    \STATE {\revcolor the proposer selects} minimum-cost discriminating controls
           (Eq.~\ref{eq:control-compiler})
    \STATE freeze and execute the experiment packet under $\mathcal C$
    \STATE retain programs by measured performance
    \STATE resolve the residual and check the required response {\revcolor with an execution probe}
    \STATE remeasure residuals on the changed program; open the next
           intervention space
  \end{algorithmic}
\end{algorithm}

{\revcolor
\subsection{Production Discovery Record}
\label{app:production-record}

The production run that produced \HiRes follows the algorithm-discovery
protocol of Appendix~\ref{app:search-information}. It starts from
strict-MXFP4 GPTQ, scores candidates by Fit-3 on Qwen3-4B, and accepts a
visible improvement as transferable only after the Llama-3.1-8B gate. The
proposer had web access, received no baseline code, and was given no prompt
guidance toward existing methods or papers. A coding agent implemented and ran
every round under the controller's rules (Appendix~\ref{app:search-cycle});
the authors provided the environment and checked the results.
The run took 56 research rounds and 174 recorded calls: 46 complete primary
evaluations and 36 transfer-gate evaluations, plus 59 control conditions, 30
execution probes, and 3 entry checks run on local calibration fixtures. Twenty-nine rounds promoted a program and seven
reused an existing one. Round~56 froze \HiRes; no component was added or
changed afterwards. Of the 29 promotions, 21 implemented a code response
required by an earlier verdict. Tables~\ref{tab:production-rounds}
and~\ref{tab:production-rounds-b} list every round,
Appendix~\ref{app:round-records} reproduces the full structured cards of 23 selected rounds, and
Appendix~\ref{app:hires-provenance} dates each component. Figure~\ref{fig:production-trajectory} visualizes the full 56-round Fit-3 trajectory, verdict distribution, and transfer-gate filtering across all three stages of \HiRes.

\begin{figure}[htbp]
  \centering
  \includegraphics[width=\textwidth]{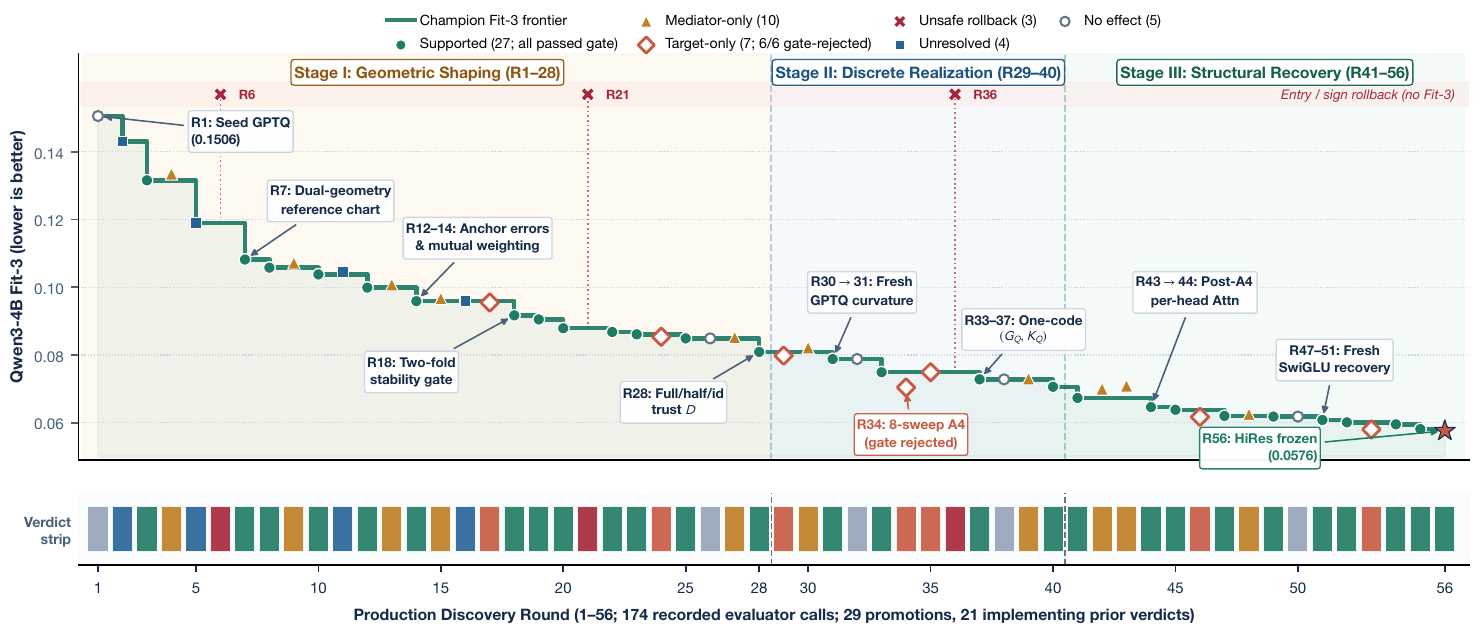}
  \caption{\textbf{Full 56-round production discovery trajectory on Qwen3-4B (174 recorded evaluator calls).} \textbf{Top:} Champion Fit-3 frontier (teal step curve) descending from the strict-MXFP4 GPTQ seed ($0.1506$, Round~1) to frozen \HiRes ($0.0576$, Round~56) across three emergent stages: Geometric Shaping (R1--28), Discrete Realization (R29--40), and Structural Recovery (R41--56). Hollow coral diamonds mark Target-only branches whose visible-model gains were rejected by the Llama-3.1-8B transfer gate (6/6 rejected); amber triangles mark Mediator-only diagnostics that triggered subsequent code responses; crimson crosses at top mark entry/sign safety rollbacks (R6, R21, R36). \textbf{Bottom:} Color-coded per-round verdict strip aligned across Rounds~1--56.}
  \label{fig:production-trajectory}
\end{figure}

Two rounds show how a verdict changes the next program. In round~13, the
interaction correction was updated from activation errors alone. The control
measured both operands: activation error fell by 5.6\%, weight error rose by
6.8\%, and the joint error rose by 0.7\%. The mediator-only verdict ruled out
accepting an update on activation error alone, and round~14 weighted each
operand's error by the energy of the other, lowering output error by 3.7\%
against 0.8\% for a parameter-matched block-average control. In round~43,
attention recovery was fitted before A4 encoding. With the same two
coefficients per head, the pre-encoding fit lowered local MSE by 0.02\% and
the post-encoding fit by 7.1\%; the verdict required fitting on the installed
W4/A4 path, which became Eq.~\ref{eq:hires-vo-fit}.

\paragraph{Verdicts and the transfer gate.}
Verdicts rest on controls run on Qwen3-4B calibration fixtures; the transfer gate evaluates on
Llama-3.1-8B and returns an aggregate acceptance signal to the proposer.
Table~\ref{tab:verdict-gate} cross-tabulates the two. All 27 supported changes
passed the gate. None of the six target-only changes that reached it passed,
although each improved Fit-3 over its parent on Qwen3-4B; the gate thus kept
\HiRes from changes that helped only the search-visible model.
The gate signal also feeds the later search, and the record keeps no time stamp
for each verdict. We therefore read this agreement as retrospective
consistency. The independent test of the design choices they shaped is the post-discovery
ablations on models that never entered the search
(Appendix~\ref{app:experiments}). Round~33 shows how a supported change keeps
being refined after it passes: its verdict, that A4 code choice must read the
installed W4, carries into \HiRes; round~35 found that its $G_Q$-only score
omits the weight-mismatch term, round~36 rejected an implementation with the
wrong sign, and round~37 added the term with the sign fixed by the algebra. Without that term the rule
trails nearest rounding in the final pipeline
(Table~\ref{tab:onecode-rules}).

\begin{table}[htbp]
\centering\revcolor
\caption{Verdicts on Qwen3-4B and the Llama-3.1-8B transfer gate in the
production run. The gate runs only for visible improvements; one program that passed (round 11) was not promoted, so 30 passes yield 29 promotions.}
\label{tab:verdict-gate}
\small\setlength{\tabcolsep}{6pt}
\begin{tabular}{lrrr}
\toprule
Verdict & Rounds & Reached the gate & Passed \\
\midrule
Supported & 27 & 27 & 27 \\
Unresolved & 4 & 3 & 3 \\
Target-only & 7 & 6 & 0 \\
Mediator-only & 10 & 0 & -- \\
No effect & 5 & 0 & -- \\
Unsafe & 3 & 0 & -- \\
\midrule
Total & 56 & 36 & 30 \\
\bottomrule
\end{tabular}
\end{table}

\begin{table}[htbp]
\centering\revcolor
\caption{All 56 rounds of the production run (rounds 1--28: Stage~I Geometric Shaping). Fit-3 is
measured on Qwen3-4B (bold marks champion promotions); ``--'' marks rounds rejected at the entry or contract
check. Transfer: Llama-3.1-8B gate outcome; ``--'' means not triggered. A
program can pass the gate and stay in the executable frontier without being
promoted (round 11). Resp.: the earlier round whose verdict the code change
implements; ``--'' means none.}
\label{tab:production-rounds}
\scriptsize\setlength{\tabcolsep}{4pt}
\renewcommand{\arraystretch}{1.02}
\begin{tabular}{rp{.39\linewidth}rllll}
\toprule
Rd. & Intervention & Fit-3 & Verdict & Transfer & Champion & Resp. \\
\midrule
\rowcolor{qfBronzeBg}\multicolumn{7}{l}{\sffamily\bfseries\color{qfBronze}Stage I: Geometric Shaping (Rounds 1--28)} \\
\midrule
1 & Seed interface and encoding-boundary check & 0.1506 & \vNef & seed & \#00 & -- \\
2 & One-sided diagonal equalization & \textbf{0.1431} & \vUnr & accepted & \#02 & -- \\
3 & Fixed orthogonal mixing & \textbf{0.1316} & \vSup & accepted & \#03 & -- \\
4 & Stronger peak clipping & 0.1335 & \vMed & -- & \#03 & -- \\
5 & Activation-covariance whitening & \textbf{0.1190} & \vUnr & accepted & \#05 & R4 \\
6 & One-sided whitening on ill-conditioned blocks & -- & \vUns & -- & \#05 & -- \\
7 & Dual-geometry reference chart & \textbf{0.1082} & \vSup & accepted & \#07 & -- \\
8 & Volume and spectral bounds & \textbf{0.1058} & \vSup & accepted & \#08 & R6 \\
9 & Larger spectral freedom & 0.1072 & \vMed & -- & \#08 & -- \\
10 & Block reordering and exponent discrimination & \textbf{0.1038} & \vSup & accepted & \#10 & R9 \\
11 & Return to fixed physical blocks & 0.1046 & \vUnr & accepted & \#10 & -- \\
12 & Cross-moments from an actual encoding & \textbf{0.0999} & \vSup & accepted & \#12 & -- \\
13 & Activation-error-only update & 0.1008 & \vMed & -- & \#12 & -- \\
14 & Mutual weighting of both operands & \textbf{0.0959} & \vSup & accepted & \#14 & R13 \\
15 & Strictly orthogonal rotation branch & 0.0967 & \vMed & -- & \#14 & -- \\
16 & Pair-mode diagnostic revisit & 0.0959 & \vUnr & -- & \#14 & -- \\
17 & Full-matrix update without thresholds & 0.0955 & \vTar & \textcolor{qfCoral}{rejected} & \#14 & -- \\
18 & Two-fold stability gate & \textbf{0.0917} & \vSup & accepted & \#18 & R17 \\
19 & Separating owner conflicts & \textbf{0.0905} & \vSup & accepted & \#19 & R18 \\
20 & Directed-load constraint & \textbf{0.0879} & \vSup & accepted & \#20 & R19 \\
21 & Independent radius per pair & -- & \vUns & -- & \#20 & -- \\
22 & Joint small-ball closed-form correction & \textbf{0.0868} & \vSup & accepted & \#22 & R21 \\
23 & Volume normalization and zero diagonal & \textbf{0.0861} & \vSup & accepted & \#23 & R22 \\
24 & Repeated optimization with a stronger anchor & 0.0854 & \vTar & \textcolor{qfCoral}{rejected} & \#23 & -- \\
25 & Mutual imbalance of encoding risk & \textbf{0.0849} & \vSup & accepted & \#25 & -- \\
26 & Scale recheck of the zero-sum projection & 0.0849 & \vNef & -- & \#25 & -- \\
27 & Accepting $D$ by risk ratio only & 0.0852 & \vMed & -- & \#25 & -- \\
28 & Full/half/identity trust comparison & \textbf{0.0809} & \vSup & accepted & \#28 & R27 \\
\bottomrule
\end{tabular}
\end{table}

\begin{table}[htbp]
\centering\revcolor
\caption{All 56 rounds of the production run (rounds 29--56: Stage~II Discrete Realization \& Stage~III Structural Recovery; continued).}
\label{tab:production-rounds-b}
\scriptsize\setlength{\tabcolsep}{4pt}
\renewcommand{\arraystretch}{1.02}
\begin{tabular}{rp{.39\linewidth}rllll}
\toprule
Rd. & Intervention & Fit-3 & Verdict & Transfer & Champion & Resp. \\
\midrule
\rowcolor{qfBlueBg}\multicolumn{7}{l}{\sffamily\bfseries\color{qfBlue}Stage II: Discrete Realization (Rounds 29--40)} \\
\midrule
29 & Per-layer strength of $D$ & 0.0798 & \vTar & \textcolor{qfCoral}{rejected} & \#28 & -- \\
30 & Further curvature-cache shortcut & 0.0822 & \vMed & -- & \#28 & -- \\
31 & Fresh pre-A4 curvature after $DT$ & \textbf{0.0788} & \vSup & accepted & \#31 & R30 \\
32 & Recheck of block identity and access order & 0.0788 & \vNef & -- & \#31 & -- \\
33 & Installed-weight-guided A4 & \textbf{0.0749} & \vSup & accepted & \#33 & -- \\
34 & Repeated full coordinate sweeps & 0.0704 & \vTar & \textcolor{qfCoral}{rejected} & \#33 & -- \\
35 & Residual revisit of the $G_Q$-only program & 0.0749 & \vTar & -- & \#33 & -- \\
36 & Sign check of the joint residual expansion & -- & \vUns & -- & \#33 & -- \\
37 & Complete one-code output objective & \textbf{0.0728} & \vSup & accepted & \#37 & R36 \\
38 & Candidate enumeration and tie-break check & 0.0728 & \vNef & -- & \#37 & -- \\
39 & Recheck of the one-code information boundary & 0.0730 & \vMed & -- & \#37 & -- \\
40 & $D$ acceptance on the installed one-code rule & \textbf{0.0706} & \vSup & accepted & \#40 & R28 \\
\midrule
\rowcolor{qfTealBg}\multicolumn{7}{l}{\sffamily\bfseries\color{qfTeal}Stage III: Structural Recovery (Rounds 41--56)} \\
\midrule
41 & Residual shifts from operators to paths & \textbf{0.0673} & \vSup & accepted & \#41 & -- \\
42 & Layer-wide attention scale only & 0.0700 & \vMed & -- & \#41 & -- \\
43 & Attention fitted before encoding & 0.0709 & \vMed & -- & \#41 & R42 \\
44 & Post-encoding per-head affine & \textbf{0.0646} & \vSup & accepted & \#44 & R43 \\
45 & Product coupling in the MLP residual & \textbf{0.0638} & \vSup & accepted & \#45 & -- \\
46 & Affine on the final MLP product & 0.0617 & \vTar & \textcolor{qfCoral}{rejected} & \#45 & -- \\
47 & Gate shift and two branch scales & \textbf{0.0620} & \vSup & accepted & \#47 & R46 \\
48 & BF16-magnitude parameterization control & 0.0625 & \vMed & -- & \#47 & -- \\
49 & Regularized realized gate magnitude & \textbf{0.0618} & \vSup & accepted & \#49 & R48 \\
50 & Stale MLP fit before attention & 0.0618 & \vNef & -- & \#49 & -- \\
51 & Fresh MLP fit after attention & \textbf{0.0608} & \vSup & accepted & \#51 & R50 \\
52 & Down-projection-weighted objective & \textbf{0.0601} & \vSup & accepted & \#52 & R51 \\
53 & Ill-conditioned directions and ridge & 0.0580 & \vTar & \textcolor{qfCoral}{rejected} & \#52 & -- \\
54 & Robust per-cell estimation & \textbf{0.0595} & \vSup & accepted & \#54 & R53 \\
55 & Scale adaptation and deterministic aggregation & \textbf{0.0581} & \vSup & accepted & \#55 & R54 \\
56 & Uncertainty shrinkage; \HiRes frozen & \textbf{0.0576} & \vSup & accepted & \#56 & R55 \\
\bottomrule
\end{tabular}
\end{table}
}

{\revcolor
\subsection{Round Records of the Production Run}
\label{app:round-records}

This subsection reproduces the records of 23 rounds selected from the
production run: rounds that shaped the stages of the final \HiRes, a round
rejected by the transfer gate, and the round that froze the algorithm. Each record gives the program and its
parent, the observation that opened the round, the registered explanations
with the focal one marked, the executed control and its outcome, the task
result, the verdict with the required response, the implementation check,
and the transfer outcome. Fit-3 is measured on Qwen3-4B, and $\Delta$ is the
change from the parent. Programs are labelled \#00--\#56. The fields follow the
record's layout rather than the order in which they were completed. Tables~\ref{tab:production-rounds}
and~\ref{tab:production-rounds-b} summarize all 56 rounds.

\stagebanner{qfBronzeBg}{qfBronze}{Stage I: Geometric Shaping (Selected Rounds 6--28)}{Dual-geometry reference chart, anchor error moments, two-fold stability gate, and discrete trust-region balance $D$}
\begin{roundcard}{qfCrimsonBg}{qfCrimsonBd}{qfCrimson}{Round 6: one-sided whitening on ill-conditioned blocks}{(\#06 from \#05; calls 15--17)}{\fbadge{Fit-3: -- (halted)}}{\vUns\,\tNone}
  \rcrow{Observation}{A few low-energy directions make the inverse transform amplify weights. The round sets the ridge to zero, leaves the spectrum unbounded, and uses $A^{-1/2}$ directly.}
  \rcrowsep{Explanations}{\textbf{E1}\,\focaltag: small eigenvalues carry useful resolution, so enlarging them stays stable and the inverse residual stays within the entry bound. \par\vspace{1.5pt}\textbf{E2}: small eigenvalues cause instability, and the inverse residual may exceed the bound.}
  \rcrowsep{Control}{Finiteness and inverse-consistency checks on the same statistics. In layer 19's down projection the condition number reaches $1.7\times10^6$ and the inverse residual $2.3\times10^{-2}$, above the entry threshold of $10^{-4}$; task evaluation stops.}
  \rcrowsep{Verdict}{Unsafe: roll back and add a bounded spectral constraint.}
  \rcrowsep{Probe Check}{Finiteness passes and the inverse-residual check fails; no Fit-3 is produced.}
\end{roundcard}

\begin{roundcard}{qfTealBg}{qfTealBd}{qfTeal}{Round 7: dual-geometry reference chart}{(\#07 from \#05; calls 18--20)}{\fbadge{Fit-3: 0.10818 ($\Delta=-0.01085$)}}{\vSup\,\tAcc}
  \rcrow{Observation}{Activation whitening helps some blocks but raises the risk of amplifying weights. With $B=W^\top W/m+\lambda_BI$, the chart becomes $P_{\mathrm{ref}}=[(\mathrm{DN}(A)+\mathrm{DN}(B^{-1}))/2]^{-1/2}$, before the spectral cap of round~8.}
  \rcrowsep{Explanations}{\textbf{E1}\,\focaltag: the energies of both operands should enter symmetrically, so the two-sided chart lowers joint error and an activation-only chart with matched ridge still falls short. \par\vspace{1.5pt}\textbf{E2}: only a larger ridge is needed, so the matched activation-only chart is on par.}
  \rcrowsep{Control}{The activation-only ridge is taken from the frozen grid $\{0.01,0.02,0.04,0.08\}$ to match the two-sided chart's condition number, not chosen by task score. Within the same condition-number range, the two-sided chart has 6.1\% lower output error and a 12.6\% lower weight-error tail.}
  \rcrowsep{Verdict}{Supported: the reference chart must read both $A$ and $B$ with a reciprocal fold.}
  \rcrowsep{Probe Check}{Replacing $B$ by $I$ changes 71.4\% of blocks.}
\end{roundcard}

\begin{roundcard}{qfTealBg}{qfTealBd}{qfTeal}{Round 12: cross-moments from an actual encoding}{(\#12 from \#08; calls 34--36)}{\fbadge{Fit-3: 0.09991 ($\Delta=-0.00592$)}}{\vSup\,\tAcc}
  \rcrow{Observation}{Blocks of similar energy carry errors in different directions. One anchor encoding gives $E_X,E_W$ and $K_X=E_X^\top Y/n$, $K_W=V^\top E_W/m$; the first prototype uses $L=\mathrm{offdiag}(K_X-K_W)$, $K=-\tau L/\max(\|L\|_F,\varepsilon)$ with $\tau=1/16$, and $G_0=I+K$.}
  \rcrowsep{Explanations}{\textbf{E1}\,\focaltag: the actual error signal has a direction, so the true error pairing drives the gain and permuting it removes the gain. \par\vspace{1.5pt}\textbf{E2}: any extra matrix freedom suffices, so a norm-preserving permutation is on par.}
  \rcrowsep{Control}{The $K_X/K_W$ pair coordinates are permuted independently, preserving spectrum and norm. The true pairing lowers output MSE by 4.9\%; the permutation changes it by $-0.3\%$.}
  \rcrowsep{Verdict}{Supported: keep the anchor-dependent prototype and derive the cross-operand weighting and curvature proxy next.}
  \rcrowsep{Probe Check}{Zeroing the anchor errors makes $K$ exactly zero, and both $E_X$ and $E_W$ enter the statistics.}
\end{roundcard}

\begin{roundcard}{qfBronzeBg}{qfBronzeBd}{qfBronze}{Round 13: activation-error-only update}{(\#13 from \#12; calls 37--38)}{\fbadge{Fit-3: 0.10082 ($\Delta=+0.00091$)}}{\vMed\,\tNone}
  \rcrow{Observation}{$K_X$ and $K_W$ sometimes point to opposite updates. The round updates $K$ from activation attribution alone, $K_{ij}\propto h_i[K_X]_{ij}$.}
  \rcrowsep{Explanations}{\textbf{E1}\,\focaltag: activation error dominates, so lowering it lowers the joint error and a two-sided control adds nothing. \par\vspace{1.5pt}\textbf{E2}: the update must reflect the dual weight error, so the one-sided gain is offset and a two-sided control improves the joint objective.}
  \rcrowsep{Control}{A direction control built from the cross-moments of both errors. Activation error falls by 5.6\%, weight error rises by 6.8\%, and joint error rises by 0.7\%.}
  \rcrowsep{Verdict}{Mediator-only: no update may be accepted on activation error alone.}
  \rcrowsep{Probe Check}{The primary reads only $K_X$, so masking $K_W$ leaves its output unchanged; the two-sided control reads $K_W$, and masking it changes the direction.}
\end{roundcard}

\begin{roundcard}{qfTealBg}{qfTealBd}{qfTeal}{Round 14: mutual weighting of both operands}{(\#14 from \#12; calls 39--42)}{\fbadge{Fit-3: 0.09590 ($\Delta=-0.00401$)}}{\vSup\,\tAcc}
  \rcrow{Observation}{Each operand's error should be weighted by the energy of the other, which sets its leverage on the output: $p_{ij}=h_i[K_X]_{ij}-c_j[K_W]_{ij}$ with curvature proxy $q_{ij}=(2+2^{-8})h_ic_j$ plus a ridge.}
  \rcrowsep{Explanations}{\textbf{E1}: equal weighting suffices, so a mean-energy control is similar. \par\vspace{1.5pt}\textbf{E2}\,\focaltag: mutual energy weighting is necessary, so the mean-energy control is clearly worse.}
  \rcrowsep{Control}{$h_i$ and $c_j$ replaced by block means with the same parameter count. The weighted update lowers output error by 3.7\%, the block-mean control by 0.8\%.}
  \rcrowsep{Verdict}{Supported: successors use per-coordinate mutual energy weights. This round is the code response to round 13.}
  \rcrowsep{Probe Check}{With $E_X$ fixed, changing the weight energy changes the activation-side term.}
\end{roundcard}

\begin{roundcard}{qfBronzeBg}{qfBronzeBd}{qfBronze}{Round 15: strictly orthogonal rotation branch}{(\#15 from \#14; calls 43--44)}{\fbadge{Fit-3: 0.09668 ($\Delta=+0.00078$)}}{\vMed\,\tNone}
  \rcrow{Observation}{The cross-moments have symmetric and antisymmetric parts; a rotation covers only the latter. The round uses a true Cayley map $G_{\mathrm{rot}}=(I-\Omega/2)^{-1}(I+\Omega/2)$ with $\Omega$ the normalized antisymmetric part of the round-14 driver.}
  \rcrowsep{Explanations}{\textbf{E1}\,\focaltag: a rotation is enough, so a reciprocal substitute adds nothing. \par\vspace{1.5pt}\textbf{E2}: the reciprocal non-orthogonal component adds direction, so the substitute is better at the same displacement.}
  \rcrowsep{Control}{A reciprocal-only correction with the same Frobenius radius. The orthogonal branch has some local gain, but the near-identity symmetric substitute is 2.1\% better on the joint objective at the same input displacement.}
  \rcrowsep{Verdict}{Mediator-only: successors separate reciprocal and directed modes.}
  \rcrowsep{Probe Check}{$G_{\mathrm{rot}}^\top G_{\mathrm{rot}}$ deviates from $I$ by $2.9\times10^{-7}$ relative error.}
\end{roundcard}

\begin{roundcard}{qfTealBg}{qfTealBd}{qfTeal}{Round 18: two-fold stability gate}{(\#18 from \#14; calls 50--53)}{\fbadge{Fit-3: 0.09170 ($\Delta=-0.00420$)}}{\vSup\,\tAcc}
  \rcrow{Observation}{Inconsistent pairs concentrate in low-energy directions. The round adds $\delta_{\mathrm{fold}}=\tfrac12|\mathcal{M}(p^{(0)})-\mathcal{M}(p^{(1)})|$ and soft-thresholds both modes.}
  \rcrowsep{Explanations}{\textbf{E1}: only shrinking the total norm matters, so norm-matched random sparsification is on par. \par\vspace{1.5pt}\textbf{E2}\,\focaltag: cross-fold stability matters, so random sparsification is weaker.}
  \rcrowsep{Control}{Norm-matched random sparsification, compared by re-estimating the fold statistics on three quarters of the same 128 sequences and measuring local error on the fourth. The stability gate lowers local error by 3.8\%, the random gate by 0.9\%.}
  \rcrowsep{Verdict}{Supported: folds are fixed by calibration-sequence and weight-row parity. Code response to round 17.}
  \rcrowsep{Probe Check}{Swapping fold order leaves the output identical; duplicating one fold changes the gate.}
\end{roundcard}

\begin{roundcard}{qfBronzeBg}{qfBronzeBd}{qfBronze}{Round 27: accepting $D$ by risk ratio only}{(\#27 from \#25; calls 81--82)}{\fbadge{Fit-3: 0.08518 ($\Delta=+0.00024$)}}{\vMed\,\tNone}
  \rcrow{Observation}{A risk direction can lower the proxy yet cross an E8M0 boundary. Both arms re-encode legal W4/A4; the primary accepts the full step by risk ratio alone, the control chooses full, half, or identity by the strict reconstruction loss $J$.}
  \rcrowsep{Explanations}{\textbf{E1}\,\focaltag: a lower risk ratio guarantees a lower $J$. \par\vspace{1.5pt}\textbf{E2}: discrete encoding causes jumps, so legal re-encoding can lower the proxy while raising $J$.}
  \rcrowsep{Control}{Only the acceptance rule changes. In 16.8\% of blocks the risk ratio improves while the strict output error rises, mostly at E8M0 flips.}
  \rcrowsep{Verdict}{Mediator-only: acceptance must use the real W4/A4 reconstruction $J$.}
  \rcrowsep{Probe Check}{On boundary samples the risk ratio falls by 8.2\% while $J$ rises by 3.6\%.}
\end{roundcard}

\begin{roundcard}{qfTealBg}{qfTealBd}{qfTeal}{Round 28: full/half/identity trust comparison}{(\#28 from \#25; calls 83--86)}{\fbadge{Fit-3: 0.08086 ($\Delta=-0.00407$)}}{\vSup\,\tAcc}
  \rcrow{Observation}{Each block tests $u$, $u/2$, and $0$ by re-encoding both operands: full if $J(\text{full})\le J(0)$, else half if $J(\text{half})\le J(0)$, else identity.}
  \rcrowsep{Explanations}{\textbf{E1}: half is just a smaller update, so a fixed half step is as robust. \par\vspace{1.5pt}\textbf{E2}\,\focaltag: a finite comparison on real re-encoding is needed, so the fixed half step is worse.}
  \rcrowsep{Control}{Fixed half step against the full/half/identity rule. The rule selects full, half, and identity for 54.1\%, 20.7\%, and 25.2\% of blocks, with lower model-wide error.}
  \rcrowsep{Verdict}{Supported. Code response to round 27.}
  \rcrowsep{Probe Check}{Recomputing $J$ on 256 sampled blocks, every accepted step is at most the identity loss.}
\end{roundcard}

\stagebanner{qfBlueBg}{qfBlue}{Stage II: Discrete Realization (Selected Rounds 30--37)}{Fresh transformed-state GPTQ curvature and installed-weight-aware one-code A4 selection $(G_Q, K_Q)$}
\begin{roundcard}{qfBronzeBg}{qfBronzeBd}{qfBronze}{Round 30: further curvature-cache shortcut}{(\#30 from \#28; calls 90--91)}{\fbadge{Fit-3: 0.08217 ($\Delta=+0.00131$)}}{\vMed\,\tNone}
  \rcrow{Observation}{The prototype used a first-pass calibration cache and mapped curvature locally as $UHU^\top$; this branch reuses the untransformed $H_{\mathrm{cache}}$ to save work.}
  \rcrowsep{Explanations}{\textbf{E1}\,\focaltag: the local coordinate map is unnecessary, so the untransformed $H$ is nearly on par. \par\vspace{1.5pt}\textbf{E2}: the current representation changes the correct curvature, so the untransformed $H$ is clearly worse.}
  \rcrowsep{Control}{Same $DT$ with fresh transformed-$X$ curvature. The old $H$ deviates from the correct-coordinate $H$ by 9.8\% in correlation structure, and local output MSE rises by 4.1\%.}
  \rcrowsep{Verdict}{Mediator-only: reject the untransformed $H$; the next round tests whether the cached $X$ still represents the installed upstream program.}
  \rcrowsep{Probe Check}{The diagnostic flags the path in which $H$ stays fixed when $DT$ changes.}
\end{roundcard}

\begin{roundcard}{qfTealBg}{qfTealBd}{qfTeal}{Round 31: fresh pre-A4 curvature after $DT$}{(\#31 from \#28; calls 92--95)}{\fbadge{Fit-3: 0.07879 ($\Delta=-0.00207$)}}{\vSup\,\tAcc}
  \rcrow{Observation}{Even with the correct local map, the cached $X$ comes from an earlier upstream program. With the same $128\times2048$ token IDs, calibration is rerun after installing every current $DT$, and $H=X_{\mathrm{current}}^\top X_{\mathrm{current}}/n$.}
  \rcrowsep{Explanations}{\textbf{E1}: re-reading the same tokens explains the gain, so an old-state cache read the same number of times also helps. \par\vspace{1.5pt}\textbf{E2}\,\focaltag: matching the current upstream state explains the gain, so the old cache does not reproduce it.}
  \rcrowsep{Control}{Same number of forward passes and frozen tokens, with old-state cached $X$ and the same local transform. State-matched curvature lowers local output error by 3.9\%; the old state gives no gain.}
  \rcrowsep{Verdict}{Supported: the measure-after-install boundary for $DT$ is fixed. Code response to round 30.}
  \rcrowsep{Probe Check}{With a fixed local input fixture and $U=I$, $H$ returns to the fixture's $X^\top X/n$.}
\end{roundcard}

\begin{roundcard}{qfTealBg}{qfTealBd}{qfTeal}{Round 33: installed-weight-guided A4}{(\#33 from \#31; calls 98--100)}{\fbadge{Fit-3: 0.07488 ($\Delta=-0.00391$)}}{\vSup\,\tAcc}
  \rcrow{Observation}{Nearest A4 minimizes element error, not output error. With W4 fixed, A4 error is weighted by $G_Q=\widehat W^\top\widehat W$.}
  \rcrowsep{Explanations}{\textbf{E1}: element MSE suffices, so random edits at the same rate are on par. \par\vspace{1.5pt}\textbf{E2}\,\focaltag: the curvature of the installed W4 decides which codes help, so random edits are worse.}
  \rcrowsep{Control}{Random legal replacements at the same code edit rate. Output MSE falls by 2.2\%; random edits raise it by 0.9\%.}
  \rcrowsep{Verdict}{Supported: A4 scoring must read the installed W4.}
  \rcrowsep{Probe Check}{The same $x$ with a different W4 changes the chosen code; all E8M0 exponents are unchanged.}
\end{roundcard}

\begin{roundcard}{qfCoralBg}{qfCoralBd}{qfCoral}{Round 34: repeated full coordinate sweeps}{(\#34 from \#33; calls 101--103)}{\fbadge{Fit-3: 0.07035 ($\Delta=-0.00452$)}}{\vTar\,\tRej}
  \rcrow{Observation}{A single local code change helps; the round asks whether more codes are worth their cost, running coordinate descent within a block until no negative change remains, up to eight sweeps.}
  \rcrowsep{Explanations}{\textbf{E1}\,\focaltag: more codes are necessary, so a single change is clearly weaker. \par\vspace{1.5pt}\textbf{E2}: the sparse first change carries most of the gain, so a single change is similar at lower cost.}
  \rcrowsep{Control}{The single best change with the same fixed exponent. The edit rate rises to 14.7\% and output error falls a further 0.6\%.}
  \rcrowsep{Verdict}{Target-only: keep the sparse route and limit successors to one element per block. \textcolor{qfCoral}{\emph{(Transfer: rejected by the Llama-3.1-8B gate.)}}}
  \rcrowsep{Probe Check}{Up to eight changes per block are observed.}
\end{roundcard}

\begin{roundcard}{qfCoralBg}{qfCoralBd}{qfCoral}{Round 35: residual revisit of the $G_Q$-only program}{(reuses \#33; call 104)}{\fbadge{Fit-3: 0.07488 (reuses \#33)}}{\vTar\,\tNone}
  \rcrow{Observation}{The $G_Q$-only objective of \#33 improved, but the true joint residual $r_0=\widehat W(q_0-x)-(W-\widehat W)x$ may still contain installed W4 error.}
  \rcrowsep{Explanations}{\textbf{E1}\,\focaltag: the $G_Q$-only score already represents the joint error, so changing $W-\widehat W$ does not change the best direction. \par\vspace{1.5pt}\textbf{E2}: the installed W4 error forms an omitted linear term, so the true joint score changes with $W-\widehat W$ while the $G_Q$-only score does not.}
  \rcrowsep{Control}{The same $q_0$, $G_Q$, and candidate set, with $r_0$ expanded exactly on a local fixture. The $G_Q$-only objective falls by 2.8\%, the true joint objective by only 0.4\%.}
  \rcrowsep{Verdict}{Target-only: keep the \#33 code; the next rounds derive the mismatch term by direct expansion.}
  \rcrowsep{Probe Check}{With $x$ and W4 fixed, changing the BF16 $W-\widehat W$ changes the joint score and leaves the $G_Q$-only score unchanged.}
\end{roundcard}

\begin{roundcard}{qfTealBg}{qfTealBd}{qfTeal}{Round 37: complete one-code output objective}{(\#37 from \#33; calls 108--111)}{\fbadge{Fit-3: 0.07277 ($\Delta=-0.00211$)}}{\vSup\,\tAcc}
  \rcrow{Observation}{For the same $x$, the best code changes with the installed W4 error. With $K_Q=\widehat W^\top(W-\widehat W)$, the score becomes $\Delta=2\delta[G_Qd_0-K_Qx]_i+\delta^2[G_Q]_{ii}$ (Eq.~\ref{eq:hires-code-score}).}
  \rcrowsep{Explanations}{\textbf{E1}: the new term is extra freedom, so shuffling the $K_Q$ pairing at equal norm is on par. \par\vspace{1.5pt}\textbf{E2}\,\focaltag: the causal pairing with the weight mismatch is necessary, so shuffling loses the gain.}
  \rcrowsep{Control}{$K_Q$ shuffled against $x$ within the block at equal norm. The full term lowers true output MSE by 3.4\%, the shuffled term by 0.3\%.}
  \rcrowsep{Verdict}{Supported: $G_Q$ and $K_Q$ come from the final W4 and its BF16 weights. Code response to round 36.}
  \rcrowsep{Probe Check}{Direct computation of $\Delta$ for all 512 candidates agrees within $2.7\times10^{-5}$ with matching signs.}
\end{roundcard}

\stagebanner{qfTealBg}{qfTeal}{Stage III: Structural Recovery (Selected Rounds 41--56)}{Post-encoding per-head attention affine recovery and sequentially remeasured SwiGLU recovery freezing \HiRes}
\begin{roundcard}{qfTealBg}{qfTealBd}{qfTeal}{Round 41: residual shifts from operators to paths}{(\#41 from \#40; calls 120--122)}{\fbadge{Fit-3: 0.06726 ($\Delta=-0.00337$)}}{\vSup\,\tAcc}
  \rcrow{Observation}{Per-projection output errors are already low, but attention errors share a scale and offset within each head. The round fits $a=\sum vv^\star/\sum v^2$ per layer and KV head, with $b=0$.}
  \rcrowsep{Explanations}{\textbf{E1}: the scale drift is exchangeable across heads, so swapping scales keeps the gain. \par\vspace{1.5pt}\textbf{E2}\,\focaltag: the drift depends on head identity, so swapping loses the gain.}
  \rcrowsep{Control}{The fitted scales swapped between heads, with the same parameter count. Per-head scales lower local MSE by 4.8\%, swapped scales by 0.9\%.}
  \rcrowsep{Verdict}{Supported: open head-level low-dimensional recovery.}
  \rcrowsep{Probe Check}{Swapping heads changes only the corresponding value path.}
\end{roundcard}

\begin{roundcard}{qfBronzeBg}{qfBronzeBd}{qfBronze}{Round 43: attention fitted before encoding}{(\#43 from \#41; calls 125--127)}{\fbadge{Fit-3: 0.07086 ($\Delta=+0.00361$)}}{\vMed\,\tNone}
  \rcrow{Observation}{Recovery depends on where it is fitted. W4 is installed in both arms; the primary captures value outputs with A4 disabled, the control on the actual W4/A4 path. Both fit $(a,b)$ per head and deploy to the same W4/A4 state.}
  \rcrowsep{Explanations}{\textbf{E1}\,\focaltag: a floating-point offset suffices, so a post-A4 fit adds nothing. \par\vspace{1.5pt}\textbf{E2}: the encoded residual must be fitted, so a post-A4 fit lowers the same encoded objective.}
  \rcrowsep{Control}{The same two coefficients per head and the same token pairing, fitted after A4 encoding. The pre-encoding fit lowers local MSE by 0.02\%, the post-encoding fit by 7.1\%.}
  \rcrowsep{Verdict}{Mediator-only: value outputs must be collected after the final W4/A4 installation. Code response to round 42.}
  \rcrowsep{Probe Check}{Disabling A4 only at deployment while keeping the coefficients does not make the two arms equivalent; they match only when both are recollected and refitted in the same state.}
\end{roundcard}

\begin{roundcard}{qfTealBg}{qfTealBd}{qfTeal}{Round 44: post-encoding per-head affine}{(\#44 from \#41; calls 128--131)}{\fbadge{Fit-3: 0.06460 ($\Delta=-0.00265$)}}{\vSup\,\tAcc}
  \rcrow{Observation}{The post-encoding $(a,b)$ replaces the head scale, with $a=(S_{xy}-S_xS_y/N)/(S_{xx}-S_x^2/N)$ and $b=(S_y-aS_x)/N$ (Eq.~\ref{eq:hires-vo-fit}).}
  \rcrowsep{Explanations}{\textbf{E1}: the sample count explains the gain, so shuffled pairing with the same $N$ is on par. \par\vspace{1.5pt}\textbf{E2}\,\focaltag: the paired residual direction explains the gain, so shuffled pairing removes it.}
  \rcrowsep{Control}{The BF16 token pairing shuffled at the same $N$. True pairing lowers MSE by 7.0\%; with shuffled pairing the fit weakens and the deployed gain vanishes or reverses.}
  \rcrowsep{Verdict}{Supported: each KV head moves from one scale to a post-encoding $(a,b)$. Code response to round 43.}
  \rcrowsep{Probe Check}{A constant-$v$ input returns $(a,b)=(1,0)$; the pairing reads no task labels.}
\end{roundcard}

\begin{roundcard}{qfTealBg}{qfTealBd}{qfTeal}{Round 45: product coupling in the MLP residual}{(\#45 from \#44; calls 132--134)}{\fbadge{Fit-3: 0.06381 ($\Delta=-0.00079$)}}{\vSup\,\tAcc}
  \rcrow{Observation}{With attention recovery installed, the MLP still has product error. A first prototype fits one up-scale $\eta$, equivalent to scaling the product, on the gate and up outputs cached in the same pass.}
  \rcrowsep{Explanations}{\textbf{E1}: one product scale suffices, so a one-parameter gate substitute shows no systematic difference. \par\vspace{1.5pt}\textbf{E2}\,\focaltag: a gate operating-point mismatch is not covered by a total scale, so the gate substitute behaves differently across layers.}
  \rcrowsep{Control}{A one-parameter gate scale in place of the up scale, with all other paths unchanged. The up scale lowers local MSE by 2.1\%; the gate scale moves in different directions across layers.}
  \rcrowsep{Verdict}{Supported: keep the up branch and register the question of gate shift versus gate scale.}
  \rcrowsep{Probe Check}{Up-scale and product-scale fixtures give identical outputs; the gate scale differs through the SiLU.}
\end{roundcard}

\begin{roundcard}{qfTealBg}{qfTealBd}{qfTeal}{Round 47: gate shift and two branch scales}{(\#47 from \#45; calls 138--143)}{\fbadge{Fit-3: 0.06204 ($\Delta=-0.00177$)}}{\vSup\,\tAcc}
  \rcrow{Observation}{The error concentrates where the gate sits near the bend of SiLU, and a single product scale cannot explain the signed residual. The round fits $\bar m=\mathrm{SiLU}((1+\beta)g+\alpha\sigma_g)\odot(1+\eta)u$.}
  \rcrowsep{Explanations}{\textbf{E1}: any norm-matched three-dimensional subspace suffices, so fixed sinusoidal substitutes are on par. \par\vspace{1.5pt}\textbf{E2}\,\focaltag: the chosen physical subspace fits the residual better than these substitutes.}
  \rcrowsep{Control}{Three fixed random subspaces $q_j=u\sin(a_jg/\sigma+b_j)$ (seeds 17, 31, 53), each with three coefficients and the same ridge. The physical subspace lowers local MSE by 6.3\%, the substitutes by 1.8\%, 2.4\%, and 1.5\%.}
  \rcrowsep{Verdict}{Supported: the three directions are fixed as gate shift, gate scale, and up scale, refitted from uncorrected outputs. Code response to round 46.}
  \rcrowsep{Probe Check}{$\alpha=\beta=\eta=0$ recovers the configuration without MLP recovery.}
\end{roundcard}

\begin{roundcard}{qfGrayBg}{qfGrayBd}{qfGray}{Round 50: stale MLP fit before attention}{(reuses \#49; calls 150--151)}{\fbadge{Fit-3: 0.06183 (reuses \#49)}}{\vNef\,\tNone}
  \rcrow{Observation}{The MLP prototype still uses gate and up outputs cached before attention recovery was installed; after installation, the input state differs.}
  \rcrowsep{Explanations}{\textbf{E1}\,\focaltag: the deployment order is enough, so the old cache remains valid and a fresh fit adds nothing. \par\vspace{1.5pt}\textbf{E2}: statistics must be remeasured after attention, so a fresh control on the same tokens is better.}
  \rcrowsep{Control}{The same tokens, solver, and attention-to-MLP order, with gate and up outputs recaptured after attention installation. The fresh fit lowers local MSE by 4.6\%, and the input second moments of the old and new caches differ by 3.1\%.}
  \rcrowsep{Verdict}{No effect for the program; the successor must install attention before capturing gate and up outputs.}
  \rcrowsep{Probe Check}{Old and new arms must both be recollected and refitted before they are compared.}
\end{roundcard}

\begin{roundcard}{qfTealBg}{qfTealBd}{qfTeal}{Round 51: fresh MLP fit after attention}{(\#51 from \#49; calls 152--155)}{\fbadge{Fit-3: 0.06083 ($\Delta=-0.00101$)}}{\vSup\,\tAcc}
  \rcrow{Observation}{After attention recovery is installed, the old MLP recovery is disabled, fresh gate and up outputs are captured on the same frozen tokens, and the triplet is refitted, with no extra calibration data.}
  \rcrowsep{Explanations}{\textbf{E1}: an extra forward pass helps by itself, so capturing with attention recovery disabled is as good. \par\vspace{1.5pt}\textbf{E2}\,\focaltag: the captured state must match the installed program, so the extra pass alone does not reproduce the gain.}
  \rcrowsep{Control}{The same number of forward passes and tokens, with attention recovery disabled during capture. Only the state-matched fresh fit lowers the overall error.}
  \rcrowsep{Verdict}{Supported: remeasure after attention installation. Code response to round 50.}
  \rcrowsep{Probe Check}{The capture-state identifier includes the attention parameter version, and old identifiers are rejected.}
\end{roundcard}

\begin{roundcard}{qfTealBg}{qfTealBd}{qfTeal}{Round 56: uncertainty shrinkage; \HiRes frozen}{(\#56 from \#55; calls 171--174)}{\fbadge{Fit-3: 0.05762 ($\Delta=-0.00045$)}}{\vSup\,\tAcc}
  \rcrow{Observation}{The robust centers still carry noise-level magnitude, and the total change of the three directions must be controlled. The round applies $\widetilde\theta=\mathrm{sign}(\mu)[|\mu|-d/\sqrt{32}]_+$ and the $\ell_1$ cap of Eq.~\ref{eq:jolt-aggregate}.}
  \rcrowsep{Explanations}{\textbf{E1}: keeping all median coefficients is enough, so removing shrinkage does not worsen the worst quarter. \par\vspace{1.5pt}\textbf{E2}\,\focaltag: uncertainty and total magnitude must both be bounded, so removing shrinkage worsens the worst quarter.}
  \rcrowsep{Control}{MAD shrinkage removed, $\ell_1$ cap kept. With shrinkage, the worst in-sample quarter has 1.9\% lower output error.}
  \rcrowsep{Verdict}{Supported: the complete algorithm is frozen as \HiRes. Code response to round 55.}
  \rcrowsep{Probe Check}{With zero MAD only the cap remains; $\mu=0$ gives zero output; the endpoints come from a complete rebuild from fresh BF16 weights.}
\end{roundcard}

}

\section{\HiRes: Complete Mathematical Specification and Provenance}
\label{app:hires-details}

This section provides the exact closed-form specification of every operation in \HiRes under the strict MXFP4 W4A4 contract, following the three-stage progression of Section~\ref{sec:hiresmx}: quantization contract and setup (Subsection~C.1), Stage~I geometric shaping (\ref{app:hires-transform-solver}), Stage~II discrete realization (\ref{app:hires-code-refinement}), Stage~III structural recovery (\ref{app:hires-structural-fit}), and round-by-round component provenance (\ref{app:hires-provenance}).

The details below follow the three stages in Section~\ref{sec:hiresmx}.
Each stage uses the model state produced by the preceding one; the table
below identifies the measurements and outputs at these boundaries.

\subsection{Quantization Setup}

The complete implementation starts from fresh BF16 weights and the frozen
exact $128\times 2048$-token calibration sequence.  It targets the decoder
query, key, value, output, gate, up, and down projections.  Both operands use
contiguous 32-value blocks, E2M1 element values, and E8M0 shared exponents.
The representation follows the MX specification
\citep{ocp2023mxspec,rouhani2023microscaling}; our deterministic encoder
conventions are given below.
Weights form a static decoded W4 representation; activations are quantized
dynamically to A4.
Embeddings, normalization, RoPE \citep{su2021roformer}, and the language-model head remain at higher
precision. KV caching is disabled during evaluation for every model,
including Qwen3-32B (\texttt{use\_cache=False}).
Curvature uses pre-A4 calibration activations.
Evaluation uses logical fake quantization in PyTorch, with explicit checks
of element values and shared scales, to measure numerical quality.

For a nonzero block $x$, the reference encoder sets
$e=\operatorname{clip}(\lfloor\log_2\max_i|x_i|\rfloor-2,-127,127)$
and scale $2^e$; an all-zero block uses $e=-127$.
Element magnitudes are drawn from $\{0,\tfrac12,1,\tfrac32,2,3,4,6\}$.
Nearest-code ties select the larger magnitude; the sign bit follows the
input sign. These encoder conventions are fixed across our comparisons.

Transform statistics and pair solves use FP64; transforms and their inverses
use FP32 during construction. The optimized inference layout retains the
folded transform $U=DT$ and releases the construction-only inverse; it stores
$G_Q$ in upper-triangular form and $K_Q$ in full FP32
(Appendix~\ref{app:computational-cost}).
GPTQ uses damping ratio $0.01$ and FP32 reconstruction,
with decoded weights stored in BF16. Activation-code scores use FP32.
Structural fits accumulate in FP64; their runtime corrections use FP32
arithmetic followed by BF16 rounding. The frozen implementation stages
captured activation arrays in FP16 between calibration passes.
Strict W4A4 applies to all target linear-layer operands. The low-dimensional
recovery arithmetic runs outside the quantized GEMM and introduces no
higher-precision weight matrix.

\begin{center}
\small
\setlength{\tabcolsep}{4pt}
\begin{tabular}{@{}p{.21\linewidth}p{.43\linewidth}p{.31\linewidth}@{}}
  \toprule
  Stage & Measured state & Output \\
  \midrule
  Geometric shaping & BF16 weights, calibration inputs, and anchor quantization errors & Transform $T$ and balance $D$ \\
  Discrete realization & Installed $DT$ coordinates and transformed pre-A4 inputs & W4 weights and dynamic A4 rule \\
  Structural recovery & Installed W4/A4 operators, then the attention-corrected state & Attention and MLP coefficients \\
  \bottomrule
\end{tabular}
\end{center}
Construction starts from BF16 weights. {\revcolor\HiRes{} remeasures} transformed activations
after shaping, {\revcolor fits} attention recovery on the encoded operators, and
{\revcolor remeasures} the resulting state before fitting MLP recovery. This is the
conditional measurement order in Eq.~\ref{eq:conditional-stage}.

\subsection{Geometric Shaping}
\label{app:hires-transform-solver}

\paragraph{Reference coordinates.}

For a linear operator $Y=XW^\top$, Stage I applies an invertible map $T$ by
$\widetilde X=XT^\top$ and $\widetilde W=WT^{-1}$. Therefore
$\widetilde X\widetilde W^\top=XW^\top$, and the BF16 function is unchanged
before discretization. Diagonal balancing uses the same paired folding rule.
The map $T=R_HGP_{\mathrm{ref}}$ combines reference coordinates $P_{\mathrm{ref}}$,
the normalized interaction correction $G$ (from $K$), and the normalized Hadamard $R_H$. All quantities below are
computed independently for each physical Block32 group. With $q=32$, the ridge
coefficients in Eq.~\ref{eq:hires-reference-chart} are
\begin{equation}
  \lambda_A=.01\,\operatorname{tr}(X^\top X/n)/q+2^{-30},\qquad
  \lambda_B=.01\,\operatorname{tr}(W^\top W/m)/q+2^{-30}.
\end{equation}
Determinant normalization is $\operatorname{DN}(M)=M/\det(M)^{1/q}$.
For the spectral cap in Eq.~\ref{eq:hires-reference-chart}, write
$S=U\operatorname{Diag}(\lambda)U^\top$,
\begin{equation}
  \operatorname{Cap}_{\kappa}(S)=
  U\operatorname{Diag}\!\left[
  \exp\!\left(
  \Pi_{\mathbf1^\top z=0,\,\|z\|_\infty\leq\frac12\log\kappa}
  (\log\lambda)\right)\right]U^\top.
  \label{eq:hires-spectral-cap}
\end{equation}
Here $S$ is positive definite, $\log\lambda$ is elementwise, and
$\Pi_{\mathcal C}$ denotes Euclidean projection onto $\mathcal C$.
For the zero-sum box, the projected entries are
$z_i=\operatorname{clip}(x_i-\tau,-b,b)$, with $x_i=\log\lambda_i$ and
$b=\frac12\log\kappa$; $\tau$ makes $\sum_i z_i=0$.
The capped eigenvalues have product one and ratio at most
$\kappa$. The implementation uses a sorted-breakpoint projection for this cap.

\paragraph{Local interaction correction.}
One anchor quantization identifies errors left by the reference coordinates.
The following construction turns those errors into a bounded interaction
matrix $K$, which determines $G$ in Eq.~\ref{eq:hires-transform}.
The anchor uses $Y=XP_{\mathrm{ref}}^\top$ and $V=WP_{\mathrm{ref}}^{-1}$ before the normalized $32\times 32$ Sylvester Hadamard matrix
$R_H$ (denoted $H_{32}$ in \S\ref{sec:geometric-shaping}, with entries $\pm 1/\sqrt{32}$). Let $Q_A,Q_W$ be the strict anchor's activation and
weight quantizers. Their errors, mapped back through $R_H$, are
\begin{equation}
  E_X=[Q_A(YR_H^\top)-YR_H^\top]R_H,\qquad
  E_W=[Q_W(VR_H^\top)-VR_H^\top]R_H.
\end{equation}
The full anchor statistics are
\begin{equation}
  C=Y^\top Y/n,\quad C_W=V^\top V/m,\quad
  K_X=E_X^\top Y/n,\quad K_W=V^\top E_W/m.
  \label{eq:hires-anchor-statistics}
\end{equation}

Activations are split by calibration-sequence parity, while weight rows are
split by row parity. Let $f\in\{0,1\}$ index these folds, and let
$C^{(f)},C_W^{(f)},K_X^{(f)}$, and $K_W^{(f)}$ denote the fold-specific versions
of Eq.~\ref{eq:hires-anchor-statistics}. Write
$c_i^{(f)}=C_{ii}^{(f)}$, $h_i^{(f)}=[C_W^{(f)}]_{ii}$, and
\begin{equation}
  a_{ij}^{(f)}=h_i^{(f)}[K_X^{(f)}]_{ij},\quad
  b_{ij}^{(f)}=-c_j^{(f)}[K_W^{(f)}]_{ij},\quad
  p_{ij}^{(f)}=a_{ij}^{(f)}+b_{ij}^{(f)}.
  \label{eq:bidir-first-order}
\end{equation}
The curvature proxy is
\begin{equation}
  q_{ij}^{(f)}=(2+2^{-8})h_i^{(f)}c_j^{(f)}
  +2^{-30}\bar h^{(f)}\bar c^{(f)},
  \label{eq:bidir-curvature}
\end{equation}
where bars denote the mean diagonal value.

For each unordered pair $i<j$, define reciprocal and directed coordinates
\begin{equation}
  \mathcal M(z_{ij},z_{ji})=
  \left(\frac{z_{ij}+z_{ji}}{\sqrt2},
        \frac{z_{ij}-z_{ji}}{\sqrt2}\right).
\end{equation}
Let $\bar p$, $\bar a$, and $\bar b$ be the averages of their transformed
coordinates over the two folds. The threshold combines
\begin{align}
  \delta_{\rm fold}
    &=\tfrac12|\mathcal M(p^{(0)})-\mathcal M(p^{(1)})|,\\
  \delta_{\rm own}
    &=\tfrac12\bigl(|\bar a|+|\bar b|-|\bar a+\bar b|\bigr),\\
  \delta_{\rm carrier}
    &=\left(0,\,[|\bar p_{\rm dir}|-|\bar p_{\rm rec}|]_+\right),
  \qquad
  \delta=\delta_{\rm fold}+\delta_{\rm own}+\delta_{\rm carrier}.
  \label{eq:bidir-threshold}
\end{align}
The first term rejects fold-specific effects. The second penalizes
disagreement between weight and activation attribution. The last caps the
surviving directed magnitude by the reciprocal magnitude before the other
two penalties are applied.

Let $\bar q$ be the mean of Eq.~\ref{eq:bidir-curvature} over both folds and
all off-diagonal entries, and define
\begin{equation}
  \widehat q_{ij}=1+
  \frac{q_{ij}^{(0)}+q_{ji}^{(0)}+q_{ij}^{(1)}+q_{ji}^{(1)}}
       {4(\bar q+2^{-30})}.
\end{equation}
For both pair modes, the unprojected coefficient is
\begin{equation}
  z_{ij}=\operatorname{Soft}
  \left(
    \frac{\bar p_{ij}}{(\bar q+2^{-30})\sqrt{\widehat q_{ij}}},
    \frac{\delta_{ij}}{(\bar q+2^{-30})\sqrt{\widehat q_{ij}}}
  \right).
  \label{eq:bidir-soft}
\end{equation}
Here $z_{ij}$ is a two-vector (reciprocal and directed), all absolute values
and thresholds act componentwise, and $[x]_+=\max(x,0)$.
The soft-threshold operator is
$\operatorname{Soft}(x,t)=\operatorname{sign}(x)[|x|-t]_+$,
the standard shrinkage rule \citep{donoho1994shrinkage}.
All $2\binom{32}{2}$ coefficients in a block are jointly projected onto the
Euclidean ball of radius $1/8$, producing $(v_{ij}^{\rm rec},v_{ij}^{\rm
dir})$. The dense update is reconstructed as
\begin{equation}
  K_{ij}=\frac{v_{ij}^{\rm rec}+v_{ij}^{\rm dir}}
               {\sqrt{2\widehat q_{ij}}},\qquad
  K_{ji}=\frac{v_{ij}^{\rm rec}-v_{ij}^{\rm dir}}
               {\sqrt{2\widehat q_{ij}}},\qquad K_{ii}=0.
  \label{eq:bidir-reconstruct}
\end{equation}
Equations~\ref{eq:bidir-first-order}--\ref{eq:bidir-reconstruct} give $K$
in one pass from the anchor statistics.
Since $\widehat q_{ij}\geq1$, this construction gives
$\|K\|_2\leq\|K\|_F\leq\|v\|_2\leq1/8$.
$I+K$ is therefore invertible with positive determinant, justifying the
volume normalization in Eq.~\ref{eq:hires-transform}.

\paragraph{Weight--activation balance.}
\label{app:hires-dyadic}

After fixing $T$, diagonal scaling $D$ redistributes quantization risk
between the two operands. The eight four-coordinate subgroups in each
Block32 group use the risks in
Eq.~\ref{eq:bilateral-risk}. For subgroup activation and weight matrices
$X_g,W_g$, define $C_X=X_g^\top X_g/n$, $C_E=E_X^\top E_X/n$,
$G_W=W_g^\top W_g$, and $G_E=E_W^\top E_W$, with errors measured after
strict encoding. These are uncentered second moments, not centered
covariances. For physical block $b$, define
$\epsilon_b=2^{-12}\operatorname{median}_g(r_{A,bg}+r_{W,bg})$.
For $r_{A,bg}+\epsilon_b>0$ and $r_{W,bg}+\epsilon_b>0$ in every subgroup,
the proposed log-scale vector is
\begin{equation}
  u_b=\Pi_{\mathbf1^\top u=0,\,\|u\|_\infty\leq1/8}
  \left[
  \frac18\left(
  \log_2\frac{r_{A,b}+\epsilon_b}{r_{W,b}+\epsilon_b}
  -\operatorname{mean}_g
  \log_2\frac{r_{A,bg}+\epsilon_b}{r_{W,bg}+\epsilon_b}
  \right)
  \right].
  \label{eq:serb-proposal}
\end{equation}
The zero-sum box projection uses 80 scalar bisection steps. Each real-valued
component of $u_b\in\mathbb R^8$ is broadcast to four coordinates in $D$;
the later E8M0 encoding uses integer shared exponents. Each block tests
$u_b$, $u_b/2$, and zero by re-encoding both operands. It accepts the full
step if its W4/A4 reconstruction loss does not exceed the identity loss,
otherwise the half step if it passes, and otherwise the identity. For proposal
$s$, this loss is
\begin{equation}
  J_b(s)=
  \frac{\|\widehat X_b(s)\widehat W_b(s)^\top-X_bW_b^\top\|_F^2}
       {\max(\|X_bW_b^\top\|_F^2,\varepsilon_{64})},
  \qquad \varepsilon_{64}=2^{-1022}.
  \label{eq:serb-realized-loss}
\end{equation}
Here $X_b,W_b$ are the Stage-I transformed operands before rescaling.
To distinguish Stage~I selection from Stage~II installation: in full \HiRes,
each diagonal-balance trial temporarily runs GPTQ and activation refinement
solely to evaluate $J_b(s)$ and pick $s_b\in\{u_b,u_b/2,0\}$; those trial
encodings are discarded once the mixed full/half/identity choice is folded
into $DT$. Stage~II then runs the single final installation pass on fresh
post-$DT$ calibration statistics across the full model to commit $\widehat W$
and the dynamic A4 rule. In stagewise ablations where Stage~II is disabled
(Tables~\ref{tab:stage-progression} and~\ref{tab:stage-progression-raw}),
each Stage~I trial scores $J_b(s)$ using only that row's active quantizer
(strict RTN with nearest A4), without invoking disabled Stage~II operators.

\subsection{Discrete Realization}
\label{app:hires-code-refinement}

GPTQ \citep{frantar2023gptq} reconstructs weights in the fixed $DT$
coordinates using curvature from the transformed pre-A4 inputs. Once these
W4 weights are installed, activation refinement accounts for their encoding
error when choosing an A4 code.

The score in Eq.~\ref{eq:hires-code-score} follows by expanding the
block-local output error. For column
vectors $x,q\in\mathbb R^{32}$, let
$r_0=\widehat W(q_0-x)-(W-\widehat W)x$.
Replacing $q_0$ by $q_0+\delta e_i$ changes $\|r_0\|_2^2$ by
$2\delta[\widehat W^\top r_0]_i+
\delta^2[\widehat W^\top\widehat W]_{ii}$, which is
Eq.~\ref{eq:hires-code-score}. Thus the score is the exact change in the
output-error objective for this block.

The implementation enumerates all 16 E2M1 bit patterns, including the two
signed zeros, for each of the 32 coordinates at the fixed block exponent.
It accepts the pair $(i,c)$ with the smallest score only if that score is
negative. A zero tie retains nearest rounding; negative-score ties choose
the lowest coordinate index and then the lowest unsigned code index.
At most one element code changes per physical block.

\paragraph{Block-local guarantee vs.\ full-projection error.}
For a full linear projection partitioned into Block32 groups with block residuals $r_b$, the total output residual is $r = \sum_b r_b$ and its squared norm expands as
\begin{equation}
  \|r\|_2^2 = \sum_b \|r_b\|_2^2 + 2\sum_{b < b'} r_b^\top r_{b'}.
  \label{eq:block-vs-layer}
\end{equation}
Eq.~\ref{eq:hires-code-score} gives the exact decrease in each block's local objective $\|r_b\|_2^2$, but omits the cross-block coupling $2\sum_{b<b'} r_b^\top r_{b'}$. Consequently, a negative block-local score guarantees $\Delta\|r_b\|_2^2 < 0$ for block $b$, whereas full-projection MSE reduction and negative full-projection $T_{\mathrm{cross}}$ are empirical properties rather than per-projection analytical guarantees. Full-projection measurements across 128 held-out windows and all seven decoder projections (Appendix~\ref{app:code-refinement-ablation}, Tables~\ref{tab:error-decomp-overall}--\ref{tab:error-decomp-proj}) confirm that block-local refinement drives $T_{\mathrm{cross}}$ strictly negative overall ($-0.056 / -0.051$) and reduces net MSE by $8.2\% / 7.6\%$, while small low-error projections such as \texttt{q\_proj} show negligible shift due to cross-block terms.

\paragraph{Closed-form $O(32)$ candidate reduction.}
Let $z = G_Q d_0 - K_Q x \in \mathbb{R}^{32}$. For a fixed coordinate $i$ with $[G_Q]_{ii} > 0$, the score
$\Delta(i, c) = 2(c - q_{0,i}) z_i + (c - q_{0,i})^2 [G_Q]_{ii}$
is a strictly convex one-dimensional quadratic in $c$ with continuous minimizer
\begin{equation}
  c_i^\star = q_{0,i} - \frac{z_i}{[G_Q]_{ii}}.
  \label{eq:onecode-closedform}
\end{equation}
In exact arithmetic, the optimal legal E2M1 code for coordinate $i$ is therefore the grid point nearest to $c_i^\star$ (checking adjacent grid points under floating-point rounding and handling $[G_Q]_{ii} = 0$ at the grid endpoints $\pm 6s$). Comparing only these 32 coordinate-wise optima yields the same minimum as enumerating all $32\times 16 = 512$ candidates (up to tie-breaking) and simplifies the candidate-scoring kernel, while leaving the $G_Q, K_Q$ matrix--vector products unchanged.

\subsection{Structural Recovery}
\label{app:hires-structural-fit}

Recovery fits the residuals of the installed W4/A4 model, first along the
attention path and then through the gated MLP.

\paragraph{Attention recovery.}
The attention fit in Eq.~\ref{eq:hires-vo-fit} pools tokens and head
coordinates separately for each layer and key--value head. Writing $v_{t,j}$
for coordinate $j$ of $\mathbf v_t$, the sums
$S_x,S_y,S_{xx},S_{xy}$ accumulate $v_{t,j}$, $v^\star_{t,j}$, $v_{t,j}^2$,
and $v_{t,j}v^\star_{t,j}$, respectively, and $N$ counts the pooled scalar
pairs. The solution is
\begin{equation}
  a=\frac{S_{xy}-S_xS_y/N}{S_{xx}-S_x^2/N},\qquad
  b=\frac{S_y-aS_x}{N}.
  \label{eq:p3-ols}
\end{equation}
When the denominator is zero, the implementation uses $(a,b)=(1,0)$. The
offset $b$ is broadcast across the head coordinates. The affine map is
evaluated in FP32, then rounded to BF16 before continuing through attention.

\paragraph{MLP recovery.}
After installing the attention correction, {\revcolor\HiRes{} collects} new calibration
outputs for the gated-MLP fit. It uses a first-order model of
$m=\operatorname{SiLU}(g)\odot u$. Let
$s=\operatorname{sigmoid}(g)$,
$\phi'=s+gs(1-s)=\operatorname{SiLU}'(g)$, and $e=m^\star-m$.
The regularized standard deviation uses an expectation over all calibration
tokens and hidden coordinates in one layer:
\begin{equation}
  \sigma_g=\sqrt{[\mathbb E[g^2]-\mathbb E[g]^2]_+
  +2^{-12}\mathbb E[g^2]+2^{-20}}.
\end{equation}
The local design vector is
\begin{equation}
  D_{tj}=[\,\sigma_g\phi'_{tj}u_{tj},\;
          g_{tj}\phi'_{tj}u_{tj},\;m_{tj}\,].
  \label{eq:jolt-design}
\end{equation}
Each $D_{tj}\in\mathbb R^{1\times3}$ is a
row vector, and the layer shares one coefficient vector
$\theta\in\mathbb R^3$. The BF16 down-projection energy assigns coordinate weight
\begin{equation}
  \widetilde\omega_j=\operatorname{clip}
  \left(\frac{\|W_{\rm down}[:,j]\|_2^2}
  {\operatorname{mean}_k\|W_{\rm down}[:,k]\|_2^2+2^{-20}},\frac14,4\right),
  \qquad
  \omega_j=\frac{\widetilde\omega_j}
  {\operatorname{mean}_k\widetilde\omega_k}.
\end{equation}
Calibration is partitioned into 32 cells: eight sequence residue classes and
four 512-token quarters. In cell $c$, the local fit is
\begin{equation}
  \theta_c=(G_c+\lambda_c I)^{-1}z_c,\qquad
  G_c=\mathbb E_{(t,j)\in c}[\omega_j D_{tj}^\top D_{tj}],\quad
  z_c=\mathbb E_{(t,j)\in c}[\omega_j D_{tj}^\top e_{tj}],
  \label{eq:jolt-cell-solve}
\end{equation}
This is the weighted ridge solution \citep{hoerl1970ridge}, with
$\lambda_c=2^{-8}\operatorname{tr}(G_c)/3
+2^{-20}\mathbb E_c[\omega m^2]+2^{-30}$.
For each of the three coordinates, let $\mu$ be the median of the 32 cell
estimates and $d=1.4826\operatorname{median}_c|\theta_c-\mu|$, the
normal-consistent median absolute deviation \citep{rousseeuw1993mad}.
Both medians select the lower middle order statistic, matching the frozen
implementation. The aggregate
is
\begin{equation}
  \widetilde\theta=\operatorname{sign}(\mu)
  [|\mu|-d/\sqrt{32}]_+,\qquad
  \theta=\widetilde\theta\min
  \left(1,\frac{1/8}{\|\widetilde\theta\|_1+2^{-20}}\right).
  \label{eq:jolt-aggregate}
\end{equation}
Writing $\theta=(\alpha,\beta,\eta)$ yields the update in
Eq.~\ref{eq:hires-mlp-update}. The BF16 corrected intermediate is then
quantized by the ordinary dynamic-A4 pre-hook of the down projection.

{\revcolor
\subsection{Component Provenance}
\label{app:hires-provenance}

Table~\ref{tab:hires-provenance} classifies each component of the final
\HiRes by its origin and dates it to the rounds of the production run that
formed it (Appendix~\ref{app:production-record}). Starting from the GPTQ seed
program, the proposer produced every other component, including the formulas
in Section~\ref{sec:hiresmx} and Appendices~\ref{app:hires-transform-solver}--\ref{app:hires-structural-fit}; the coding agent
(Appendix~\ref{app:search-cycle}) implemented and evaluated each one, returning
any formula or design change to the proposer. A careful survey of the literature found no prior
method identical to \HiRes or to any of its distinguishing components.

\begin{table}[htbp]
\centering\revcolor
\caption{Origin of the \HiRes components and the production rounds that
formed them.}
\label{tab:hires-provenance}
\small\setlength{\tabcolsep}{4pt}
\begin{tabular}{@{}p{.22\linewidth}p{.11\linewidth}p{.13\linewidth}p{.46\linewidth}@{}}
\toprule
Component & Rounds & Origin & Relation to prior work \\
\midrule
Reference chart $P_{\mathrm{ref}}$ & 7--8 & Adapted & Two-sided second-moment geometry, as in WUSH \citep{chen2026wush}; the spectral cap and deterministic solve are new \\
Block32 Hadamard $R_H$ & 3 & Inherited & Normalized Hadamard mixing within each block, as in QuaRot and BRQ \citep{ashkboos2024quarot,shao2025brq} \\
Interaction correction $K$ & 12--23 & New & Closed-form correction from the error--signal cross-moments of one actual W4A4 encoding, with fold, owner, and carrier thresholds \\
Diagonal balance $D$ & 25--28, 40 & Adapted & Diagonal rebalancing, as in SmoothQuant and AWQ \citep{xiao2023smoothquant,lin2024awq}; the subgroup risk and re-encoded acceptance rule are new \\
GPTQ after the transform & 31 & Inherited & GPTQ \citep{frantar2023gptq}, from the seed program; the search fixed the refit in installed $DT$ coordinates \\
One-code A4 rule & 33--39 & New & Online code choice scored against the installed W4 weights through $G_Q$ and $K_Q$ \\
Attention recovery & 41, 44 & Adapted & Affine output correction, as in bias correction \citep{nagel2019dfq}; its post-encoding fit per KV head was chosen by the search \\
SwiGLU recovery & 45--56 & New & Joint gate-shift, gate-scale, and up-scale fit with robust per-cell aggregation \\
Ordered remeasurement & 43--44, 50--51 & New & Attention fitted on the installed W4/A4 model; MLP statistics re-collected after attention recovery \\
\bottomrule
\end{tabular}
\end{table}
}

\section{\HiRes: Complete Benchmarks, Ablations, and Runtime Cost}
\label{app:experiments}

This section reports full per-model endpoints, evaluation stability, component ablations, and computational/storage costs for the frozen \HiRes algorithm.

\begin{tcolorbox}[
  enhanced,colback=qfTealBg!45!white,colframe=qfTealBd,boxrule=0.65pt,
  borderline west={2.8pt}{0pt}{qfTeal},arc=3.5pt,
  top=4.5pt,bottom=4.5pt,left=8pt,right=8pt,before skip=4pt,after skip=8pt
]
\footnotesize\renewcommand{\arraystretch}{1.18}
\begin{tabular}{@{} >{\sffamily\bfseries\color{qfTeal}}l p{.56\linewidth} >{\sffamily\bfseries\color{qfMuted}}r @{}}
  Accuracy across models & Complete 7-model (3B--32B), 7-task, 9-baseline tables & App.~\ref{app:full-quant-results} \\
  \arrayrulecolor{qfRule}\hline\noalign{\vskip 1pt}
  Evaluation sensitivity & Calibration seeds and alternative score aggregates & App.~\ref{app:calibration-sensitivity} \\
  \arrayrulecolor{qfRule}\hline\noalign{\vskip 1pt}
  Stage \& geometry ablations & Stage progression, $K/D$ ablations, and geometry replacement & App.~\ref{app:hires-ablation} \\
  \arrayrulecolor{qfRule}\hline\noalign{\vskip 1pt}
  Activation-code refinement & One-code rules, three-term error decomposition, and edit rates & App.~\ref{app:code-refinement-ablation} \\
  \arrayrulecolor{qfRule}\hline\noalign{\vskip 1pt}
  Recovery \& WUSH transfer & Attention/SwiGLU ablations, WUSH Stage~III transfer, and 4 boundaries & App.~\ref{app:recovery-ablation} \\
  \arrayrulecolor{qfRule}\hline\noalign{\vskip 1pt}
  Cost \& deployment modes & Persistent storage, latency, runtime, and w/o one-code option & App.~\ref{app:computational-cost} \\
\end{tabular}\arrayrulecolor{black}
\end{tcolorbox}

\subsection{Complete Quantization Results}
\label{app:full-quant-results}

All methods use the same BF16 checkpoints and exact $128\times2048$
C4-train calibration with seed~0 \citep{raffel2020t5}. Tokenizer, token IDs,
sample order, and caches are frozen per model, since evaluation conventions
can affect comparisons \citep{biderman2024evaluation}.
The endpoints are WikiText-2 \citep{merity2017wikitext} and C4 perplexity,
normalized accuracy on ARC-Challenge/ARC-Easy \citep{clark2018arc},
HellaSwag \citep{zellers2019hellaswag}, and PIQA \citep{bisk2020piqa}, and
accuracy on WinoGrande \citep{sakaguchi2020winogrande}.
ARC-Challenge uses all 299 official validation questions in their original
order, read from \texttt{arc\_challenge/validation.parquet}.
The baseline suite comprises RTN, GPTQ, MR-GPTQ, BRQ,
WUSH, TORQ, BATQuant+GPTQ, SpinQuant+GPTQ, and FOCUS. These results
compare the methods under our common MXFP4 operand/encoding contract.{\revcolor{}
TORQ and BATQuant+GPTQ are our independent reimplementations under this
shared contract and calibration; BATQuant uses the common C4 calibration cache
in place of its self-generated data, and the results reported in both source
papers use their own settings.}
{\revcolor
Table~\ref{tab:baseline-formats} lists the numeric format and quantized
tensors that each baseline uses in its source paper, and the setting in which
we evaluate it. Every method in Table~\ref{tab:cross-model-fit7} is evaluated
as MXFP4 W4A4 under the same contract.

\begin{table}[htbp]
\centering\revcolor
\caption{Baseline formats in their source papers and in our evaluation.}
\label{tab:baseline-formats}
\small\setlength{\tabcolsep}{5pt}
\begin{tabular}{lll}
\toprule
Method & Format in source paper & Evaluated here \\
\midrule
RTN & format-agnostic & MXFP4 W4A4 \\
GPTQ \citep{frantar2023gptq} & integer grid, weight-only & MXFP4 W4A4 \\
SpinQuant+GPTQ \citep{liu2025spinquant} & INT, W4A4KV4 and W4A8KV8 & MXFP4 W4A4 \\
MR-GPTQ \citep{cook2025mrgptq} & MXFP4 / NVFP4, W4A4 & MXFP4 W4A4 \\
BRQ \citep{shao2025brq} & MXFP4, W4A4 & MXFP4 W4A4 \\
WUSH \citep{chen2026wush} & MXFP4 / NVFP4, W4A4 & MXFP4 W4A4 \\
TORQ \citep{xu2026torq} & MXFP4, W4A4 & MXFP4 W4A4 \\
BATQuant+GPTQ \citep{li2026batquant} & MXFP4, W4A4 & MXFP4 W4A4 \\
FOCUS \citep{yan2026focus} & NVFP4 / MXFP4, W4A4 & MXFP4 W4A4 \\
\bottomrule
\end{tabular}
\end{table}
}

\paragraph{MR-GPTQ.}
We adapt the official FP-Quant implementation \citep{cook2025mrgptq}, pinned
at \texttt{239c4168}. It retains its transform and intrinsic GPTQ
reconstruction, with encoding and decoding adapted to our E2M1/E8M0
contract. For OLMo, the loader divides the decoded exported scale
by four to recover the intended scale; the codes and calibration-built
artifact are unchanged. We reevaluate all seven tasks with this loader.

\paragraph{WUSH.}
We adapt the official implementation \citep{chen2026wush}, pinned at
\texttt{0b69f66c}, preserving its transform and intrinsic GPTQ reconstruction.
Encoding and decoding follow our E2M1/E8M0 contract.

\paragraph{BRQ.}
BRQ \citep{shao2025brq} uses an official-code direct-core adaptation
(\texttt{0e4ab0b9}): normalized Sylvester $H_{32}$ transforms on consecutive
input-feature blocks, followed by strict GPTQ. This port fixes the block
transform; the paper's stochastic-sign, sharing, and RNG choices remain
unresolved. The short table label includes the GPTQ reconstruction step.

\paragraph{TORQ.}
TORQ \citep{xu2026torq} uses our independent reimplementation
of its two-level inter/intra-block rotation, followed by strict GPTQ.
The short table label includes this reconstruction step.

\paragraph{BATQuant+GPTQ.}
Our independent BATQuant reimplementation \citep{li2026batquant} learns
GPK transforms and clipping parameters for 160 optimizer steps per
transformation group, then applies strict GPTQ. It uses the common C4
calibration cache in place of the paper's self-generated calibration data.
GPTQ reconstructs the transformed weights once, with damping $0.01$;
all seven tasks restore the resulting payload with dynamic A4.

\paragraph{SpinQuant+GPTQ.}
We use the Brevitas implementation of learned SpinQuant rotations
\citep{liu2025spinquant}, pinned at \texttt{6ba27e0b}, fuse them into the
float model, and apply one strict GPTQ pass. On OLMo2's post-norm topology,
the port uses paired V/O Hadamard rotations across attention, optimized for
100 steps with global batch size eight. The fused float model preserves
Base logits to relative RMS error $5.29\times10^{-7}$; GPTQ then reconstructs
all 280 target projections. The OLMo row reports this topology-safe paired
adaptation, rather than the original Llama residual-rotation recipe.
For these compositions, \texttt{+GPTQ} denotes terminal weight reconstruction,
not a second quantization of an already-built payload.

\paragraph{FOCUS.}
We use the official AngelSlim implementation of FOCUS \citep{yan2026focus},
with coupled-relaxation scaling and four 8-value subgroups per physical
Block32. Optimization keeps BF16 weights frozen and learns scales for one
epoch with global batch size 32. Scale and subgroup-relaxation learning
rates are $0.02$ and $0.05$, respectively; the distillation loss uses the
top 1,000 teacher logits. Activation quantization remains dynamic MXFP4
without learned activation scales. Export retains FP4 codes and E8M0
dequantization scales, discarding subgroup-relaxation parameters.
We replace the public example's calibration data with our frozen C4
$128\times2048$ token cache and adapt module registration for the
architectures beyond its Qwen3-4B example. The quantization algorithm is unchanged.
All seven tasks evaluate the same exported artifact per model.
FOCUS improves on GPTQ's Fit-7 across all seven models, with Robust Fit
$0.14718$, compared with $0.18624$ for GPTQ and $0.09300$ for \HiRes.

Tables~\ref{tab:raw-llama3b}--\ref{tab:qwen32-transfer} report every raw endpoint
used in the seven-model comparison. Accuracy values are percentages, except
for the fractions in Table~\ref{tab:qwen32-transfer}.
On Llama-3-8B, \HiRes lowers WikiText-2/C4 perplexity from
WUSH's $7.133/11.117$ to $7.111/11.026$; on Mistral-7B, the corresponding
values fall from $5.689/8.834$ to $5.664/8.810$.

\begin{table}[htbp]
\centering
\caption{Complete seven-task results on Llama-3.2-3B Base.}
\label{tab:raw-llama3b}
\small\setlength{\tabcolsep}{3.2pt}
\begin{tabular}{lrrrrrrrr}
\toprule
Method & WT2$\downarrow$ & C4$\downarrow$ & ARC-C$\uparrow$ & ARC-E$\uparrow$ & HS$\uparrow$ & PIQA$\uparrow$ & Wino$\uparrow$ & Fit-7$\downarrow$ \\
\midrule
BF16 & 7.818 & 11.216 & 41.47 & 70.50 & 72.79 & 77.58 & 61.17 & .0000 \\
RTN & 10.266 & 15.189 & 35.79 & 61.83 & 65.94 & 72.42 & 57.30 & .2594 \\
GPTQ & 9.768 & 14.177 & 37.12 & 62.71 & 67.04 & 74.10 & 57.46 & .1973 \\
MR-GPTQ & 9.011 & 13.333 & 36.45 & 67.30 & 68.67 & 76.61 & 58.48 & .1387 \\
BRQ & 8.969 & 13.182 & 39.46 & 67.63 & 68.39 & 75.90 & 58.56 & .1260 \\
WUSH & 8.905 & 13.056 & 37.46 & 65.87 & 69.14 & 76.28 & 59.19 & .1224 \\
TORQ & 10.558 & 15.597 & 31.77 & 57.87 & 63.94 & 71.55 & 57.46 & .2964 \\
BATQuant+GPTQ & 10.429 & 15.255 & 36.12 & 61.95 & 64.96 & 73.88 & 56.12 & .2645 \\
SpinQuant+GPTQ & 10.867 & 16.058 & 33.44 & 54.34 & 65.22 & 75.35 & 57.06 & .3191 \\
FOCUS & 9.481 & 14.035 & 36.79 & 65.87 & 68.62 & 74.76 & 59.51 & .1803 \\
\midrule
\textbf{\HiRes} & 8.889 & 13.045 & 38.80 & 68.56 & 69.30 & 76.01 & 58.64 & .1174 \\
\bottomrule
\end{tabular}
\end{table}

\begin{table}[htbp]
\centering
\caption{Complete seven-task results on Qwen3-4B.}
\label{tab:raw-qwen4b}
\small\setlength{\tabcolsep}{3.2pt}
\begin{tabular}{lrrrrrrrr}
\toprule
Method & WT2$\downarrow$ & C4$\downarrow$ & ARC-C$\uparrow$ & ARC-E$\uparrow$ & HS$\uparrow$ & PIQA$\uparrow$ & Wino$\uparrow$ & Fit-7$\downarrow$ \\
\midrule
BF16 & 13.661 & 19.824 & 47.16 & 77.15 & 66.73 & 74.86 & 57.22 & .0000 \\
RTN & 18.182 & 24.181 & 42.47 & 69.65 & 62.12 & 71.55 & 56.12 & .2283 \\
GPTQ & 16.013 & 22.538 & 43.48 & 72.26 & 62.56 & 73.23 & 56.51 & .1252 \\
MR-GPTQ & 15.528 & 21.981 & 46.82 & 71.89 & 62.93 & 73.12 & 55.96 & .0986 \\
BRQ & 15.104 & 21.455 & 47.16 & 72.43 & 63.30 & 73.94 & 55.96 & .0767 \\
WUSH & 14.269 & 21.132 & 47.83 & 73.36 & 63.37 & 72.96 & 57.93 & .0479 \\
TORQ & 16.998 & 24.886 & 41.47 & 67.42 & 59.81 & 71.33 & 56.27 & .1929 \\
BATQuant+GPTQ & 17.436 & 24.937 & 42.81 & 68.27 & 59.83 & 69.75 & 55.88 & .2051 \\
SpinQuant+GPTQ & 14.526 & 22.073 & 39.46 & 63.13 & 59.63 & 72.25 & 55.01 & .1409 \\
FOCUS & 13.651 & 21.324 & 45.82 & 68.94 & 63.98 & 73.12 & 56.43 & .0738 \\
\midrule
\textbf{\HiRes} & 14.740 & 21.117 & 48.83 & 74.79 & 63.71 & 74.65 & 57.62 & .0524 \\
\bottomrule
\end{tabular}
\end{table}

\begin{table}[htbp]
\centering
\caption{Complete seven-task results on Mistral-7B-v0.3.}
\label{tab:raw-mistral7b}
\small\setlength{\tabcolsep}{3.2pt}
\begin{tabular}{lrrrrrrrr}
\toprule
Method & WT2$\downarrow$ & C4$\downarrow$ & ARC-C$\uparrow$ & ARC-E$\uparrow$ & HS$\uparrow$ & PIQA$\uparrow$ & Wino$\uparrow$ & Fit-7$\downarrow$ \\
\midrule
BF16 & 5.318 & 8.439 & 45.48 & 73.53 & 79.60 & 81.28 & 70.40 & .0000 \\
RTN & 6.392 & 9.898 & 41.47 & 68.94 & 74.94 & 78.78 & 64.09 & .1512 \\
GPTQ & 6.090 & 9.334 & 44.48 & 69.65 & 75.40 & 79.43 & 64.64 & .1071 \\
MR-GPTQ & 5.731 & 8.919 & 45.48 & 73.11 & 77.17 & 80.58 & 67.88 & .0541 \\
BRQ & 5.734 & 8.923 & 44.48 & 71.76 & 77.28 & 80.20 & 68.19 & .0573 \\
WUSH & 5.689 & 8.834 & 44.82 & 72.35 & 77.20 & 79.87 & 68.82 & .0503 \\
TORQ & 6.120 & 9.412 & 39.13 & 68.18 & 75.20 & 79.43 & 63.93 & .1217 \\
BATQuant+GPTQ & 6.366 & 9.775 & 40.47 & 69.07 & 74.13 & 79.43 & 64.88 & .1482 \\
SpinQuant+GPTQ & 6.075 & 9.299 & 39.13 & 67.80 & 76.14 & 80.09 & 64.56 & .1142 \\
FOCUS & 5.831 & 9.070 & 42.14 & 71.55 & 77.67 & 80.20 & 65.82 & .0749 \\
\midrule
\textbf{\HiRes} & 5.664 & 8.810 & 46.15 & 73.57 & 77.56 & 80.25 & 67.17 & .0452 \\
\bottomrule
\end{tabular}
\end{table}

\begin{table}[htbp]
\centering
\caption{Complete seven-task results on Qwen3-8B.}
\label{tab:raw-qwen8b}
\small\setlength{\tabcolsep}{3.2pt}
\begin{tabular}{lrrrrrrrr}
\toprule
Method & WT2$\downarrow$ & C4$\downarrow$ & ARC-C$\uparrow$ & ARC-E$\uparrow$ & HS$\uparrow$ & PIQA$\uparrow$ & Wino$\uparrow$ & Fit-7$\downarrow$ \\
\midrule
BF16 & 9.727 & 15.665 & 54.85 & 80.43 & 73.46 & 77.69 & 58.01 & .0000 \\
RTN & 11.834 & 18.829 & 46.82 & 72.35 & 67.85 & 74.48 & 56.12 & .1666 \\
GPTQ & 11.554 & 18.290 & 47.16 & 75.04 & 68.47 & 74.86 & 58.41 & .1411 \\
MR-GPTQ & 10.522 & 16.831 & 53.85 & 79.63 & 70.22 & 75.57 & 58.41 & .0587 \\
BRQ & 10.701 & 16.906 & 54.85 & 79.50 & 70.53 & 76.39 & 57.38 & .0685 \\
WUSH & 10.419 & 16.594 & 53.85 & 80.81 & 71.25 & 76.22 & 57.46 & .0500 \\
TORQ & 12.357 & 19.117 & 46.15 & 74.71 & 67.31 & 75.41 & 56.20 & .1970 \\
BATQuant+GPTQ & 11.666 & 18.472 & 46.82 & 74.83 & 68.07 & 74.81 & 58.17 & .1498 \\
SpinQuant+GPTQ & 10.742 & 17.236 & 51.84 & 75.42 & 67.61 & 75.79 & 57.30 & .0835 \\
FOCUS & 10.423 & 17.191 & 46.49 & 74.03 & 70.67 & 75.19 & 58.33 & .1095 \\
\midrule
\textbf{\HiRes} & 10.055 & 16.299 & 54.85 & 77.61 & 70.94 & 76.33 & 57.46 & .0324 \\
\bottomrule
\end{tabular}
\end{table}

\begin{table}[htbp]
\centering
\caption{Complete seven-task results on Llama-3-8B Base.}
\label{tab:raw-llama8b}
\small\setlength{\tabcolsep}{3.2pt}
\begin{tabular}{lrrrrrrrr}
\toprule
Method & WT2$\downarrow$ & C4$\downarrow$ & ARC-C$\uparrow$ & ARC-E$\uparrow$ & HS$\uparrow$ & PIQA$\uparrow$ & Wino$\uparrow$ & Fit-7$\downarrow$ \\
\midrule
BF16 & 6.138 & 9.597 & 53.18 & 76.81 & 78.50 & 80.52 & 62.83 & .0000 \\
RTN & 8.236 & 12.650 & 41.81 & 69.91 & 72.37 & 76.55 & 59.91 & .2522 \\
GPTQ & 7.962 & 12.148 & 40.47 & 66.29 & 73.25 & 75.63 & 59.75 & .2283 \\
MR-GPTQ & 7.249 & 11.324 & 45.15 & 71.21 & 75.07 & 77.80 & 61.40 & .1394 \\
BRQ & 7.260 & 11.245 & 45.15 & 72.52 & 75.69 & 78.94 & 59.98 & .1386 \\
WUSH & 7.133 & 11.117 & 44.82 & 73.32 & 75.38 & 78.35 & 61.40 & .1248 \\
TORQ & 8.253 & 13.060 & 44.15 & 65.53 & 72.88 & 75.03 & 61.09 & .2654 \\
BATQuant+GPTQ & 8.427 & 12.999 & 44.15 & 67.97 & 71.16 & 76.22 & 59.75 & .2727 \\
SpinQuant+GPTQ & 8.383 & 13.125 & 46.15 & 71.55 & 73.27 & 77.42 & 62.12 & .2588 \\
FOCUS & 7.628 & 11.790 & 45.82 & 74.24 & 75.88 & 77.37 & 61.72 & .1737 \\
\midrule
\textbf{\HiRes} & 7.111 & 11.026 & 47.49 & 72.47 & 75.24 & 79.87 & 60.77 & .1187 \\
\bottomrule
\end{tabular}
\end{table}

\paragraph{OLMo-2-13B transfer.}
We evaluate OLMo-2-1124-13B Base \citep{olmo2024olmo2} after freezing
\HiRes, extending the test to a new architecture and model family at 13B
parameters. Its results and diagnostics do not enter algorithm design or
hyperparameter selection. The calibration and evaluation protocol is
unchanged, including disabled KV caching. \HiRes achieves the lowest
quantized Fit-7 and C4 perplexity, and the highest accuracy on
ARC-Challenge, ARC-Easy, HellaSwag, and PIQA
(Table~\ref{tab:raw-olmo13b}).

\begin{table}[htbp]
\centering
\caption{Complete seven-task results on OLMo-2-1124-13B Base with the frozen
algorithm. Accuracy values are percentages; bold marks the best quantized
value in each column. {\revcolor$\ddagger$}: topology-safe paired V/O SpinQuant+GPTQ.}
\label{tab:raw-olmo13b}
\small\setlength{\tabcolsep}{3.2pt}
\begin{tabular}{lrrrrrrrr}
\toprule
Method & WT2$\downarrow$ & C4$\downarrow$ & ARC-C$\uparrow$ & ARC-E$\uparrow$ & HS$\uparrow$ & PIQA$\uparrow$ & Wino$\uparrow$ & Fit-7$\downarrow$ \\
\midrule
BF16 & 5.003 & 10.887 & 52.84 & 80.01 & 82.95 & 81.77 & 69.85 & .0000 \\
RTN & 6.043 & 11.842 & 48.16 & 79.00 & 81.02 & 80.14 & 69.22 & .1360 \\
GPTQ & 6.060 & 11.787 & 50.50 & 78.41 & 80.68 & 80.09 & 68.51 & .1360 \\
MR-GPTQ & \textbf{5.657} & 11.368 & 49.50 & 79.59 & 81.25 & 80.36 & \textbf{70.72} & .0846 \\
BRQ & 5.672 & 11.389 & 52.84 & 81.19 & 81.12 & 81.01 & 69.69 & .0811 \\
WUSH & 5.699 & 11.376 & 54.52 & 80.85 & 81.49 & 80.25 & 69.85 & .0823 \\
TORQ & 5.852 & 11.588 & 52.17 & 80.30 & 80.90 & 80.52 & 68.19 & .1067 \\
BATQuant+GPTQ & 6.003 & 11.785 & 48.16 & 78.66 & 81.20 & 79.98 & 69.46 & .1311 \\
SpinQuant+GPTQ{\revcolor$^{\ddagger}$} & 6.033 & 11.809 & 50.50 & 80.30 & 80.95 & 80.79 & 67.40 & .1317 \\
FOCUS & 5.911 & 11.647 & 50.84 & 78.87 & 79.63 & 79.54 & 67.64 & .1195 \\
\midrule
\textbf{\HiRes} & 5.658 & \textbf{11.339} & \textbf{54.85} & \textbf{81.31} & \textbf{81.66} & \textbf{81.12} & 68.19 & \textbf{.0773} \\
\bottomrule
\end{tabular}
\end{table}

\paragraph{Qwen3-32B scale transfer.}
The complete comparison includes BF16 and ten quantized methods under
strict OCP MXFP4 W4A4 with E2M1 values, E8M0 shared scales, and
contiguous block size 32. KV caching is disabled during evaluation.
Calibration uses C4-train, seed~0, and exactly $128\times2048$ tokens.
\HiRes achieves the lowest quantized Fit-7 ($0.0274$), followed by
WUSH ($0.0365$) and BRQ ($0.0425$). Among quantized methods, it leads
on five of seven tasks: WikiText-2/C4 perplexity ($7.926/13.045$),
ARC-E, HellaSwag, and PIQA. Its Fit-7 is $25.0\%$ lower than WUSH's
(Table~\ref{tab:qwen32-transfer}).

\begin{table}[htbp]
  \centering
  \caption{Complete Qwen3-32B seven-task results under strict MXFP4
  W4A4 with KV caching disabled. Accuracy values are fractions. Quantized methods are ordered
  by Fit-7; bold marks the best quantized value in each column.}
  \label{tab:qwen32-transfer}
  \small
  \setlength{\tabcolsep}{3pt}
  \begin{tabular}{lrrrrrrrr}
    \toprule
    Method & WT2$\downarrow$ & C4$\downarrow$ & ARC-C$\uparrow$ & ARC-E$\uparrow$ & HS$\uparrow$ & PIQA$\uparrow$ & Wino$\uparrow$ & Fit-7$\downarrow$ \\
    \midrule
    BF16 & 7.611 & 12.689 & 0.5953 & 0.8262 & 0.8157 & 0.8199 & 0.6140 & 0.0000 \\
    \midrule
    \textbf{\HiRes} & \textbf{7.926} & \textbf{13.045} & 0.5753 & \textbf{0.8321} & \textbf{0.8104} & \textbf{0.8150} & 0.6235 & \textbf{0.0274} \\
    WUSH & 7.950 & 13.119 & 0.5652 & 0.8169 & 0.8041 & 0.8041 & 0.6243 & 0.0365 \\
    BRQ & 8.138 & 13.301 & \textbf{0.6054} & 0.8199 & 0.8019 & 0.8074 & \textbf{0.6330} & 0.0425 \\
    MR-GPTQ & 8.196 & 13.485 & 0.5819 & 0.8308 & 0.7992 & 0.7862 & 0.6275 & 0.0524 \\
    FOCUS & 8.141 & 13.910 & 0.5819 & 0.8152 & 0.7895 & 0.7927 & 0.6140 & 0.0671 \\
    GPTQ & 8.420 & 13.763 & 0.5786 & 0.8018 & 0.7935 & 0.8047 & 0.6101 & 0.0746 \\
    RTN & 8.614 & 14.147 & 0.5686 & 0.8131 & 0.7932 & 0.7965 & 0.6140 & 0.0919 \\
    TORQ & 8.667 & 14.188 & 0.5953 & 0.8051 & 0.7978 & 0.7905 & 0.6212 & 0.0929 \\
    SpinQuant+GPTQ & 8.416 & 13.845 & 0.5151 & 0.7837 & 0.7745 & 0.7911 & 0.6022 & 0.1023 \\
    BATQuant+GPTQ & 8.973 & 14.627 & 0.5753 & 0.7963 & 0.7894 & 0.7927 & 0.6243 & 0.1217 \\
    \bottomrule
  \end{tabular}
\end{table}

\subsection{Calibration and Evaluation Sensitivity}
\label{app:calibration-sensitivity}

\paragraph{Calibration seeds.}
We repeat calibration with seeds $\{0,3,5\}$ on Qwen3-8B and
Meta-Llama-3-8B, abbreviated as Qwen and Llama in the tables below.
Seed~0 matches the main comparison. Each configuration collects its own
recovery statistics and reruns downstream evaluation. Table captions
distinguish Fit-3 (WT2, C4, ARC-C; Eq.~\ref{eq:fit3}) from Fit-7
(Eq.~\ref{eq:fit7}); both use the BF16 references in
Table~\ref{tab:ablation-bf16}.

Before running the comparisons, we specified minimum differences of
$0.002$ for Fit-7 and $0.003$ for Fit-3. These thresholds apply to each hypothesis's designated
comparison between the full method and its strongest control.
Intervals use the $n=3$ $t$ distribution, with $t_{0.975,2}=4.303$;
paired directions and intervals are reported together.
Accuracy sample counts are 299 for ARC-C, 2376 for ARC-E, 10042 for
HellaSwag, 1838 for PIQA, and 1267 for WinoGrande.

\begin{table}[htbp]
\centering
\caption{BF16 references for the multi-seed ablations. Accuracy is reported as a fraction.}
\label{tab:ablation-bf16}
\small\setlength{\tabcolsep}{4pt}
\begin{tabular}{lrrrrrrr}
\toprule
Model & WT2$\downarrow$ & C4$\downarrow$ & ARC-C$\uparrow$ & ARC-E$\uparrow$ & Hella$\uparrow$ & PIQA$\uparrow$ & Wino$\uparrow$ \\
\midrule
Qwen & 9.727 & 15.665 & 0.5485 & 0.8043 & 0.7346 & 0.7769 & 0.5801 \\
Llama & 6.138 & 9.597 & 0.5318 & 0.7681 & 0.7850 & 0.8052 & 0.6283 \\
\bottomrule
\end{tabular}
\end{table}

\HiRes improves on WUSH across all $3/3$ calibration seeds on both models
(Table~\ref{tab:calib-seed-stability}). On Qwen3-8B, mean$\pm$SD Fit-7 is
$0.0327\pm0.0003$ for \HiRes versus $0.0498\pm0.0029$ for WUSH
(mean paired gain $-0.0171\pm0.0028$).
On Llama-3-8B, the corresponding values are $0.1162\pm0.0040$ versus
$0.1216\pm0.0036$ (mean paired gain $-0.0055\pm0.0011$).
Per-seed paired differences are listed in Table~\ref{tab:geometry-paired}.
\begin{table}[htbp]
\centering
\caption{Calibration-seed stability under the seven-task protocol. Accuracy columns are fractions; lower Fit-7 is better.}
\label{tab:calib-seed-stability}
\small\setlength{\tabcolsep}{4pt}
\begin{tabular}{llrrrrr}
\toprule
Model & Method & Seed & WT2$\downarrow$ & C4$\downarrow$ & ARC-C$\uparrow$ & ARC-E$\uparrow$ \\
\midrule
Qwen & \HiRes & 0 & 10.055 & 16.299 & 0.5485 & 0.7761 \\
Qwen & \HiRes & 3 & 10.160 & 16.315 & 0.5351 & 0.7963 \\
Qwen & \HiRes & 5 & 10.150 & 16.306 & 0.5485 & 0.7744 \\
Qwen & WUSH & 0 & 10.419 & 16.594 & 0.5385 & 0.8081 \\
Qwen & WUSH & 3 & 10.378 & 16.552 & 0.5418 & 0.7896 \\
Qwen & WUSH & 5 & 10.446 & 16.615 & 0.5418 & 0.8026 \\
\midrule
Llama & \HiRes & 0 & 7.111 & 11.026 & 0.4749 & 0.7247 \\
Llama & \HiRes & 3 & 7.115 & 10.994 & 0.4649 & 0.7517 \\
Llama & \HiRes & 5 & 7.101 & 10.988 & 0.4983 & 0.7555 \\
Llama & WUSH & 0 & 7.133 & 11.117 & 0.4482 & 0.7332 \\
Llama & WUSH & 3 & 7.138 & 11.097 & 0.4615 & 0.7378 \\
Llama & WUSH & 5 & 7.131 & 11.072 & 0.4749 & 0.7609 \\
\bottomrule
\end{tabular}\par\medskip
\begin{tabular}{llrrrrr}
\toprule
Model & Method & Seed & HellaSwag$\uparrow$ & PIQA$\uparrow$ & WinoGrande$\uparrow$ & Fit-7$\downarrow$ \\
\midrule
Qwen & \HiRes & 0 & 0.7094 & 0.7633 & 0.5746 & 0.0324 \\
Qwen & \HiRes & 3 & 0.7115 & 0.7797 & 0.5817 & 0.0327 \\
Qwen & \HiRes & 5 & 0.7129 & 0.7622 & 0.5872 & 0.0330 \\
Qwen & WUSH & 0 & 0.7125 & 0.7622 & 0.5746 & 0.0500 \\
Qwen & WUSH & 3 & 0.7109 & 0.7606 & 0.5927 & 0.0468 \\
Qwen & WUSH & 5 & 0.7043 & 0.7628 & 0.5746 & 0.0525 \\
\midrule
Llama & \HiRes & 0 & 0.7524 & 0.7987 & 0.6077 & 0.1187 \\
Llama & \HiRes & 3 & 0.7507 & 0.7916 & 0.6109 & 0.1182 \\
Llama & \HiRes & 5 & 0.7544 & 0.7987 & 0.6062 & 0.1115 \\
Llama & WUSH & 0 & 0.7538 & 0.7835 & 0.6140 & 0.1248 \\
Llama & WUSH & 3 & 0.7513 & 0.7943 & 0.6117 & 0.1224 \\
Llama & WUSH & 5 & 0.7539 & 0.7900 & 0.6117 & 0.1178 \\
\bottomrule
\end{tabular}
\end{table}

\paragraph{Alternative summaries.}
We also summarize the seven-task endpoints by their mean relative damage
and the equally weighted mean of the five accuracy tasks:
\begin{equation}
\operatorname{MeanDamage}_m=\frac17\sum_{j=1}^7 d_{mj},\qquad
\operatorname{Acc5}_m=\frac{100}{5}\sum_{j\in\mathcal A}
\operatorname{Acc}_{mj},
\label{eq:alternative-aggregates}
\end{equation}
where $\mathcal A$ contains ARC-C, ARC-E, HellaSwag, PIQA, and WinoGrande,
and accuracies inside the sum are fractions.
Damage follows Eq.~\ref{eq:fit7}, including negative values for gains over
BF16. Fit-7 retains the reported table values; the alternative aggregates
are recomputed from the same seven-task endpoints.

\HiRes lowers MeanDamage on all seven models
(Table~\ref{tab:alternative-aggregates}).
On Qwen3-8B, WT2, C4, and Fit-7 improve while Acc5 is slightly lower.
The completed Qwen3-32B evaluation gives MeanDamage $0.0224$ for WUSH
and $0.0133$ for \HiRes.
Their worst-task damages are $0.0506$ on ARC-C and $0.0414$ on WT2,
respectively; the seven endpoints appear in Table~\ref{tab:qwen32-transfer}.

\begin{table}[htbp]
\centering
\caption{Alternative aggregates from the frozen seven-task evaluations at
calibration seed~0. All results use the complete 299-question ARC-C
validation split.}
\label{tab:alternative-aggregates}
\small\setlength{\tabcolsep}{4pt}
\begin{tabular}{llrrr}
\toprule
Model & Method & Fit-7$\downarrow$ & MeanDamage$\downarrow$ & Acc5 (\%)$\uparrow$ \\
\midrule
Llama-3.2-3B & WUSH & 0.1224 & 0.0807 & 61.5880 \\
& \HiRes & 0.1174 & 0.0716 & 62.2620 \\
Qwen3-4B & WUSH & 0.0479 & 0.0298 & 63.0900 \\
& \HiRes & 0.0524 & 0.0258 & 63.9200 \\
Mistral-7B & WUSH & 0.0503 & 0.0310 & 68.6120 \\
& \HiRes & 0.0452 & 0.0254 & 68.9400 \\
\midrule
Qwen3-8B & WUSH & 0.0500 & 0.0289 & 67.9176 \\
& \HiRes & 0.0324 & 0.0244 & 67.4385 \\
Llama-3-8B & WUSH & 0.1248 & 0.0875 & 66.6534 \\
& \HiRes & 0.1187 & 0.0790 & 67.1706 \\
OLMo-2-13B & WUSH & 0.0823 & 0.0254 & 73.3907 \\
& \HiRes & 0.0773 & 0.0236 & 73.4266 \\
\midrule
Qwen3-32B & WUSH & 0.0365 & 0.0224 & 72.2941 \\
& \HiRes & 0.0274 & 0.0133 & 73.1251 \\
\bottomrule
\end{tabular}
\end{table}

\subsection{Stage and Geometry Ablations}
\label{app:hires-ablation}

We first add the three stages in deployment order, then isolate the
geometric corrections at fixed downstream procedures. The stage comparison
in Table~\ref{tab:stage-progression-raw} reports Fit-3 on Qwen3-8B and
Meta-Llama-3-8B. Each stage lowers damage on both models; when Stage~II is
disabled (``+ Geometric shaping''), Stage~I's diagonal-balance trial evaluates
re-encoding loss under that row's active RTN/nearest-A4 operators rather than
invoking GPTQ or one-code refinement.

\begin{table}[htbp]
\centering
\caption{Post-discovery stage ablations with raw endpoints. Each row retains the preceding
stages. The Discrete row corresponds to the configuration with both structural
recovery paths disabled.}
\label{tab:stage-progression-raw}
\small\setlength{\tabcolsep}{4pt}
\begin{tabular}{llrrrrrr}
\toprule
Model & Program & WT2$\downarrow$ & C4$\downarrow$ & ARC-C$\uparrow$ & Mean dmg.$\downarrow$ & Worst dmg.$\downarrow$ & Fit-3$\downarrow$ \\
\midrule
Qwen3-8B & Strict RTN & 11.8342 & 18.8285 & 46.82 & .1883 & .2167 & .2025 \\
& + Geometric shaping & 10.5374 & 16.7494 & 51.17 & .0732 & .0833 & .0783 \\
& + Discrete realization & 10.1858 & 16.4524 & 50.84 & .0569 & .0732 & .0650 \\
& + Structural recovery & 10.0552 & 16.2987 & 54.85 & .0247 & .0404 & .0326 \\
\midrule
Llama-3-8B & Strict RTN & 8.2362 & 12.6497 & 41.81 & .2912 & .3418 & .3165 \\
& + Geometric shaping & 7.1963 & 11.2766 & 46.15 & .1598 & .1750 & .1674 \\
& + Discrete realization & 7.1417 & 11.1163 & 46.82 & .1471 & .1635 & .1553 \\
& + Structural recovery & 7.1110 & 11.0262 & 47.49 & .1381 & .1585 & .1483 \\
\bottomrule
\end{tabular}
\end{table}

\begin{table}[htbp]
\centering
\caption{Geometric-component ablations at calibration seed 0. Downstream reconstruction and recovery procedures are fixed.}
\label{tab:component-ablations}
\small\setlength{\tabcolsep}{4pt}
\begin{tabular}{llrrrr}
\toprule
Model & Configuration & WT2$\downarrow$ & C4$\downarrow$ & ARC-C$\uparrow$ & Fit-3$\downarrow$ \\
\midrule
Qwen3-8B & Full \HiRes & 10.0552 & 16.2987 & 54.85 & .0326 \\
& Remove local interaction $K$ & 10.1215 & 16.3613 & 53.51 & .0405 \\
& $D=I$ & 10.1480 & 16.3925 & 53.18 & .0433 \\
& $K=0, D=I$ & 10.2436 & 16.5013 & 52.17 & .0526 \\
Llama-3-8B & Full \HiRes & 7.1110 & 11.0262 & 47.49 & .1483 \\
& Remove local interaction $K$ & 7.1393 & 11.0706 & 46.49 & .1553 \\
& $D=I$ & 7.1525 & 11.0931 & 45.82 & .1592 \\
& $K=0, D=I$ & 7.1813 & 11.1419 & 44.48 & .1674 \\
\bottomrule
\end{tabular}
\end{table}

The local interaction $K$ improves fitness on both models, by $0.0079$ on
Qwen3-8B and $0.0070$ on Llama-3-8B. Diagonal balance contributes on both as
well: removing $D$ degrades Qwen3-8B by $0.0107$ and Llama-3-8B by $0.0109$,
and removing both is worse than removing either alone.

Removing either geometric correction increases damage
(Table~\ref{tab:kd-multiseed}). In all $6/6$ model--seed pairs, the full
method is best, followed by removing $K$, removing $D$, and removing both.
Relative to removing $K$, the strongest control, the full method
lowers Fit-7 by $0.0053$--$0.0066$ on Qwen and
$0.0054$--$0.0061$ on Llama.
Both ranges exceed the $0.002$ threshold.
Removing both corrections is worse than removing either alone.
Each correction also lowers Fit-7 when used without the other.
\begin{table}[htbp]
\centering
\caption{Geometry ablations across calibration seeds. Reference denotes $K=0$, $D=I$. Each configuration refits recovery statistics on its own encoded state. Accuracy is reported as a fraction.}
\label{tab:kd-multiseed}
\small\setlength{\tabcolsep}{4pt}
\begin{tabular}{llrrrrr}
\toprule
Model & Config. & Seed & WT2$\downarrow$ & C4$\downarrow$ & ARC-C$\uparrow$ & ARC-E$\uparrow$ \\
\midrule
Qwen & Full & 0 & 10.055 & 16.299 & 0.5485 & 0.7761 \\
Qwen & $K=0$ & 0 & 10.121 & 16.361 & 0.5351 & 0.7736 \\
Qwen & $D=I$ & 0 & 10.148 & 16.393 & 0.5318 & 0.7719 \\
Qwen & Reference & 0 & 10.244 & 16.501 & 0.5217 & 0.7664 \\
\midrule
Qwen & Full & 3 & 10.160 & 16.315 & 0.5351 & 0.7963 \\
Qwen & $K=0$ & 3 & 10.225 & 16.376 & 0.5217 & 0.7938 \\
Qwen & $D=I$ & 3 & 10.244 & 16.400 & 0.5184 & 0.7925 \\
Qwen & Reference & 3 & 10.314 & 16.480 & 0.5151 & 0.7883 \\
\midrule
Qwen & Full & 5 & 10.150 & 16.306 & 0.5485 & 0.7744 \\
Qwen & $K=0$ & 5 & 10.208 & 16.361 & 0.5351 & 0.7723 \\
Qwen & $D=I$ & 5 & 10.231 & 16.387 & 0.5351 & 0.7706 \\
Qwen & Reference & 5 & 10.333 & 16.502 & 0.5217 & 0.7652 \\
\midrule
Llama & Full & 0 & 7.111 & 11.026 & 0.4749 & 0.7247 \\
Llama & $K=0$ & 0 & 7.139 & 11.071 & 0.4649 & 0.7197 \\
Llama & $D=I$ & 0 & 7.152 & 11.093 & 0.4582 & 0.7159 \\
Llama & Reference & 0 & 7.181 & 11.142 & 0.4448 & 0.7088 \\
\midrule
Llama & Full & 3 & 7.115 & 10.994 & 0.4649 & 0.7517 \\
Llama & $K=0$ & 3 & 7.147 & 11.044 & 0.4548 & 0.7458 \\
Llama & $D=I$ & 3 & 7.156 & 11.060 & 0.4482 & 0.7433 \\
Llama & Reference & 3 & 7.174 & 11.091 & 0.4415 & 0.7382 \\
\midrule
Llama & Full & 5 & 7.101 & 10.988 & 0.4983 & 0.7555 \\
Llama & $K=0$ & 5 & 7.128 & 11.029 & 0.4883 & 0.7508 \\
Llama & $D=I$ & 5 & 7.140 & 11.051 & 0.4816 & 0.7471 \\
Llama & Reference & 5 & 7.170 & 11.100 & 0.4682 & 0.7399 \\
\bottomrule
\end{tabular}
\end{table}

\begin{table}[htbp]
\centering
\vspace{\abovecaptionskip}
{\normalsize Table~\ref*{tab:kd-multiseed}: Geometry ablations across calibration seeds (continued).\par}
\vspace{\belowcaptionskip}
\small\setlength{\tabcolsep}{4pt}
\begin{tabular}{llrrrrr}
\toprule
Model & Config. & Seed & HellaSwag$\uparrow$ & PIQA$\uparrow$ & WinoGrande$\uparrow$ & Fit-7$\downarrow$ \\
\midrule
Qwen & Full & 0 & 0.7094 & 0.7633 & 0.5746 & 0.0324 \\
Qwen & $K=0$ & 0 & 0.7076 & 0.7606 & 0.5738 & 0.0377 \\
Qwen & $D=I$ & 0 & 0.7068 & 0.7590 & 0.5730 & 0.0399 \\
Qwen & Reference & 0 & 0.7044 & 0.7557 & 0.5691 & 0.0474 \\
\midrule
Qwen & Full & 3 & 0.7115 & 0.7797 & 0.5817 & 0.0327 \\
Qwen & $K=0$ & 3 & 0.7098 & 0.7769 & 0.5809 & 0.0392 \\
Qwen & $D=I$ & 3 & 0.7092 & 0.7758 & 0.5801 & 0.0421 \\
Qwen & Reference & 3 & 0.7074 & 0.7737 & 0.5770 & 0.0476 \\
\midrule
Qwen & Full & 5 & 0.7129 & 0.7622 & 0.5872 & 0.0330 \\
Qwen & $K=0$ & 5 & 0.7113 & 0.7601 & 0.5864 & 0.0391 \\
Qwen & $D=I$ & 5 & 0.7107 & 0.7584 & 0.5856 & 0.0410 \\
Qwen & Reference & 5 & 0.7081 & 0.7546 & 0.5817 & 0.0508 \\
\midrule
Llama & Full & 0 & 0.7524 & 0.7987 & 0.6077 & 0.1187 \\
Llama & $K=0$ & 0 & 0.7505 & 0.7954 & 0.6046 & 0.1243 \\
Llama & $D=I$ & 0 & 0.7491 & 0.7922 & 0.6022 & 0.1277 \\
Llama & Reference & 0 & 0.7462 & 0.7878 & 0.5983 & 0.1343 \\
\midrule
Llama & Full & 3 & 0.7507 & 0.7916 & 0.6109 & 0.1182 \\
Llama & $K=0$ & 3 & 0.7487 & 0.7878 & 0.6077 & 0.1243 \\
Llama & $D=I$ & 3 & 0.7475 & 0.7851 & 0.6054 & 0.1270 \\
Llama & Reference & 3 & 0.7456 & 0.7824 & 0.6030 & 0.1315 \\
\midrule
Llama & Full & 5 & 0.7544 & 0.7987 & 0.6062 & 0.1115 \\
Llama & $K=0$ & 5 & 0.7526 & 0.7954 & 0.6030 & 0.1169 \\
Llama & $D=I$ & 5 & 0.7512 & 0.7927 & 0.6006 & 0.1201 \\
Llama & Reference & 5 & 0.7483 & 0.7878 & 0.5967 & 0.1269 \\
\bottomrule
\end{tabular}
\end{table}

\paragraph{Geometry replacement.}
\label{app:geometry-replacement}
To separate geometry from downstream corrections,
we compare WUSH, WUSH+R, and the complete \HiRes on Qwen3-8B and
Meta-Llama-3-8B, using calibration seeds $0,3,5$.
WUSH+R uses WUSH geometry through a common transform interface, rebuilds
W4 in those coordinates, and applies the same one-code refinement and
attention-to-MLP recovery as \HiRes.
Both configurations use GPTQ damping $0.01$, FP32 reconstruction, and the
same $128\times2048$ C4 calibration sequences and order.
They share E2M1/E8M0 Block32 encoding, the one-code rule, and solver budgets.

Each geometry is followed by freshly fitted attention and then MLP
recovery, with two affine coefficients per KV head and three MLP
coefficients per layer; $\sigma_g$ is a separate scale statistic.
The comparison fixes refinement and recovery procedures while changing
the geometry and the encoded state it produces.

With refinement and recovery disabled, the common interface reproduces
WUSH in all six model--seed configurations: W4 codes and E8M0 exponents
match elementwise, the five accuracy tasks give identical correctness
vectors, and WT2, C4, and Fit-7 agree at reporting precision. GPTQ ordering,
calibration files, and data order are unchanged. We reuse these WUSH and
\HiRes results and add WUSH+R.

Adding refinement and recovery to WUSH lowers mean Fit-7 by $0.0108$
on Qwen and $0.0029$ on Llama.
With the same recovery capacity, \HiRes lowers it by another $0.0063$
and $0.0026$, respectively (Table~\ref{tab:geometry-replacement}).
Both steps improve Fit-7 in all three paired seeds on each model
(Table~\ref{tab:geometry-paired}).
Mean WT2 and C4 perplexity decrease at both steps on both models.
Mean accuracy across the five classification tasks decreases slightly
on Qwen and increases on Llama.

\begin{table}[htbp]
\centering
\caption{Geometry replacement with matched refinement and recovery.
WT2, C4, and Acc5 are arithmetic means over calibration seeds $0,3,5$.
Acc5 averages the five accuracy tasks equally within each seed before
averaging across seeds. Fit-7 is computed per seed; $\pm$ denotes sample SD.}
\label{tab:geometry-replacement}
\small\setlength{\tabcolsep}{4pt}
\begin{tabular}{llrrrr}
\toprule
Model & Method & Fit-7$\downarrow$ & WT2$\downarrow$ & C4$\downarrow$ & Acc5 (\%)$\uparrow$ \\
\midrule
Qwen & WUSH & $0.0498\pm0.0029$ & 10.414 & 16.587 & 67.85 \\
& WUSH+R & $0.0390\pm0.0022$ & 10.239 & 16.428 & 67.79 \\
& \HiRes & $0.0327\pm0.0003$ & 10.122 & 16.307 & 67.74 \\
\midrule
Llama & WUSH & $0.1216\pm0.0036$ & 7.134 & 11.095 & 67.21 \\
& WUSH+R & $0.1188\pm0.0033$ & 7.121 & 11.045 & 67.44 \\
& \HiRes & $0.1162\pm0.0040$ & 7.109 & 11.003 & 67.61 \\
\bottomrule
\end{tabular}
\end{table}

\begin{table}[htbp]
\centering
\caption{Paired Fit-7 differences for geometry replacement across three calibration seeds (negative values favor the first method).}
\label{tab:geometry-paired}
\small\setlength{\tabcolsep}{4pt}
\begin{tabular}{llrrrr}
\toprule
Model & Difference & Seed 0$\downarrow$ & Seed 3$\downarrow$ & Seed 5$\downarrow$ & Mean $\pm$ SD$\downarrow$ \\
\midrule
Qwen & WUSH+R $-$ WUSH & $-0.0125$ & $-0.0089$ & $-0.0110$ & $-0.0108\pm0.0018$ \\
& \HiRes $-$ WUSH+R & $-0.0052$ & $-0.0052$ & $-0.0086$ & $-0.0063\pm0.0020$ \\
& \HiRes $-$ WUSH & $-0.0176$ & $-0.0141$ & $-0.0195$ & $-0.0171\pm0.0028$ \\
\midrule
Llama & WUSH+R $-$ WUSH & $-0.0039$ & $-0.0020$ & $-0.0028$ & $-0.0029\pm0.0009$ \\
& \HiRes $-$ WUSH+R & $-0.0022$ & $-0.0022$ & $-0.0034$ & $-0.0026\pm0.0007$ \\
& \HiRes $-$ WUSH & $-0.0060$ & $-0.0042$ & $-0.0062$ & $-0.0055\pm0.0011$ \\
\bottomrule
\end{tabular}
\end{table}

\subsection{Activation-Code Refinement}
\label{app:code-refinement-ablation}

Sequential reconstruction and one-code refinement both improve Fit-3
(Table~\ref{tab:discrete-ablations}), with the larger reduction coming
from sequential reconstruction. We next isolate the activation-code rule
while retaining the same weight reconstruction.

\begin{table}[htbp]
\centering
\caption{Discrete-realization ablations at calibration seed 0. Geometry and structural recovery procedures are fixed.}
\label{tab:discrete-ablations}
\small\setlength{\tabcolsep}{4pt}
\begin{tabular}{llrrrr}
\toprule
Model & Configuration & WT2$\downarrow$ & C4$\downarrow$ & ARC-C$\uparrow$ & Fit-3$\downarrow$ \\
\midrule
Qwen3-8B & Strict RTN realization & 10.2749 & 16.4117 & 49.83 & .0783 \\
& Sequential reconstruction & 10.1817 & 16.3898 & 53.18 & .0440 \\
& Full discrete realization & 10.0552 & 16.2987 & 54.85 & .0326 \\
Llama-3-8B & Strict RTN realization & 7.2654 & 11.3159 & 43.48 & .1827 \\
& Sequential reconstruction & 7.1464 & 11.0486 & 47.16 & .1536 \\
& Full discrete realization & 7.1110 & 11.0262 & 47.49 & .1483 \\
\bottomrule
\end{tabular}
\end{table}

We compare nearest rounding, sensitivity-only refinement, and the full
mismatch-aware rule from Appendix~\ref{app:hires-code-refinement}.
Payload format, geometry, weight reconstruction, and structural recovery
are fixed. Nearest rounding corresponds to the sequential-reconstruction
row in Table~\ref{tab:discrete-ablations}; the full rule gives \HiRes.

The full rule has lower Fit-7 than both controls in all $6/6$
model--seed pairs (Table~\ref{tab:onecode-rules}).
Its mean gain over nearest rounding and sensitivity-only is, respectively,
$0.0066$ and $0.0102$ on Qwen, and $0.0055$ and $0.0076$ on Llama.
The gains range over $0.0055$--$0.0109$ on Qwen and
$0.0050$--$0.0081$ on Llama.
All endpoint directions also favor the full rule across the 12 configuration
comparisons, covering 84 task-level differences across the paired seeds.
\begin{table}[htbp]
\centering
\caption{One-code rule ablations with the same payload format and remaining pipeline. Nearest uses nearest rounding; Sensitivity omits the realized-weight mismatch term; Full includes it.}
\label{tab:onecode-rules}
\small\setlength{\tabcolsep}{4pt}
\begin{tabular}{llrrrrr}
\toprule
Model & Method & Seed & WT2$\downarrow$ & C4$\downarrow$ & ARC-C$\uparrow$ & ARC-E$\uparrow$ \\
\midrule
Qwen & Nearest & 0 & 10.182 & 16.390 & 0.5318 & 0.7744 \\
Qwen & Sensitivity & 0 & 10.234 & 16.416 & 0.5418 & 0.7727 \\
Qwen & Full & 0 & 10.055 & 16.299 & 0.5485 & 0.7761 \\
\midrule
Qwen & Nearest & 3 & 10.234 & 16.369 & 0.5251 & 0.7950 \\
Qwen & Sensitivity & 3 & 10.308 & 16.412 & 0.5284 & 0.7938 \\
Qwen & Full & 3 & 10.160 & 16.315 & 0.5351 & 0.7963 \\
\midrule
Qwen & Nearest & 5 & 10.213 & 16.352 & 0.5418 & 0.7731 \\
Qwen & Sensitivity & 5 & 10.282 & 16.393 & 0.5418 & 0.7719 \\
Qwen & Full & 5 & 10.150 & 16.306 & 0.5485 & 0.7744 \\
\midrule
Llama & Nearest & 0 & 7.146 & 11.049 & 0.4716 & 0.7210 \\
Llama & Sensitivity & 0 & 7.154 & 11.068 & 0.4582 & 0.7193 \\
Llama & Full & 0 & 7.111 & 11.026 & 0.4749 & 0.7247 \\
\midrule
Llama & Nearest & 3 & 7.146 & 11.014 & 0.4515 & 0.7475 \\
Llama & Sensitivity & 3 & 7.154 & 11.031 & 0.4515 & 0.7466 \\
Llama & Full & 3 & 7.115 & 10.994 & 0.4649 & 0.7517 \\
\midrule
Llama & Nearest & 5 & 7.131 & 11.007 & 0.4883 & 0.7513 \\
Llama & Sensitivity & 5 & 7.142 & 11.027 & 0.4816 & 0.7500 \\
Llama & Full & 5 & 7.101 & 10.988 & 0.4983 & 0.7555 \\
\bottomrule
\end{tabular}
\end{table}

\begin{table}[htbp]
\centering
\vspace{\abovecaptionskip}
{\normalsize Table~\ref*{tab:onecode-rules}: {\revcolor One-code rule ablations across calibration seeds} (continued).\par}
\vspace{\belowcaptionskip}
\small\setlength{\tabcolsep}{4pt}
\begin{tabular}{llrrrrr}
\toprule
Model & Method & Seed & HellaSwag$\uparrow$ & PIQA$\uparrow$ & WinoGrande$\uparrow$ & Fit-7$\downarrow$ \\
\midrule
Qwen & Nearest & 0 & 0.7084 & 0.7584 & 0.5722 & 0.0401 \\
Qwen & Sensitivity & 0 & 0.7085 & 0.7595 & 0.5714 & 0.0421 \\
Qwen & Full & 0 & 0.7094 & 0.7633 & 0.5746 & 0.0324 \\
\midrule
Qwen & Nearest & 3 & 0.7107 & 0.7753 & 0.5801 & 0.0394 \\
Qwen & Sensitivity & 3 & 0.7107 & 0.7764 & 0.5793 & 0.0436 \\
Qwen & Full & 3 & 0.7115 & 0.7797 & 0.5817 & 0.0327 \\
\midrule
Qwen & Nearest & 5 & 0.7122 & 0.7590 & 0.5856 & 0.0385 \\
Qwen & Sensitivity & 5 & 0.7122 & 0.7595 & 0.5848 & 0.0429 \\
Qwen & Full & 5 & 0.7129 & 0.7622 & 0.5872 & 0.0330 \\
\midrule
Llama & Nearest & 0 & 0.7518 & 0.7943 & 0.6054 & 0.1237 \\
Llama & Sensitivity & 0 & 0.7516 & 0.7933 & 0.6038 & 0.1268 \\
Llama & Full & 0 & 0.7524 & 0.7987 & 0.6077 & 0.1187 \\
\midrule
Llama & Nearest & 3 & 0.7500 & 0.7867 & 0.6077 & 0.1243 \\
Llama & Sensitivity & 3 & 0.7500 & 0.7867 & 0.6077 & 0.1252 \\
Llama & Full & 3 & 0.7507 & 0.7916 & 0.6109 & 0.1182 \\
\midrule
Llama & Nearest & 5 & 0.7537 & 0.7943 & 0.6038 & 0.1169 \\
Llama & Sensitivity & 5 & 0.7536 & 0.7933 & 0.6022 & 0.1194 \\
Llama & Full & 5 & 0.7544 & 0.7987 & 0.6062 & 0.1115 \\
\bottomrule
\end{tabular}
\end{table}

The local objectives explain why sensitivity alone is insufficient in
these comparisons (Table~\ref{tab:onecode-local}).
Sensitivity-only achieves the lowest value of its own objective,
$0.912/0.921$ on Qwen/Llama, but raises the mismatch term to $1.043/1.036$.
The full rule accepts sensitivity terms of $0.947/0.953$ while lowering
mismatch to $0.918/0.926$, and obtains the best endpoints.
Local errors are measured on all layers using refinement.
Each layer's error is normalized by its output-element count, then
averaged across layers. Measurement segments are independent of, and
disjoint from, the rule-fitting data.
\begin{table}[htbp]
\centering
\caption{Local rule objectives at seed 0, each normalized to nearest rounding. Lower is better for both objectives and Fit-7.}
\label{tab:onecode-local}
\small\setlength{\tabcolsep}{4pt}
\begin{tabular}{llrrr}
\toprule
Model & Rule & \shortstack{Sensitivity-weighted\\local error$\downarrow$} & \shortstack{Mismatch-weighted\\local error$\downarrow$} & Fit-7$\downarrow$ \\
\midrule
Qwen & Nearest & 1.000 & 1.000 & 0.0401 \\
Qwen & Sensitivity & 0.912 & 1.043 & 0.0421 \\
Qwen & Full & 0.947 & 0.918 & 0.0324 \\
\midrule
Llama & Nearest & 1.000 & 1.000 & 0.1237 \\
Llama & Sensitivity & 0.921 & 1.036 & 0.1268 \\
Llama & Full & 0.953 & 0.926 & 0.1187 \\
\bottomrule
\end{tabular}
\end{table}

The observed edit rate is below the $1/32=3.125\%$ element-level ceiling
(Table~\ref{tab:onecode-edits}). Qwen changes an average of $0.61$ codes
per Block32, affecting $1.91\%$ of A4 elements; Llama changes $0.68$,
affecting $2.12\%$.
Table~\ref{tab:onecode-cost} separates one-time quantization cost from
recurring decoding overhead, measured against the same W4A4 model with
one-code refinement disabled.
\begin{table}[htbp]
\centering
\caption{Observed one-code edits. Codes per block and the fraction of all A4 elements use different denominators.}
\label{tab:onecode-edits}
\small\setlength{\tabcolsep}{4pt}
\begin{tabular}{lrrr}
\toprule
Model & \shortstack{Edited codes\\per Block32} & \shortstack{Edited A4\\elements} & Rule ceiling \\
\midrule
Qwen3-8B & 0.61 & 1.91\% & 3.125\% \\
Llama-3-8B & 0.68 & 2.12\% & 3.125\% \\
\bottomrule
\end{tabular}
\end{table}

\begin{table}[htbp]
\centering
\caption{One-time quantization cost (upper panel) and per-forward decoding overhead (lower panel) of one-code refinement. Decoding uses batch=1, seq=2048, and greedy generation. Measurements use a single NVIDIA A100-SXM4-80GB GPU with the PyTorch 2.9.1 (CUDA 12.6) eager/reference fake-quant backend.}
\label{tab:onecode-cost}
\small\setlength{\tabcolsep}{4pt}
\begin{tabular}{lrrr}
\toprule
Model & Without refinement$\downarrow$ & With refinement$\downarrow$ & Change$\downarrow$ \\
\midrule
Qwen3-8B & 58.2 min & 61.6 min & +3.4 min (+5.8\%) \\
Llama-3-8B & 54.7 min & 57.8 min & +3.1 min (+5.7\%) \\
\bottomrule
\end{tabular}
\par\medskip
\begin{tabular}{lrrr}
\toprule
Model & Without refinement$\uparrow$ & With refinement$\uparrow$ & Change$\uparrow$ \\
\midrule
Qwen3-8B & 41.7 tok/s & 40.6 tok/s & $-$2.6\% \\
Llama-3-8B & 43.2 tok/s & 42.1 tok/s & $-$2.5\% \\
\bottomrule
\end{tabular}
\end{table}

\paragraph{Three-term full-projection error decomposition.}
\label{app:error-decomposition}
To directly test whether the one-code A4 rule activates negative cross-cancellation (Eq.~\ref{eq:bilinear-decomp}), we decompose the full-projection output MSE into activation risk $T_A = \operatorname{tr}(G_Q E_X^\top E_X)$, weight risk $T_W = \operatorname{tr}(G_{E_W} X^\top X)$, and the signed cross-term $T_{\mathrm{cross}} = -2\operatorname{tr}(K_Q X^\top E_X)$, with $T_{\mathrm{total}} = T_A + T_W + T_{\mathrm{cross}}$. We evaluate Qwen3-8B (36 layers) and Llama-3-8B (32 layers) at seed~0 across 128 held-out 2048-token windows (64 C4, 64 WikiText-2) disjoint from calibration. All seven linear projections are evaluated under fixed-input replay using cached layer inputs from Stage-III-disabled \HiRes Nearest A4, so the three \HiRes A4 variants share the exact same Stage~I transform, GPTQ weights, and $T_W$ while isolating A4 code selection. Each projection's MSE is normalized by output-element count, averaged equally across projections, and scaled per model so that \HiRes Nearest A4 equals $1.000$.

\begin{figure}[htbp]
  \centering
  \includegraphics[width=0.98\textwidth]{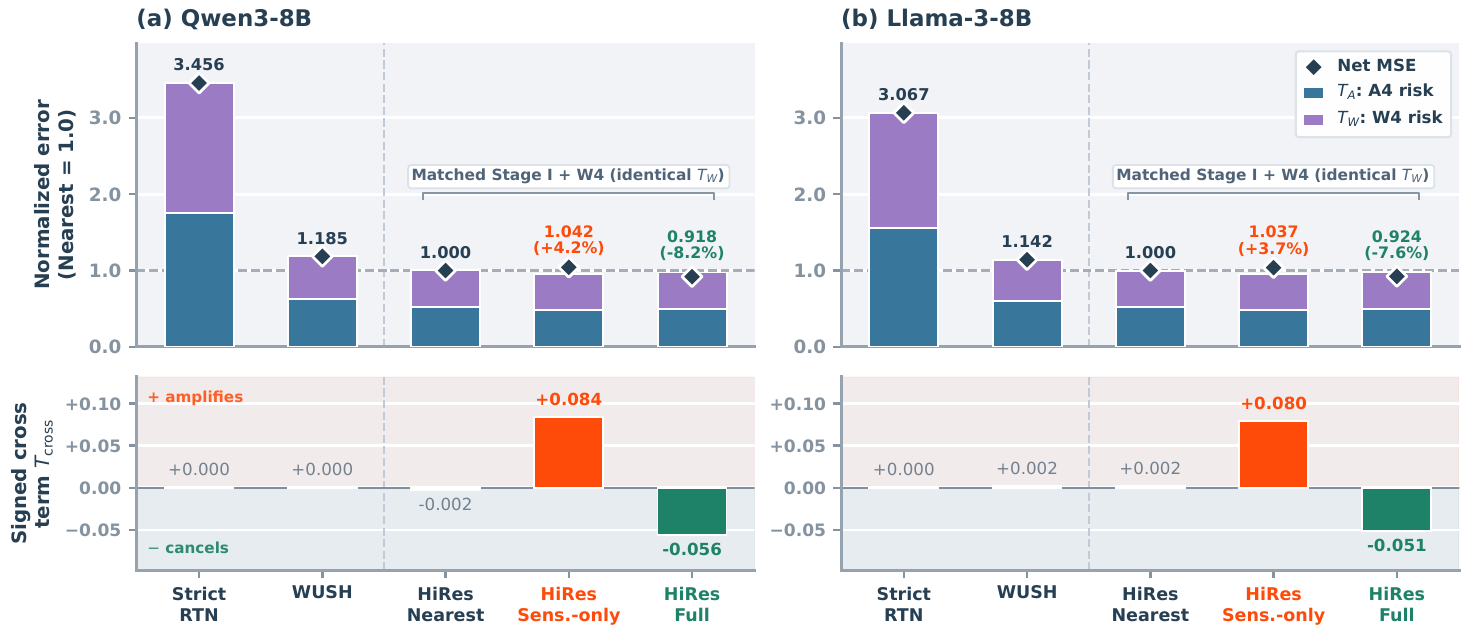}
  \caption{\textbf{Full-projection three-term error decomposition across 128 held-out windows and all seven decoder linear projections.} \textbf{Top:} Stacked A4 risk ($T_A$) and W4 risk ($T_W$) with net MSE diamonds, normalized per model to \HiRes Nearest A4 ($=1.000$). \textbf{Bottom:} Signed cross-term $T_{\mathrm{cross}}$. Sensitivity-only A4 lowers $T_A$ but drives $T_{\mathrm{cross}}$ strongly positive ($+0.084 / +0.080$), worsening net MSE; full one-code A4 turns $T_{\mathrm{cross}}$ negative ($-0.056 / -0.051$), reducing net MSE by $8.2\%$ on Qwen3-8B and $7.6\%$ on Llama-3-8B.}
  \label{fig:error-decomposition}
\end{figure}

Figure~\ref{fig:error-decomposition} and Table~\ref{tab:error-decomp-overall} confirm the predicted sign reversal. On both models, sensitivity-only A4 lowers $T_A$ ($0.522 \to 0.478$ on Qwen3-8B; $0.519 \to 0.478$ on Llama-3-8B) but drives $T_{\mathrm{cross}}$ sharply positive ($+0.084$ and $+0.080$), increasing net MSE by $4.2\%$ and $3.7\%$. Full one-code A4 flips $T_{\mathrm{cross}}$ negative ($-0.056$ and $-0.051$), cutting net MSE to $0.918$ ($-8.2\%$) on Qwen3-8B and $0.924$ ($-7.6\%$) on Llama-3-8B. These full-projection ratios closely track the separate local objectives in Table~\ref{tab:onecode-local}.

\begin{table}[htbp]
\centering
\caption{Full-projection three-term error decomposition normalized to \HiRes (Nearest A4) total MSE (128 held-out windows, 7 projections, seed~0).}
\label{tab:error-decomp-overall}
\small\setlength{\tabcolsep}{6pt}
\begin{tabular}{llrrrr}
\toprule
Model & Configuration & $T_A\downarrow$ & $T_W\downarrow$ & $T_{\mathrm{cross}}$ & Total MSE$\downarrow$ \\
\midrule
Qwen3-8B & Strict RTN & 1.748 & 1.708 & $+$0.000 & 3.456 \\
 & WUSH & 0.623 & 0.561 & $+$0.000 & 1.185 \\
 & \HiRes (Nearest A4) & 0.522 & 0.481 & $-$0.002 & 1.000 \\
 & \HiRes (Sensitivity-only A4) & 0.478 & 0.481 & $+$0.084 & 1.042 \\
 & \textbf{\HiRes (Full one-code A4)} & 0.494 & 0.481 & \textbf{$-$0.056} & \textbf{0.918} \\
\midrule
Llama-3-8B & Strict RTN & 1.554 & 1.512 & $+$0.000 & 3.067 \\
 & WUSH & 0.597 & 0.542 & $+$0.002 & 1.142 \\
 & \HiRes (Nearest A4) & 0.519 & 0.479 & $+$0.002 & 1.000 \\
 & \HiRes (Sensitivity-only A4) & 0.478 & 0.479 & $+$0.080 & 1.037 \\
 & \textbf{\HiRes (Full one-code A4)} & 0.496 & 0.479 & \textbf{$-$0.051} & \textbf{0.924} \\
\bottomrule
\end{tabular}
\end{table}

The same sign reversal holds on C4 and WikiText-2 individually (Table~\ref{tab:error-decomp-corpus}) and across all major projections (Table~\ref{tab:error-decomp-proj}). Small low-error projections such as \texttt{q\_proj} show negligible shift under full-projection cross-block interactions, while high-error projections (\texttt{v\_proj}, \texttt{o\_proj}, \texttt{gate\_proj}, \texttt{up\_proj}, and especially \texttt{down\_proj}, with $T_{\mathrm{cross}} = -0.145 / -0.120$) account for most of the cross-cancellation gain.

\begin{table}[htbp]
\centering
\caption{Per-corpus three-term error decomposition (64 windows per corpus, using the same per-model Nearest denominator as Table~\ref{tab:error-decomp-overall}).}
\label{tab:error-decomp-corpus}
\scriptsize\setlength{\tabcolsep}{5pt}
\begin{tabular}{lllrrrr}
\toprule
Model & Corpus & Configuration & $T_A$ & $T_W$ & $T_{\mathrm{cross}}$ & Total MSE \\
\midrule
Qwen3-8B & C4 & Strict RTN & 1.881 & 1.831 & $+$0.000 & 3.713 \\
 & C4 & WUSH & 0.669 & 0.601 & $+$0.001 & 1.270 \\
 & C4 & \HiRes (Nearest A4) & 0.560 & 0.513 & $-$0.003 & 1.071 \\
 & C4 & \HiRes (Sensitivity-only) & 0.514 & 0.513 & $+$0.093 & 1.121 \\
 & C4 & \textbf{\HiRes (Full one-code)} & 0.527 & 0.513 & \textbf{$-$0.062} & \textbf{0.979} \\
\cmidrule(lr){2-7}
 & WT2 & Strict RTN & 1.614 & 1.584 & $+$0.000 & 3.199 \\
 & WT2 & WUSH & 0.577 & 0.522 & $+$0.000 & 1.100 \\
 & WT2 & \HiRes (Nearest A4) & 0.484 & 0.448 & $-$0.002 & 0.929 \\
 & WT2 & \HiRes (Sensitivity-only) & 0.441 & 0.448 & $+$0.075 & 0.964 \\
 & WT2 & \textbf{\HiRes (Full one-code)} & 0.461 & 0.448 & \textbf{$-$0.050} & \textbf{0.858} \\
\midrule
Llama-3-8B & C4 & Strict RTN & 1.617 & 1.569 & $+$0.000 & 3.186 \\
 & C4 & WUSH & 0.620 & 0.562 & $+$0.002 & 1.185 \\
 & C4 & \HiRes (Nearest A4) & 0.538 & 0.494 & $+$0.002 & 1.034 \\
 & C4 & \HiRes (Sensitivity-only) & 0.499 & 0.494 & $+$0.084 & 1.076 \\
 & C4 & \textbf{\HiRes (Full one-code)} & 0.511 & 0.494 & \textbf{$-$0.058} & \textbf{0.948} \\
\cmidrule(lr){2-7}
 & WT2 & Strict RTN & 1.492 & 1.456 & $+$0.000 & 2.948 \\
 & WT2 & WUSH & 0.574 & 0.523 & $+$0.002 & 1.098 \\
 & WT2 & \HiRes (Nearest A4) & 0.501 & 0.463 & $+$0.002 & 0.966 \\
 & WT2 & \HiRes (Sensitivity-only) & 0.458 & 0.463 & $+$0.076 & 0.997 \\
 & WT2 & \textbf{\HiRes (Full one-code)} & 0.481 & 0.463 & \textbf{$-$0.044} & \textbf{0.900} \\
\bottomrule
\end{tabular}
\end{table}

\begin{table}[htbp]
\centering
\caption{Per-projection three-term error decomposition across all decoder layers and 128 windows (same per-model Nearest denominator; $T_W$ is shared across the three A4 rules).}
\label{tab:error-decomp-proj}
\scriptsize\setlength{\tabcolsep}{4.2pt}
\begin{tabular}{llrrrr@{\hspace{12pt}}rrrr}
\toprule
 & & \multicolumn{4}{c}{Qwen3-8B} & \multicolumn{4}{c}{Llama-3-8B} \\
\cmidrule(lr){3-6}\cmidrule(lr){7-10}
Projection & A4 Rule & $T_A$ & $T_W$ & $T_{\mathrm{cross}}$ & Total & $T_A$ & $T_W$ & $T_{\mathrm{cross}}$ & Total \\
\midrule
\texttt{q\_proj} & Nearest & 0.255 & 0.363 & $-$0.001 & 0.617 & 0.262 & 0.375 & $+$0.001 & 0.638 \\
 & Sensitivity & 0.246 & 0.363 & $+$0.011 & 0.620 & 0.254 & 0.375 & $+$0.008 & 0.637 \\
 & Full & 0.252 & 0.363 & $+$0.001 & 0.617 & 0.262 & 0.375 & $+$0.005 & 0.641 \\
\midrule
\texttt{k\_proj} & Nearest & 0.217 & 0.346 & $+$0.002 & 0.564 & 0.232 & 0.334 & $-$0.001 & 0.565 \\
 & Sensitivity & 0.208 & 0.346 & $+$0.015 & 0.568 & 0.223 & 0.334 & $+$0.009 & 0.566 \\
 & Full & 0.219 & 0.346 & $-$0.004 & 0.561 & 0.236 & 0.334 & $-$0.006 & 0.565 \\
\midrule
\texttt{v\_proj} & Nearest & 0.369 & 0.440 & $-$0.002 & 0.807 & 0.390 & 0.435 & $+$0.003 & 0.828 \\
 & Sensitivity & 0.350 & 0.440 & $+$0.037 & 0.827 & 0.371 & 0.435 & $+$0.038 & 0.844 \\
 & Full & 0.361 & 0.440 & $-$0.038 & 0.763 & 0.385 & 0.435 & $-$0.033 & 0.787 \\
\midrule
\texttt{o\_proj} & Nearest & 0.538 & 0.468 & $+$0.001 & 1.007 & 0.592 & 0.472 & $+$0.002 & 1.066 \\
 & Sensitivity & 0.497 & 0.468 & $+$0.093 & 1.059 & 0.548 & 0.472 & $+$0.097 & 1.117 \\
 & Full & 0.508 & 0.468 & $-$0.062 & 0.915 & 0.565 & 0.472 & $-$0.064 & 0.973 \\
\midrule
\texttt{gate\_proj} & Nearest & 0.657 & 0.579 & $-$0.004 & 1.232 & 0.652 & 0.559 & $-$0.002 & 1.210 \\
 & Sensitivity & 0.592 & 0.579 & $+$0.134 & 1.304 & 0.590 & 0.559 & $+$0.126 & 1.275 \\
 & Full & 0.598 & 0.579 & $-$0.073 & 1.104 & 0.598 & 0.559 & $-$0.071 & 1.087 \\
\midrule
\texttt{up\_proj} & Nearest & 0.630 & 0.553 & $+$0.002 & 1.184 & 0.626 & 0.503 & $+$0.001 & 1.130 \\
 & Sensitivity & 0.569 & 0.553 & $+$0.115 & 1.236 & 0.570 & 0.503 & $+$0.099 & 1.173 \\
 & Full & 0.597 & 0.553 & $-$0.074 & 1.076 & 0.599 & 0.503 & $-$0.070 & 1.033 \\
\midrule
\texttt{down\_proj} & Nearest & 0.989 & 0.614 & $-$0.015 & 1.588 & 0.882 & 0.671 & $+$0.011 & 1.563 \\
 & Sensitivity & 0.884 & 0.614 & $+$0.184 & 1.682 & 0.794 & 0.671 & $+$0.181 & 1.645 \\
 & Full & 0.923 & 0.614 & $-$0.145 & 1.392 & 0.829 & 0.671 & $-$0.120 & 1.380 \\
\bottomrule
\end{tabular}
\end{table}

\subsection{Recovery Paths and Remeasurement}
\label{app:recovery-ablation}

We test which paths benefit from recovery and when their fitting statistics
should be collected. The path ablations keep per-path capacity fixed;
the timing controls keep both operators fixed and change their fitting state.

\paragraph{Attention and MLP paths.}
\begin{table}[htbp]
\centering
\caption{Structural-recovery factorization and fresh-statistics control.}
\label{tab:structural-ablations}
\small\setlength{\tabcolsep}{4pt}
\begin{tabular}{llrrrr}
\toprule
Model & Configuration & WT2$\downarrow$ & C4$\downarrow$ & ARC-C$\uparrow$ & Fit-3$\downarrow$ \\
\midrule
Qwen3-8B & Both paths disabled & 10.1858 & 16.4524 & 50.84 & .0650 \\
& Attention only & 10.1588 & 16.4411 & 52.17 & .0486 \\
& MLP only & 10.0965 & 16.3397 & 53.51 & .0391 \\
& Both paths & 10.0552 & 16.2987 & 54.85 & .0326 \\
& Stale MLP statistics & 10.1090 & 16.3722 & 53.18 & .0417 \\
\midrule
Llama-3-8B & Both paths disabled & 7.1417 & 11.1163 & 46.82 & .1553 \\
& Attention only & 7.1328 & 11.0855 & 46.49 & .1548 \\
& MLP only & 7.1263 & 11.0618 & 46.49 & .1537 \\
& Both paths & 7.1110 & 11.0262 & 47.49 & .1483 \\
& Stale MLP statistics & 7.1454 & 11.0816 & 45.82 & .1582 \\
\bottomrule
\end{tabular}
\end{table}

The full pair is best on both models, improving over the stronger single path
by $0.0065$ on Qwen3-8B and $0.0054$ on Llama-3-8B. On Qwen3-8B each single
path also improves clearly over disabling both. On Llama-3-8B the two single
paths lie within $0.0016$ of the no-recovery control and within $0.0012$ of
each other. The full pair gives the largest separation from the no-recovery
control. Each comparison varies path presence at fixed per-path capacity.
The stale control measures the MLP recovery statistics
on the Stage-II parent before installing the attention correction, while
retaining the normal attention-to-MLP deployment order.

Both recovery paths give the lowest damage across seeds
(Table~\ref{tab:structural-multiseed}).
In all $6/6$ model--seed pairs, damage increases in the order: both paths,
MLP only, attention only, and neither path.
The full pair improves over MLP-only recovery, the strongest single-path control,
by $0.0036$--$0.0043$ on Qwen and $0.0029$--$0.0032$ on Llama.
The two single paths differ by only $0.0003$--$0.0006$ on Llama,
versus $0.0025$--$0.0064$ on Qwen.
Their Llama gains over no recovery are $0.0016$--$0.0025$.
The full pair has the largest gain, with per-path capacity fixed across
the compared configurations.
\begin{table}[htbp]
\centering
\caption{Recovery-path ablations across calibration seeds. Per-path capacity is fixed, and each configuration refits recovery statistics. Accuracy is reported as a fraction.}
\label{tab:structural-multiseed}
\small\setlength{\tabcolsep}{4pt}
\begin{tabular}{llrrrrr}
\toprule
Model & Config. & Seed & WT2$\downarrow$ & C4$\downarrow$ & ARC-C$\uparrow$ & ARC-E$\uparrow$ \\
\midrule
Qwen & None & 0 & 10.186 & 16.452 & 0.5084 & 0.7837 \\
Qwen & Attention & 0 & 10.159 & 16.441 & 0.5217 & 0.7837 \\
Qwen & MLP & 0 & 10.096 & 16.340 & 0.5351 & 0.7744 \\
Qwen & Both & 0 & 10.055 & 16.299 & 0.5485 & 0.7761 \\
\midrule
Qwen & None & 3 & 10.264 & 16.437 & 0.5050 & 0.8022 \\
Qwen & Attention & 3 & 10.232 & 16.413 & 0.5151 & 0.8013 \\
Qwen & MLP & 3 & 10.196 & 16.351 & 0.5217 & 0.7946 \\
Qwen & Both & 3 & 10.160 & 16.315 & 0.5351 & 0.7963 \\
\midrule
Qwen & None & 5 & 10.287 & 16.466 & 0.5050 & 0.7824 \\
Qwen & Attention & 5 & 10.231 & 16.418 & 0.5284 & 0.7803 \\
Qwen & MLP & 5 & 10.188 & 16.344 & 0.5351 & 0.7727 \\
Qwen & Both & 5 & 10.150 & 16.306 & 0.5485 & 0.7744 \\
\midrule
Llama & None & 0 & 7.142 & 11.116 & 0.4682 & 0.7298 \\
Llama & Attention & 0 & 7.133 & 11.085 & 0.4649 & 0.7344 \\
Llama & MLP & 0 & 7.126 & 11.062 & 0.4649 & 0.7273 \\
Llama & Both & 0 & 7.111 & 11.026 & 0.4749 & 0.7247 \\
\midrule
Llama & None & 3 & 7.144 & 11.077 & 0.4582 & 0.7563 \\
Llama & Attention & 3 & 7.137 & 11.053 & 0.4548 & 0.7614 \\
Llama & MLP & 3 & 7.133 & 11.035 & 0.4548 & 0.7546 \\
Llama & Both & 3 & 7.115 & 10.994 & 0.4649 & 0.7517 \\
\midrule
Llama & None & 5 & 7.134 & 11.082 & 0.4916 & 0.7609 \\
Llama & Attention & 5 & 7.127 & 11.059 & 0.4883 & 0.7673 \\
Llama & MLP & 5 & 7.120 & 11.030 & 0.4883 & 0.7584 \\
Llama & Both & 5 & 7.101 & 10.988 & 0.4983 & 0.7555 \\
\bottomrule
\end{tabular}
\end{table}

\begin{table}[htbp]
\centering
\vspace{\abovecaptionskip}
{\normalsize Table~\ref*{tab:structural-multiseed}: Recovery-path ablations across calibration seeds (continued).\par}
\vspace{\belowcaptionskip}
\small\setlength{\tabcolsep}{4pt}
\begin{tabular}{llrrrrr}
\toprule
Model & Config. & Seed & HellaSwag$\uparrow$ & PIQA$\uparrow$ & WinoGrande$\uparrow$ & Fit-7$\downarrow$ \\
\midrule
Qwen & None & 0 & 0.7079 & 0.7601 & 0.5770 & 0.0551 \\
Qwen & Attention & 0 & 0.7092 & 0.7650 & 0.5817 & 0.0402 \\
Qwen & MLP & 0 & 0.7088 & 0.7612 & 0.5777 & 0.0360 \\
Qwen & Both & 0 & 0.7094 & 0.7633 & 0.5746 & 0.0324 \\
\midrule
Qwen & None & 3 & 0.7103 & 0.7769 & 0.5833 & 0.0549 \\
Qwen & Attention & 3 & 0.7114 & 0.7807 & 0.5864 & 0.0434 \\
Qwen & MLP & 3 & 0.7110 & 0.7775 & 0.5848 & 0.0370 \\
Qwen & Both & 3 & 0.7115 & 0.7797 & 0.5817 & 0.0327 \\
\midrule
Qwen & None & 5 & 0.7113 & 0.7590 & 0.5896 & 0.0578 \\
Qwen & Attention & 5 & 0.7127 & 0.7633 & 0.5927 & 0.0397 \\
Qwen & MLP & 5 & 0.7124 & 0.7601 & 0.5904 & 0.0372 \\
Qwen & Both & 5 & 0.7129 & 0.7622 & 0.5872 & 0.0330 \\
\midrule
Llama & None & 0 & 0.7484 & 0.7884 & 0.6069 & 0.1241 \\
Llama & Attention & 0 & 0.7521 & 0.7933 & 0.6077 & 0.1222 \\
Llama & MLP & 0 & 0.7513 & 0.7971 & 0.6093 & 0.1216 \\
Llama & Both & 0 & 0.7524 & 0.7987 & 0.6077 & 0.1187 \\
\midrule
Llama & None & 3 & 0.7470 & 0.7818 & 0.6101 & 0.1232 \\
Llama & Attention & 3 & 0.7504 & 0.7862 & 0.6109 & 0.1216 \\
Llama & MLP & 3 & 0.7496 & 0.7900 & 0.6125 & 0.1213 \\
Llama & Both & 3 & 0.7507 & 0.7916 & 0.6109 & 0.1182 \\
\midrule
Llama & None & 5 & 0.7501 & 0.7878 & 0.6054 & 0.1171 \\
Llama & Attention & 5 & 0.7541 & 0.7922 & 0.6062 & 0.1154 \\
Llama & MLP & 5 & 0.7531 & 0.7971 & 0.6077 & 0.1147 \\
Llama & Both & 5 & 0.7544 & 0.7987 & 0.6062 & 0.1115 \\
\bottomrule
\end{tabular}
\end{table}

\paragraph{Remeasurement after attention.}
With W4/A4 operators already installed in both conditions, we vary whether
MLP statistics are measured before or after the attention correction.
Remeasuring after attention improves WT2, C4,
and ARC-C individually in all $6/6$ model--seed pairs
(Table~\ref{tab:fresh-stale-multiseed}).
Across the three calibration seeds, remeasuring after attention reduces Fit-3
by $0.0080\pm0.0014$ on Qwen3-8B (range $0.0064$--$0.0091$) and by
$0.0107\pm0.0019$ on Llama-3-8B (range $0.0093$--$0.0130$), improving all
$3/3$ seeds on each model.
The next control instead varies whether recovery is fitted before or after
A4 encoding and measures both local reconstruction error and task endpoints.
\begin{table}[htbp]
\centering
\caption{MLP statistics measured after (fresh) or before (stale) attention recovery. The last column is paired fresh-minus-stale Fit-3; both conditions retain attention-to-MLP deployment.}
\label{tab:fresh-stale-multiseed}
\small\setlength{\tabcolsep}{4pt}
\begin{tabular}{lr lrrrrr}
\toprule
Model & Seed & Statistics & WT2$\downarrow$ & C4$\downarrow$ & ARC-C$\uparrow$ & Fit-3$\downarrow$ & $\Delta$ Fit-3$\downarrow$ \\
\midrule
Qwen & 0 & Fresh & 10.055 & 16.299 & 0.5485 & 0.0326 & -0.0091 \\
Qwen & 0 & Stale & 10.109 & 16.372 & 0.5318 & 0.0417 & -- \\
Qwen & 3 & Fresh & 10.160 & 16.315 & 0.5351 & 0.0407 & -0.0064 \\
Qwen & 3 & Stale & 10.211 & 16.399 & 0.5284 & 0.0471 & -- \\
Qwen & 5 & Fresh & 10.150 & 16.306 & 0.5485 & 0.0358 & -0.0085 \\
Qwen & 5 & Stale & 10.204 & 16.373 & 0.5351 & 0.0443 & -- \\
\midrule
Llama & 0 & Fresh & 7.111 & 11.026 & 0.4749 & 0.1483 & -0.0099 \\
Llama & 0 & Stale & 7.145 & 11.082 & 0.4582 & 0.1582 & -- \\
Llama & 3 & Fresh & 7.115 & 10.994 & 0.4649 & 0.1514 & -0.0093 \\
Llama & 3 & Stale & 7.148 & 11.084 & 0.4515 & 0.1607 & -- \\
Llama & 5 & Fresh & 7.101 & 10.988 & 0.4983 & 0.1392 & -0.0130 \\
Llama & 5 & Stale & 7.140 & 11.069 & 0.4749 & 0.1522 & -- \\
\bottomrule
\end{tabular}
\end{table}

\paragraph{Remeasurement after encoding.}
\label{app:encoding-boundary}
The recovery operators and parameter counts are identical in both arms:
attention uses two affine coefficients $(a_{\ell h},b_{\ell h})$ per
key--value head (Eq.~\ref{eq:hires-vo-fit}), and MLP uses three
coefficients $(\alpha,\beta,\eta)$ per layer (Eq.~\ref{eq:hires-mlp-update}).
Both use the same weighted ridge fit, median-deviation shrinkage, and
projection to $\|[\alpha,\beta,\eta]\|_1\leq1/8$ for MLP.
We change only whether the fitting second moments are collected before
A4 encoding or after actual MXFP4 encoding.

Post-encoding statistics improve both the deployed local error and the
task endpoints (Table~\ref{tab:encoding-boundary}).
Both arms measure MSE in the same actual encoded state, relative to
BF16 layer outputs. We include every layer with attention V/O or MLP
recovery, normalize each layer's squared error by its output-element count,
and take the arithmetic mean across layers.
The measurement set contains 512 independent sequences of length 2048,
disjoint from coefficient-fitting data.
The before column measures the common model without recovery; the after
column applies the coefficients fitted by each arm.
\begin{table}[htbp]
\centering
\caption{Recovery fitted before or after A4 encoding at seed 0: task endpoints (upper panel) and MSE in the actual encoded state (lower panel). The operator forms and coefficient counts are fixed.}
\label{tab:encoding-boundary}
\small\setlength{\tabcolsep}{4pt}
\begin{tabular}{llrrrr}
\toprule
Model & Statistics & WT2$\downarrow$ & C4$\downarrow$ & ARC-C$\uparrow$ & Fit-3$\downarrow$ \\
\midrule
Qwen & pre-encoding & 10.143 & 16.451 & 0.5217 & 0.0487 \\
Qwen & post-encoding & 10.055 & 16.299 & 0.5485 & 0.0326 \\
\midrule
Llama & pre-encoding & 7.149 & 11.123 & 0.4649 & 0.1572 \\
Llama & post-encoding & 7.111 & 11.026 & 0.4749 & 0.1483 \\
\bottomrule
\end{tabular}
\par\medskip
\begin{tabular}{llrrr}
\toprule
Model & Statistics & Local MSE before$\downarrow$ & Local MSE after$\downarrow$ & Reduction$\uparrow$ \\
\midrule
Qwen & pre-encoding & $3.030\times 10^{-3}$ & $3.030\times 10^{-3}$ & 0.02\% \\
Qwen & post-encoding & $3.030\times 10^{-3}$ & $2.742\times 10^{-3}$ & 9.52\% \\
Llama & pre-encoding & $4.164\times 10^{-5}$ & $4.164\times 10^{-5}$ & 0.01\% \\
Llama & post-encoding & $4.164\times 10^{-5}$ & $3.810\times 10^{-5}$ & 8.49\% \\
\bottomrule
\end{tabular}
\end{table}

Coefficients fitted before encoding barely change deployed MSE:
the reductions are $0.019\%$ on Qwen and $0.012\%$ on Llama.
Fitting after encoding gives $9.52\%$ and $8.49\%$ reductions with the
same operator forms, counts, and locations.
Across calibration seeds, post-encoding fitting lowers Fit-3 by
$0.0161$--$0.0165$ on Qwen and $0.0083$--$0.0096$ on Llama,
above the $0.003$ threshold (Table~\ref{tab:encoding-boundary-seeds}).
\begin{table}[htbp]
\centering
\caption{Encoding-boundary control across seeds. WT2, C4, and ARC-C are pre-encoding-fit endpoints; post-encoding-fit endpoints appear in Table~\ref{tab:fresh-stale-multiseed} (fresh rows).}
\label{tab:encoding-boundary-seeds}
\small\setlength{\tabcolsep}{4pt}
\begin{tabular}{lrrrrrrr}
\toprule
Model & Seed & Pre WT2$\downarrow$ & Pre C4$\downarrow$ & Pre ARC-C$\uparrow$ & Pre Fit-3$\downarrow$ & Post Fit-3$\downarrow$ & Gain$\uparrow$ \\
\midrule
Qwen & 0 & 10.143 & 16.451 & 0.5217 & 0.0487 & 0.0326 & 0.0161 \\
Qwen & 3 & 10.220 & 16.421 & 0.5151 & 0.0571 & 0.0407 & 0.0165 \\
Qwen & 5 & 10.243 & 16.469 & 0.5217 & 0.0521 & 0.0358 & 0.0163 \\
\midrule
Llama & 0 & 7.149 & 11.123 & 0.4649 & 0.1572 & 0.1483 & 0.0089 \\
Llama & 3 & 7.158 & 11.102 & 0.4548 & 0.1610 & 0.1514 & 0.0096 \\
Llama & 5 & 7.135 & 11.074 & 0.4883 & 0.1475 & 0.1392 & 0.0083 \\
\bottomrule
\end{tabular}
\end{table}

\paragraph{Transferring Stage~III sequential recovery to WUSH.}
\label{app:wush-transfer}
To check whether post-encoding sequential recovery (Eq.~\ref{eq:conditional-stage}) transfers to a different quantization backbone, we attach Stage~III directly to our strongest baseline, WUSH, as well as to \HiRes with Nearest A4 (seed~0, Qwen3-8B and Llama-3-8B). Both WUSH variants keep WUSH's geometry, W4 weights, and dynamic nearest A4 rule unchanged, adding only the per-KV-head attention scalars and three SwiGLU scalars fitted on $128\times 2048$ C4-train sequences (Appendix~\ref{app:hires-structural-fit}). The \emph{pre-A4} variant fits recovery with A4 encoding disabled; the \emph{post-A4 sequential} variant fits attention on the live W4A4 state, installs it, and remeasures $(g, u)$ to fit SwiGLU. Block-level local MSE reductions are evaluated on 512 unseen 2048-token sequences from shared unrecovered block inputs, while task metrics use full deployed W4A4 inference. (Unlike WUSH+R in Tables~\ref{tab:geometry-replacement}--\ref{tab:geometry-paired}, which also includes Stage~II, the WUSH transplant here adds only Stage~III.)

\begin{table}[htbp]
\centering
\caption{Transplanting Stage~III recovery onto WUSH and \HiRes Nearest A4 (seed~0). Local MSE reduction is measured over 512 unseen sequences under fixed-input block replay (\HiRes Full reports the recovery-path reduction from Table~\ref{tab:encoding-boundary} and task endpoints from Tables~\ref{tab:raw-qwen8b} and~\ref{tab:raw-llama8b}).}
\label{tab:wush-transfer}
\small\setlength{\tabcolsep}{4pt}
\begin{tabular}{llrrrrrr}
\toprule
Model & Configuration & Local MSE red.\ (\%)$\uparrow$ & WT2$\downarrow$ & C4$\downarrow$ & Acc5 (\%)$\uparrow$ & Fit-7$\downarrow$ & Fit-3$\downarrow$ \\
\midrule
Qwen3-8B & WUSH & 0.00 & 10.42 & 16.59 & 67.92 & 0.0500 & 0.0604 \\
 & WUSH + Stage~III (pre-A4) & 0.03 & 10.42 & 16.59 & 67.90 & 0.0503 & 0.0605 \\
 & \textbf{WUSH + Stage~III (post-A4 seq.)} & \textbf{8.17} & 10.27 & 16.47 & \textbf{68.03} & 0.0403 & 0.0481 \\
 & \HiRes Nearest + Stage~III (pre-A4) & 0.02 & 10.26 & 16.51 & 66.62 & 0.0468 & 0.0535 \\
 & \HiRes Nearest + Stage~III (post-A4 seq.) & 8.89 & 10.18 & 16.39 & 66.91 & 0.0401 & 0.0440 \\
 & \textbf{\HiRes Full} & \textbf{9.52}$^\ast$ & \textbf{10.06} & \textbf{16.30} & 67.44 & \textbf{0.0324} & \textbf{0.0326} \\
\midrule
Llama-3-8B & WUSH & 0.00 & 7.13 & 11.12 & 66.65 & 0.1248 & 0.1606 \\
 & WUSH + Stage~III (pre-A4) & 0.01 & 7.14 & 11.12 & 66.66 & 0.1252 & 0.1612 \\
 & \textbf{WUSH + Stage~III (post-A4 seq.)} & \textbf{6.94} & 7.13 & 11.07 & 67.12 & 0.1211 & 0.1549 \\
 & \HiRes Nearest + Stage~III (pre-A4) & 0.01 & 7.17 & 11.14 & 66.63 & 0.1285 & 0.1604 \\
 & \HiRes Nearest + Stage~III (post-A4 seq.) & 7.96 & 7.15 & 11.05 & 66.88 & 0.1237 & 0.1536 \\
 & \textbf{\HiRes Full} & \textbf{8.49}$^\ast$ & \textbf{7.11} & \textbf{11.03} & \textbf{67.17} & \textbf{0.1187} & \textbf{0.1483} \\
\bottomrule
\end{tabular}
\end{table}

Table~\ref{tab:wush-transfer} shows the same sharp boundary on WUSH as on \HiRes. Fitting Stage~III before A4 encoding leaves block MSE essentially unchanged ($< 0.03\%$ across the 512 sequences) and slightly hurts Fit-7. Fitting sequentially after A4 encoding cuts WUSH block MSE by $8.17\%$ on Qwen3-8B ($3.51 \to 3.23 \times 10^{-3}$) and $6.94\%$ on Llama-3-8B ($4.68 \to 4.36 \times 10^{-5}$), lowering Fit-7 by $0.0098$ and $0.0036$. On Qwen3-8B, WUSH + Stage~III achieves higher Acc5 ($68.03\%$ vs.\ $67.44\%$), while full \HiRes remains substantially stronger on WT2, C4, Fit-7 ($0.0324$ vs.\ $0.0403$), and Fit-3 ($0.0326$ vs.\ $0.0481$), and leads on every metric on Llama-3-8B---reflecting the additional gains from Stages~I and~II.

\paragraph{Summary of all four stage-handoff boundaries.}
Table~\ref{tab:four-boundaries} brings together all four transitions in \HiRes, comparing each stage when fitted against stale or simplified statistics versus the live state installed by its predecessor.

\begin{table}[htbp]
\centering
\caption{All four stage-handoff boundaries in \HiRes (seed~0). Each row compares stale or simplified statistics against matched post-stage statistics on its own local or task metric.}
\label{tab:four-boundaries}
\scriptsize\setlength{\tabcolsep}{3.8pt}
\begin{tabular}{@{}p{.19\linewidth}p{.21\linewidth}rrp{.24\linewidth}@{}}
\toprule
Stage Handoff & Metric (lower is better) & \shortstack[r]{Qwen3-8B\\(old$\to$new)} & \shortstack[r]{Llama-3-8B\\(old$\to$new)} & Comparison \\
\midrule
$DT \to$ GPTQ curvature & Relative W4 output MSE & $1.041 \to \mathbf{1.000}$ & $1.037 \to \mathbf{1.000}$ & Pre-$DT$ $H$ vs.\ coordinate-matched $H$ \\
Installed W4 $\to$ A4 rule & Relative mismatch error & $1.043 \to \mathbf{0.918}$ & $1.036 \to \mathbf{0.926}$ & Sensitivity-only vs.\ Full (Table~\ref{tab:onecode-local}) \\
A4 $\to$ Stage~III recovery & Relative block MSE & $1.000 \to \mathbf{0.905}$ & $1.000 \to \mathbf{0.915}$ & Pre-A4 vs.\ post-A4 fit (Table~\ref{tab:encoding-boundary}) \\
Attn $\to$ SwiGLU stats & Task Fit-3 & $0.0417 \to \mathbf{0.0326}$ & $0.1582 \to \mathbf{0.1483}$ & Stale pre-attn vs.\ fresh $(g,u)$ (Table~\ref{tab:fresh-stale-multiseed}) \\
\bottomrule
\end{tabular}
\end{table}

\subsection{Computational and Storage Costs}
\label{app:computational-cost}

We compare complete \HiRes with BF16 for persistent storage and full-forward
latency, and report one-time quantization separately. The storage and latency
results use the optimized inference layout and fused online operations below.

\paragraph{Persistent storage.}
Each target linear layer retains its own Block32 matrices: the folded input
transform $U=DT$, the upper triangle of $G_Q$ including its diagonal, and
the full $K_Q$, all in FP32. Symmetric entries of $G_Q$ are recovered by
mirroring the stored values. Each input block therefore uses
$4(1024+528+1024)=10{,}304$ bytes of matrix state.
Attention recovery adds two coefficients per key--value head; MLP recovery
adds $\alpha,\beta,\eta,\sigma_g$ per layer. Their FP32 storage totals
$2{,}880$, $2{,}560$, and $5{,}120$ bytes for Qwen3-8B, Llama-3-8B,
and Qwen3-32B, respectively.
Construction-only inverses, intermediate geometry, calibration arrays,
Hessians, and BF16 target-weight copies are released after quantization.
Table~\ref{tab:hires-storage} counts decoded BF16 parameters and persistent
auxiliary tensors; activations, temporary workspaces, and allocator reserves
are outside this static accounting. The total payload exceeds the BF16
parameter payload by $1.87\%$--$2.61\%$.

\begin{table}[htbp]
\centering
\caption{Static storage in the BF16-decoded implementation. Auxiliary storage
includes all retained matrices and recovery coefficients. Blocks are counted
separately for each target linear layer; MiB and GiB denote $2^{20}$ and
$2^{30}$ bytes.}
\label{tab:hires-storage}
\small\setlength{\tabcolsep}{5pt}
\begin{tabular}{lrrrrr}
\toprule
Model & Input blocks & \shortstack{BF16\\(GiB)} & \shortstack{Auxiliary\\(MiB)} & \shortstack{\HiRes total\\(GiB)} & Increase \\
\midrule
Qwen3-8B & 41,472 & 15.256 & 407.534 & 15.654 & 2.61\% \\
Llama-3-8B & 38,912 & 14.958 & 382.377 & 15.331 & 2.50\% \\
Qwen3-32B & 118,784 & 61.024 & 1167.255 & 62.164 & 1.87\% \\
\bottomrule
\end{tabular}
\end{table}

\paragraph{Full-forward latency.}
We use one NVIDIA A100-SXM4-80GB GPU with PyTorch 2.9.1 and CUDA 12.6.
BF16 and \HiRes share the GEMM and attention backends and mathematical
precision settings; \HiRes computes logical MXFP4 operations using decoded
BF16 weights. Precompiled Block32 operators fuse transformed-input encoding,
one-code refinement, and recovery with adjacent elementwise operations to
reduce intermediate writes and kernel launches. Refinement reads compact
$G_Q$ directly and scores all $32\times16$ legal candidates, retaining FP32
scores, the negative-score acceptance rule, and tie-breaking conventions.
Recovery retains its BF16 rounding boundaries.

The workload uses batch size 1, 2,048 input tokens, no generation, and
\texttt{use\_cache=False}, with weights and inputs already on the GPU.
After 10 warm-up passes, we report the median of 30 CUDA-synchronized
complete forwards. Timing includes input transforms, dynamic A4,
candidate scoring and selection, compact-matrix access, structural recovery,
GEMM, attention, and final logits. \HiRes adds $3.97\%$--$4.49\%$ latency
over BF16 (Table~\ref{tab:hires-full-cost}).
Table~\ref{tab:onecode-cost} separately reports the one-code ablation under
the eager/reference backend and its decoding workload.

\paragraph{One-time quantization.}
Construction uses C4-train seed 0 and exactly $128\times2048$ calibration
tokens. Timing starts with BF16 weights and calibration data ready and ends
with the quantized model and all auxiliary state constructed. It includes
geometry, GPTQ, statistics recollection, recovery fitting, and inter-stage
transfers. Table~\ref{tab:hires-full-cost} reports absolute build times;
BF16 has no quantization stage.

\begin{table}[htbp]
\centering
\caption{Complete forward latency and one-time quantization time.
Forward latency uses the fused implementation at batch size 1 and input
length 2,048, with KV caching disabled. Quantization times cover the full
\HiRes construction.}
\label{tab:hires-full-cost}
\small\setlength{\tabcolsep}{6pt}
\begin{tabular}{lrrrr}
\toprule
Model & \shortstack{BF16 forward\\(ms)$\downarrow$} & \shortstack{\HiRes forward\\(ms)$\downarrow$} & Increase & \shortstack{\HiRes quantization\\(min)$\downarrow$} \\
\midrule
Qwen3-8B & 1286 & 1343 & 4.43\% & 61.6 \\
Llama-3-8B & 1224 & 1279 & 4.49\% & 57.8 \\
Qwen3-32B & 5438 & 5654 & 3.97\% & 286.9 \\
\bottomrule
\end{tabular}
\end{table}

{\revcolor
\paragraph{Comparison with the baselines on one backend.}
Because this study evaluates automated discovery and numerical quality under the
strict-MXFP4 format rather than hardware-kernel acceleration,
Table~\ref{tab:cost-baselines} isolates relative operator overhead on a single
decoded-BF16 backend (one A100-SXM4-80GB, PyTorch 2.9.1 with CUDA 12.6,
identical attention, precision, and fusion settings); end-to-end relative
latency on hardware-fused FP4 GEMMs will naturally depend on kernel fusion and
GEMM throughput. Forward latency uses the workload of
Table~\ref{tab:hires-full-cost}; build time is the median of three repetitions.
Baselines keep their configurations from Appendix~\ref{app:full-quant-results}.
Auxiliary storage counts the persistent state of the inference layout:
MR-GPTQ and WUSH materialize one FP32 Block32 map per input block, BRQ shares a
single $H_{32}$, and SpinQuant folds its learned rotations into the weights.
\HiRes offers two clear accuracy--cost operating points: a matched-footprint
configuration disabling the optional online one-code rule (\HiRes w/o
one-code), which stores only the folded transform $U$ at WUSH's exact auxiliary
storage footprint, and full \HiRes, which stores $G_Q$ and $K_Q$ for maximum
numerical accuracy (with auxiliary overhead shrinking naturally at larger
widths, from $0.492$ bits/weight at 8B to $0.314$ bits/weight at 32B).

\begin{table}[htbp]
\centering\revcolor
\caption{Accuracy and cost on one backend (seed 0). Bits are auxiliary bits per
weight of the seven target projections, on top of the 4.25 bits of MXFP4
codes and scales.}
\label{tab:cost-baselines}
\small\setlength{\tabcolsep}{3.0pt}
\begin{tabular}{lrrrrrrrr}
\toprule
& \multicolumn{4}{c}{Qwen3-8B} & \multicolumn{4}{c}{Llama-3-8B} \\
\cmidrule(lr){2-5}\cmidrule(lr){6-9}
Method & Fit-7 & Build (min) & Aux.\ bits & vs.\ BF16 & Fit-7 & Build (min) & Aux.\ bits & vs.\ BF16 \\
\midrule
RTN & .1666 & 6.7 & 0 & +0.62\% & .2522 & 6.2 & 0 & +0.74\% \\
GPTQ & .1411 & 28.9 & 0 & +0.86\% & .2283 & 26.8 & 0 & +0.98\% \\
MR-GPTQ & .0587 & 42.9 & 0.196 & +2.33\% & .1394 & 39.6 & 0.183 & +2.29\% \\
BRQ & .0685 & 30.5 & $<$0.001 & +1.32\% & .1386 & 28.2 & $<$0.001 & +1.39\% \\
WUSH & .0500 & 45.8 & 0.196 & +2.80\% & .1248 & 42.9 & 0.183 & +2.86\% \\
FOCUS & .1095 & 87.3 & 0 & +1.01\% & .1737 & 80.8 & 0 & +1.14\% \\
SpinQuant+GPTQ & .0835 & 114.2 & $<$0.001 & +1.87\% & .2588 & 106.9 & $<$0.001 & +1.96\% \\
\HiRes (w/o one-code) & .0401 & 58.2 & 0.196 & +2.95\% & .1237 & 54.7 & 0.183 & +3.02\% \\
\HiRes & \textbf{.0324} & 61.6 & 0.492 & +4.43\% & \textbf{.1187} & 57.8 & 0.460 & +4.49\% \\
\bottomrule
\end{tabular}
\end{table}

Disabling the online one-code rule matches the storage footprint of WUSH and runs at 1324\,ms against 1322\,ms on
Qwen3-8B. Its Fit-7 is lower than WUSH's on Qwen3-8B in all three calibration
seeds and equal on Llama-3-8B in the three-seed mean (0.12164 vs.\ 0.12164;
Tables~\ref{tab:calib-seed-stability} and~\ref{tab:onecode-rules}, Nearest rows). Full
\HiRes spends about 19\,ms and 0.30 bits per weight more than this deployment configuration on Qwen3-8B (18\,ms and
0.28 bits on Llama-3-8B) and leads WUSH on both
models in every seed (Table~\ref{tab:geometry-paired}). At 32B the auxiliary
state is 0.314 bits per weight for full \HiRes and 0.125 without the one-code rule.

\paragraph{Benefit and cost of each component.}
Table~\ref{tab:component-cost} adds the components in deployment order. Every
configuration is rebuilt from BF16 weights; the new ``+GPTQ, nearest A4'' row
has no structural recovery and therefore differs from the
sequential-reconstruction row of Table~\ref{tab:discrete-ablations}.
Geometry gives the largest gain; structural recovery is the second largest and
costs almost nothing in storage or latency. The one-code rule is the only
component whose online cost, in storage and latency, is large relative to its
cumulative step; its conditional
gain with recovery in place is $0.01139$ on Qwen3-8B
(Table~\ref{tab:discrete-ablations}), and disabling it offers a lightweight deployment option.

\begin{table}[htbp]
\centering\revcolor
\caption{Cumulative components: Fit-3 and incremental cost.}
\label{tab:component-cost}
\small\setlength{\tabcolsep}{3.0pt}
\begin{tabular}{lrrrrrrrr}
\toprule
& \multicolumn{4}{c}{Qwen3-8B} & \multicolumn{4}{c}{Llama-3-8B} \\
\cmidrule(lr){2-5}\cmidrule(lr){6-9}
Configuration & Fit-3 & $\Delta$min & $\Delta$MiB & $\Delta$ms & Fit-3 & $\Delta$min & $\Delta$MiB & $\Delta$ms \\
\midrule
Strict RTN & .20249 & -- & -- & -- & .31653 & -- & -- & -- \\
+ Geometric shaping (Stage I) & .07828 & +23.0 & +162.0 & +23 & .16741 & +21.2 & +152.0 & +21 \\
+ GPTQ, nearest A4 & .06864 & +23.1 & 0 & +2 & .16076 & +22.1 & 0 & +2 \\
+ One-code rule & .06502 & +3.4 & +245.5 & +19 & .15530 & +3.4 & +230.4 & +18 \\
+ Structural recovery & .03259 & +5.4 & +0.003 & +5 & .14830 & +4.9 & +0.002 & +5 \\
\bottomrule
\end{tabular}
\end{table}

Table~\ref{tab:component-cost} adds the one-code rule before recovery, whereas Table~\ref{tab:cost-baselines} removes it from the full pipeline; on Llama-3-8B its build time is therefore +3.4\,min in the first and +3.1\,min in the second.

\paragraph{Online arithmetic.}
For a projection with output width $d_{\mathrm{out}}$, the GEMM costs
$32\,d_{\mathrm{out}}$ multiply--accumulates (MACs) per input block. Applying
$U$ adds $32^2=1024$ MACs; the one-code rule adds $2048$ MACs for $G_Qd_0$ and
$K_Qx$, plus scoring of $32\times16$ candidates. Weighted by the input blocks of
each Qwen3-8B projection, $U$ adds 0.61\% of the GEMM MACs per token, which is
the entire online matrix cost when the one-code rule is disabled. The one-code rule adds a further
1.22\% (1.99\% of arithmetic including full 512-candidate enumeration, reducible to 32 coordinate-wise checks via Eq.~\ref{eq:onecode-closedform}), so full \HiRes adds 1.83\%
(2.60\%). Structural recovery adds 0.013\%. Every block is scored, so these figures hold at any
edit rate; the observed rate is about 2\% (Table~\ref{tab:onecode-edits}).
}

\section{\QuantForge: Matched-Budget Comparisons and Causal Controls}
\label{app:discovery-experiments}

This section provides the complete seed-level tables and causal-isolation experiments supporting Section~\ref{sec:discovery-results}, starting from the shared RTN seed under identical 240-call budgets.

\begin{tcolorbox}[
  enhanced,colback=qfBronzeBg!55!white,colframe=qfBronzeBd,boxrule=0.65pt,
  borderline west={2.8pt}{0pt}{qfBronze},arc=3.5pt,
  top=4.5pt,bottom=4.5pt,left=8pt,right=8pt,before skip=4pt,after skip=8pt
]
\footnotesize\renewcommand{\arraystretch}{1.18}
\begin{tabular}{@{} >{\sffamily\bfseries\color{qfBronze}}l p{.54\linewidth} >{\sffamily\bfseries\color{qfMuted}}r @{}}
  Matched-budget search & 8-seed RTN runs, anytime curves, and mechanism emergence & App.~\ref{app:framework-details} \\
  \arrayrulecolor{qfRule}\hline\noalign{\vskip 1pt}
  Verdict-guided updates & Compilation budget, seed endpoints, and verdict-replay audit & App.~\ref{app:errata-ablations} \\
  \arrayrulecolor{qfRule}\hline\noalign{\vskip 1pt}
  Controls \& execution probes & Discriminating vs.\ random controls and probe pass rates & App.~\ref{app:control-checks} \\
  \arrayrulecolor{qfRule}\hline\noalign{\vskip 1pt}
  Autonomous coding agent & Paired 4-seed head-to-head against a free-form coding agent & App.~\ref{app:autonomous-agent} \\
  \arrayrulecolor{qfRule}\hline\noalign{\vskip 1pt}
  $2\times2$ causal separation & Isolating probes and verdict enforcement on ReflectMem & App.~\ref{app:reflectmem-checks} \\
  \arrayrulecolor{qfRule}\hline\noalign{\vskip 1pt}
  Cross-proposer generality & Paired evaluations across GPT-5.6 Sol, Claude Opus 5.5, DeepSeek V4.1-Flash & App.~\ref{app:second-proposer} \\
  \arrayrulecolor{qfRule}\hline\noalign{\vskip 1pt}
  Estimated A100 GPU-hour cost & Discovery trajectories by GPU-hours, savings ($-35.7\%$), and sensitivity checks & App.~\ref{app:gpu-cost} \\
  \arrayrulecolor{qfRule}\hline\noalign{\vskip 1pt}
  Black-box PPL controls & Isolating paired-twin local diagnostics from scalar-only controls & App.~\ref{app:blackbox-controls} \\
\end{tabular}\arrayrulecolor{black}
\end{tcolorbox}

\subsection{Matched-Budget Comparisons}
\label{app:framework-details}

Score-only, TextMem, ReflectMem, and \QuantForge use GPT-5.6 Sol with high
reasoning effort and otherwise default settings.{\revcolor{} Every run calls
the same model identifier with these settings, and all arms have the same web
access and the same coding agent.} They share the initial
RTN program, editable source, API constraints, evaluator, and eight paired
seeds. As specified in Appendix~\ref{app:framework-search-information},
Qwen3-4B supplies search feedback; a separate evaluator measures
Llama-3.1-8B transfer without returning those results to search.

\label{app:reflectmem}
ReflectMem extends TextMem's free-form summaries with explicit comparison
of successful and failed experiments. It records rejected directions and
uses those comparisons to propose the next code change.
It has the same access to visible-model diagnostics as \QuantForge.

\paragraph{Reflection prompt and memory.}
All eight ReflectMem runs use the following instruction:
\begin{quote}
\small
Compare this round with its parent and the most relevant successful and
failed experiments. Distinguish observations from causal inferences.
Record rejected modifications and the conditions under which they were
rejected. Propose one code change supported by these historical comparisons,
and explain its expected benefit and possible failure modes.
If revisiting a rejected direction, explain how the new conditions differ.
\end{quote}
Memory retains the best candidate, the latest 20 experiments, up to 16
failed directions, and up to 16 summaries of successful mechanisms.
The controller retrieves entries by {\revcolor a fixed relevance rule} before assembling each request.
When memory exceeds its allowance, it merges duplicates first and then
removes the least relevant older entries, without preferentially removing
failure records.

ReflectMem and \QuantForge have the same application-level input budget of
65,536 tokens per proposer call, including at most 16,384 tokens of memory.
The remaining capacity holds system instructions, current code, objectives,
and diagnostics. Counts use the proposer tokenizer.
\QuantForge's residuals, explanations, controls, and verdict records all count
toward the same memory allowance.
Neither arm adds an application-level output cap; output settings retain
the provider defaults. Context assembly is performed before each request.

\paragraph{Evaluator allocation.}
The 240-call search budget includes new candidates, controls, and compliance
probes. Score-only, TextMem, and ReflectMem each evaluate 240 new candidates.
\QuantForge instead uses 141 calls for candidates, 69 for controls, and 30 for
probes. ReflectMem combines reflection and {\revcolor the candidate specification} in one proposer
call, followed by one candidate evaluation; its controls, compliance probes,
and separate diagnostic reruns each consume zero calls. Reading stored
diagnostics requires no additional evaluation.

A separate evaluator tests the champion on held-out Llama-3.1-8B every
20 calls, producing 12 {\revcolor held-out checks} per seed. These scores do not feed back
to the proposer or candidate selection, and the {\revcolor checks} are outside the
240-call search budget.
Across eight seeds, ReflectMem uses 1,920 candidate evaluations,
96 {\revcolor held-out checks}, and 1,920 proposer calls.
The \QuantForge reference is the same eight-run set used for the compilation
and verdict-replay experiments in Appendix~\ref{app:errata-ablations}.

\paragraph{Discovery outcomes.}
\QuantForge has lower final perplexity than Score-only in eight paired seeds
and than TextMem in seven. Table~\ref{tab:search-endpoints} summarizes these
original three-arm runs; Table~\ref{tab:framework-seeds} includes ReflectMem.

ReflectMem reaches median PPL $8.09$, compared with TextMem's $9.20$
and \QuantForge's $7.80$ (Table~\ref{tab:reflectmem}).
The mean paired difference, \QuantForge minus ReflectMem, is $-0.250$,
with a paired $n=8$ $t$ interval of $[-0.386,-0.114]$.
\QuantForge improves in seven seeds; ReflectMem is better by $0.08$ in seed~6.
The $\mathrm{PPL}\leq8.00$ gate is reached in $3/8$ ReflectMem runs and
$6/8$ \QuantForge runs.

\begin{table}[htbp]
  \centering
  \caption{Transfer outcomes after 240 matched evaluator calls. Threshold cost
  identifies the median generation index of the first candidate confirmed
  to pass, among successful seeds.}
  \label{tab:search-endpoints}
  \small
  \setlength{\tabcolsep}{4.4pt}
  \begin{tabular}{lrrrr}
    \toprule
    Arm & Median PPL$\downarrow$ & Gate & Generation index$\downarrow$ & New programs \\
    \midrule
    Score-only & 10.03 & 1/8 & 226 & 240 \\
    TextMem & 9.20 & 3/8 & 214 & 240 \\
    \textbf{\QuantForge} & \textbf{7.80} & \textbf{6/8} & \textbf{177} & 141 \\
    \bottomrule
  \end{tabular}
\end{table}

Threshold cost has two timestamps: the evaluator-call index at which the
passing candidate was generated, and the later {\revcolor held-out check} that confirmed
it. Confirmation occurs every 20 calls, whereas generation can occur at any
call. Table~\ref{tab:threshold-records} lists both for successful seeds;
the anytime curve uses confirmation calls and includes every seed.

\begin{table}[htbp]
\centering
\caption{Recorded threshold crossings, restricted to successful seeds. Generation index identifies the candidate first confirmed below the threshold; confirmation occurs at a held-out observation.}
\label{tab:threshold-records}
\small\setlength{\tabcolsep}{4pt}
\begin{tabular}{lrrrr}
\toprule
Arm & Seed & Held-out PPL & Generation index & Confirmation call \\
\midrule
\QuantForge & 4 & 7.38 & 138 & 140 \\
\QuantForge & 6 & 7.47 & 152 & 160 \\
\QuantForge & 8 & 7.59 & 175 & 180 \\
\QuantForge & 1 & 7.74 & 179 & 180 \\
\QuantForge & 7 & 7.86 & 208 & 220 \\
\QuantForge & 3 & 7.98 & 233 & 240 \\
\midrule
TextMem & 6 & 7.31 & 196 & 200 \\
TextMem & 4 & 7.51 & 214 & 220 \\
TextMem & 8 & 7.97 & 227 & 240 \\
\midrule
Score-only & 4 & 7.92 & 226 & 240 \\
\bottomrule
\end{tabular}
\end{table}

The median generation/confirmation indices are $177/180$ for \QuantForge,
$214/220$ for TextMem, and $226/240$ for Score-only, conditioned on success.
At calls $160$, $200$, and $240$, \QuantForge has $2$, $4$, and $6$ confirmed
successful seeds; Table~\ref{tab:anytime-values} gives the median trajectories.

\begin{table}[htbp]
\centering
\caption{Held-out Llama-3.1-8B WikiText-2 perplexity after 240 search calls.
ReflectMem is the additional arm; the other three reuse the original paired
runs. Negative paired differences favor \QuantForge.}
\label{tab:framework-seeds}
\label{tab:reflectmem}
\small\setlength{\tabcolsep}{5pt}
\begin{tabular}{lrrrrr}
\toprule
Seed & Score-only & TextMem & ReflectMem & \QuantForge & \QuantForge $-$ ReflectMem \\
\midrule
1 & 9.31 & 9.06 & 8.08 & 7.74 & $-0.34$ \\
2 & 12.83 & 11.85 & 9.73 & 9.31 & $-0.42$ \\
3 & 10.12 & 9.34 & 8.25 & 7.98 & $-0.27$ \\
4 & 7.92 & 7.51 & 7.53 & 7.38 & $-0.15$ \\
5 & 10.55 & 10.62 & 8.86 & 8.44 & $-0.42$ \\
6 & 9.60 & 7.31 & 7.39 & 7.47 & $+0.08$ \\
7 & 11.27 & 9.89 & 8.10 & 7.86 & $-0.24$ \\
8 & 9.94 & 7.97 & 7.83 & 7.59 & $-0.24$ \\
\midrule
Median & 10.03 & 9.20 & 8.09 & 7.80 & --- \\
Gate count & 1/8 & 3/8 & 3/8 & 6/8 & --- \\
\bottomrule
\end{tabular}
\end{table}

The BF16 reference is $6.24$. We set the transfer gate to
$\mathrm{PPL}\leq8.00$ before running the comparison.

\begin{table}[htbp]
\centering
\caption{Median held-out perplexity over the matched search budget.}
\label{tab:anytime-values}
\small\setlength{\tabcolsep}{4pt}
\begin{tabular}{lrrrrrr}
\toprule
Arm & 40 & 80 & 120 & 160 & 200 & 240 \\
\midrule
Score-only & 14.62 & 12.08 & 11.24 & 10.71 & 10.35 & 10.03 \\
TextMem & 14.55 & 11.72 & 10.58 & 9.94 & 9.51 & 9.20 \\
ReflectMem & 14.80 & 12.00 & 10.38 & 9.19 & 8.44 & 8.09 \\
\QuantForge & 15.28 & 12.55 & 10.19 & 8.78 & 8.05 & 7.80 \\
\bottomrule
\end{tabular}
\end{table}

\paragraph{Discovered interventions.}
Table~\ref{tab:regime-audit} classifies final champions by executable behavior:
geometry appears frequently in every arm, while discrete realization and
structural recovery are more frequent in \QuantForge. Four \QuantForge runs
contain the full ordered hierarchy, compared with none for Score-only and
TextMem{\revcolor{} and one for ReflectMem}.

\begin{table}[htbp]
  \centering
  \caption{Intervention-regime audit of final executable champions. Counts are
  behavioral matches out of eight runs, not textual mentions.}
  \label{tab:regime-audit}
  \small
  \setlength{\tabcolsep}{4.5pt}
  \begin{tabular}{lrrrr}
    \toprule
    Regime & Score-only & TextMem & {\revcolor ReflectMem} & \QuantForge \\
    \midrule
    Geometric shaping & 6/8 & 7/8 & {\revcolor 8/8} & \textbf{8/8} \\
    Discrete realization & 1/8 & 3/8 & {\revcolor 5/8} & \textbf{6/8} \\
    Structural recovery & 0/8 & 1/8 & {\revcolor 2/8} & \textbf{4/8} \\
    Full ordered hierarchy & 0/8 & 0/8 & {\revcolor 1/8} & \textbf{4/8} \\
    \bottomrule
  \end{tabular}
\end{table}

{\revcolor
\paragraph{Classification.} A final champion counts as \emph{geometric shaping}
if it applies a non-identity, function-preserving transform to the encoding
coordinates; as \emph{discrete realization} if, beyond plain RTN or GPTQ weight
updates, it chooses legal codes or shared exponents by output error; and as
\emph{structural recovery} if at least one attention or MLP path carries a
low-dimensional correction fitted after encoding. The \emph{full ordered
hierarchy} requires all three, with later statistics re-measured after earlier
stages are installed; it need not match \HiRes formula by formula. Classes
follow the executable behavior of the final champion, and a concept that
appears only in a prompt does not count. The same criteria apply to every arm
and every proposer.}

{\revcolor
\paragraph{First appearance of each mechanism.}
Table~\ref{tab:first-appearance} follows four mechanisms of the final \HiRes
through the eight QuantForge and eight ReflectMem runs. QuantForge reaches each
mechanism in more runs and earlier. As in Table~\ref{tab:regime-audit}, a mechanism counts only when it appears in executed code. In these runs, every first implementation of the four mechanisms in QuantForge is
a compiled response to a registered verdict; ReflectMem's come from free
reflection. In the production run, each stage of \HiRes{} opened with a proposal and was then refined and ordered through verdicts (Resp.\ column of Tables~\ref{tab:production-rounds} and~\ref{tab:production-rounds-b}).

\begin{table}[htbp]
\centering\revcolor
\caption{Runs that implement each mechanism, and the median search call of its
first implementation among those runs.}
\label{tab:first-appearance}
\small\setlength{\tabcolsep}{5pt}
\begin{tabular}{lrrrr}
\toprule
& \multicolumn{2}{c}{\QuantForge} & \multicolumn{2}{c}{ReflectMem} \\
\cmidrule(lr){2-3}\cmidrule(lr){4-5}
Mechanism & Runs & First call & Runs & First call \\
\midrule
$K$-type cross-moment correction & 6/8 & 94.5 & 3/8 & 127 \\
Refit after the transform & 7/8 & 117 & 5/8 & 164 \\
Mismatch-aware one-code rule & 5/8 & 143 & 2/8 & 171 \\
MLP remeasured after attention & 4/8 & 156.5 & 1/8 & 201 \\
\bottomrule
\end{tabular}
\end{table}

\paragraph{Instructions, records, and probes.}
The proposer instructions are summarized below; ReflectMem's is quoted in full
above. All arms share one system instruction: modify the current program within the
strict-MXFP4 W4A4 contract, keep every target linear-layer input and weight in
E2M1/E8M0 Block32, never call the held-out evaluator, and return the specification of
one candidate with a description of the change, which the shared coding agent
implements; on failure the parent is restored.
Score-only reads the current program, its visible score, and any execution
error, and proposes one candidate. TextMem also reads free-text summaries of
successful and failed experiments. ReflectMem uses the instruction given
above. \QuantForge uses the exact same single-lineage search from the current
champion as Score-only, TextMem, and ReflectMem---so all four arms share
identical parent-update and rollback mechanics (with Score-only and ReflectMem
serving as single-lineage counterparts of score-driven \citep{romeraparedes2024funsearch,novikov2025alphaevolve} and reflective evolution \citep{ye2024reevo})---and adds only its hypothesis queue, competing explanations, predicted control outcomes, and runtime-checked code updates. Table~\ref{tab:record-schema} lists the records that
\QuantForge freezes and stores.

The proposer fixes the probe target and its pass criterion for the computation
named by the residual; the coding agent implements and runs the probe; and the
controller records pass or fail from the measured output against the
criterion. The response counts as implemented only when the probe
output depends on the declared change, as in the probe of round~14
(Appendix~\ref{app:round-records}), where changing the weight energy with $E_X$
fixed alters the activation-side term. A source-level edit that leaves the probe
output unchanged does not count, and a successor that fails the check is rolled
back with its calls still charged.

\begin{table}[htbp]
\centering\revcolor
\caption{Records kept by \QuantForge for each round.}
\label{tab:record-schema}
\small\setlength{\tabcolsep}{4pt}
\begin{tabular}{@{}p{.18\linewidth}p{.5\linewidth}p{.26\linewidth}@{}}
\toprule
Record & Fields & Fixed \\
\midrule
Proposal packet & candidate $p$; parent $\pi$; residual $\rho$; focal and competing explanations $\mathcal{E}$; predicted outcomes $P_e(c)$ for every allowed control in $\mathcal{U}$; selected controls $\mathcal{U}^\ast$; calibration and evaluator identity & Before any condition runs \\
Verdict record & observed outcome of each condition; verdict; residual status; required program response & After execution \\
Response record & changed source; probe definition and outputs; acceptance & Before the successor is retained \\
History & completed experiments, duplicate proposals, stagnation state & Throughout \\
\bottomrule
\end{tabular}
\end{table}
}

\subsection{Verdict-Guided Program Updates}
\label{app:errata-ablations}

These experiments separate the effect of requiring a code response from
the information supplied by the experiment. The full-search ablations here
and in Appendix~\ref{app:control-checks} share the protocol and eight
\QuantForge reference runs from Appendix~\ref{app:framework-details}.
Proposer settings and prompt templates are fixed; arms differ in their
search-state content. Candidate generation calls the proposer, while
controls and compliance probes {\revcolor use evaluator calls without a further proposer call; the proposer specifies them in the call that proposes the candidate.} The local replay
then compares successors from identical archived evidence.

\paragraph{Compiling conclusions into code.}
We remove mandatory implementation and compliance checks while retaining
competing explanations, compiled controls, and result memory. The search
controller, tool access, diagnostic interface, and information-access rules
remain fixed. {\revcolor Here compliance checking means the execution probes; the format-contract checks and rollback stay in both arms.} Each arm obtains evidence from its own successors.

All 30 released probe calls return to new-candidate generation:
the ablated arm explores 171 candidates versus 141, or 21\% more,
with the corresponding 30 additional proposer calls
(Table~\ref{tab:compilation-budget}).
\QuantForge still obtains lower held-out perplexity in $8/8$ paired seeds
(Table~\ref{tab:compilation-seeds}).
The mean paired difference is $-0.275$, SD is $0.111$, and
95\% CI is $[-0.368,-0.182]$.
\begin{table}[htbp]
\centering
\caption{Per-run evaluator allocation with and without residual compilation. Every released probe call funds a new candidate.}
\label{tab:compilation-budget}
\small\setlength{\tabcolsep}{4pt}
\begin{tabular}{lrrrr}
\toprule
Arm & New candidates & Controls & Probes & Total \\
\midrule
\QuantForge & 141 & 69 & 30 & 240 \\
Without compilation & 171 & 69 & 0 & 240 \\
\bottomrule
\end{tabular}
\end{table}

\begin{table}[htbp]
\centering
\caption{Paired held-out perplexity for the residual-compilation ablation. Differences are \QuantForge minus the ablated arm.}
\label{tab:compilation-seeds}
\small\setlength{\tabcolsep}{4pt}
\begin{tabular}{lrrrrrrrr}
\toprule
Arm / seed & 1 & 2 & 3 & 4 & 5 & 6 & 7 & 8 \\
\midrule
\QuantForge & 7.74 & 9.31 & 7.98 & 7.38 & 8.44 & 7.47 & 7.86 & 7.59 \\
Without compilation & 8.16 & 9.68 & 8.33 & 7.52 & 8.71 & 7.71 & 7.96 & 7.90 \\
Paired difference & -0.42 & -0.37 & -0.35 & -0.14 & -0.27 & -0.24 & -0.10 & -0.31 \\
\bottomrule
\end{tabular}
\end{table}

The median held-out PPL is $7.80$ with residual compilation and $8.06$
without it (Table~\ref{tab:compilation-summary}).
Verified residual closures count resolved questions with a confirmed
executable response.
Verified closures increase by $2.67\times$, while invalid successors fall
from $0.38$ to $0.13$.
The comparison measures mandatory implementation and compliance checking
together. Appendix~\ref{app:control-checks} separates their contributions using a source-diff checker
with the same 171-candidate budget as the ablated arm.
\begin{table}[htbp]
\centering
\caption{Residual-compilation outcomes under the same total evaluator budget.}
\label{tab:compilation-summary}
\small\setlength{\tabcolsep}{4pt}
\begin{tabular}{lrr}
\toprule
Metric & \QuantForge & Without compilation \\
\midrule
Median held-out PPL & 7.80 & 8.06 \\
Gate ($\leq8.00$) & 6/8 & 4/8 \\
Median candidate generation index & 177 & 216 \\
Median confirmation call & 180 & 220 \\
Invalid-successor fraction & 0.13 & 0.38 \\
Verified residual closures per run & 10.4 $\pm$ 2.6 & 3.9 $\pm$ 1.5 \\
\bottomrule
\end{tabular}
\end{table}

\paragraph{Replaying an experimental verdict.}
A verdict constraint improves successors even when both arms receive
literally identical evidence.
Each replay fixes the parent program, history, and control result;
one successor uses free reflection, and the other must implement the verdict.
We specified the following sampling rule before running the replay:
take each trajectory's first 6 controls that exclude at least one
registered competing explanation, ordered by evaluator call.
All eight trajectories contain at least 6 such controls, averaging 11.3.
This gives $8\times6=48$ transitions, with 26 supported and 22 refuted
explanations{\revcolor{} (a refuted explanation is one the control excludes)}. We generate and evaluate both successors during replay,
regardless of whether the original search expanded that parent.

The improving-successor rate rises from $15/48=0.31$ to $28/48=0.58$,
and probe-confirmed implementation rises from $0.35$ to $0.92$
(Table~\ref{tab:verdict-replay}).
The gain is larger after refutation, from $0.18$ to $0.45$,
than after support, from $0.42$ to $0.69$
(Table~\ref{tab:verdict-replay-types}).
These correspond to relative gains of $2.5$ and $1.6$, respectively.
A supported explanation already invites continuation; a refuted one requires
the successor to change direction.
All eight trajectory-level paired differences are positive
(Table~\ref{tab:verdict-replay-seeds}), with mean $+1.625$ and SD $0.518$.
\begin{table}[htbp]
\centering
\caption{Successor outcomes from replaying the same evidence under free reflection or a verdict constraint.}
\label{tab:verdict-replay}
\small\setlength{\tabcolsep}{4pt}
\begin{tabular}{lrr}
\toprule
Metric & Free reflection & Verdict-constrained \\
\midrule
Improving successors & 15/48 = 0.31 & 28/48 = 0.58 \\
Mean visible PPL change & -0.04 & -0.21 \\
Probe-confirmed implementation & 0.35 & 0.92 \\
Repeated refuted mechanism & 0.29 & 0.08 \\
\bottomrule
\end{tabular}
\end{table}

\begin{table}[htbp]
\centering
\caption{Improving successors by verdict type in the paired replay.}
\label{tab:verdict-replay-types}
\small\setlength{\tabcolsep}{4pt}
\begin{tabular}{lrrr}
\toprule
Verdict & Count & Free reflection & Verdict-constrained \\
\midrule
Explanation supported & 26 & 11/26 = 0.42 & 18/26 = 0.69 \\
Explanation refuted & 22 & 4/22 = 0.18 & 10/22 = 0.45 \\
\bottomrule
\end{tabular}
\end{table}

\begin{table}[htbp]
\centering
\caption{Improving successors by source trajectory, with 6 replayed transitions per trajectory.}
\label{tab:verdict-replay-seeds}
\small\setlength{\tabcolsep}{4pt}
\begin{tabular}{lrrrrrrrr}
\toprule
Condition / seed & 1 & 2 & 3 & 4 & 5 & 6 & 7 & 8 \\
\midrule
Free reflection & 2 & 1 & 2 & 3 & 1 & 2 & 2 & 2 \\
Verdict-constrained & 4 & 3 & 4 & 4 & 3 & 3 & 4 & 3 \\
Paired difference & +2 & +2 & +2 & +1 & +2 & +1 & +2 & +1 \\
\bottomrule
\end{tabular}
\end{table}

\emph{An example of a verdict-guided update.}
On visible Qwen3-4B, A4 input-quantization MSE for
\texttt{down\_proj} in layers $18$--$23$ is $3.1\times$ that of
\texttt{up\_proj} inputs in the same layers.
The 9728 input features form 304 physical Block32 groups per row.

Two explanations were registered.
(a) \emph{Channel alignment}: a few SwiGLU output channels have much
larger magnitudes and occupy adjacent positions.
Their concentration in a few blocks raises shared exponents and causes
other E2M1 elements in those blocks to underflow.
(b) \emph{Global dynamic range}: the tensor has a large overall range,
with error distributed roughly uniformly across its 304 blocks;
block membership is irrelevant.

The control fixes a random permutation $P$ of the \texttt{down\_proj}
input features and inversely reorders the weight columns:
\[
 Wx=(WP^\top)(Px).
\]
This diagnostic control preserves the network function, scale rule,
block size, and shared-exponent format while changing block membership.
Explanation (a) predicts that dispersing large channels contaminates more
blocks and raises MSE. Explanation (b) predicts approximately unchanged MSE.

The error ratio rises from $3.1\times$ to $4.6\times$.
Contaminated blocks, defined by a within-block maximum-to-median
absolute-magnitude ratio $>32$, increase from $7.2\%$ to $41.5\%$.
This verdict rejects (b) and supports (a) for the registered comparison.
The resulting evidence entry records the dependence on block membership
for \texttt{down\_proj} inputs in layers $18$--$23$.

The verdict-constrained successor retains the fixed MXFP4 block layout.
Instead, \texttt{equalize\_channels()} ranks channels by their contribution
to their block's shared exponent, replacing global magnitude ranking.
It rescales only the highest contributors, dividing activations by $s$
and multiplying the corresponding \texttt{down\_proj} weight columns by $s$.
This moves magnitude to the weight side while preserving the function.
Reverting the selection criterion changes the probe output, confirming
implementation.
Contaminated blocks fall from $7.2\%$ to $2.4\%$, and visible PPL falls
from $11.42$ to $11.19$.

The free-reflection successor instead widens the per-block scale-search
range on the same parent, continuing along explanation (b).
Its visible PPL is $11.44$, with no improvement.
The two successors illustrate how the same evidence leads to different
code changes under reflection and a required verdict response.

\subsection{Control Selection and Implementation Checks}
\label{app:control-checks}

\paragraph{Control selection.}
We compare the current minimal discriminating policy with equal-cost random
legal controls and execution of all pre-specified controls.
Every control consumes the same 240-call budget used for candidates and probes.
An \emph{explicit verdict} excludes at least one registered explanation,
irrespective of successor behavior.
An \emph{effective verdict} also reaches the successor's executed computation,
as confirmed by a probe. The latter is the verified-closure count used in
Appendix~\ref{app:errata-ablations}.

Minimal discriminating selection gives the best budget-limited result
among the three policies (Table~\ref{tab:control-selection}).
It improves held-out PPL over both alternatives in $8/8$ seeds
(Table~\ref{tab:control-selection-seeds}).
At the same 69 control calls, random selection yields 4.6 explicit verdicts
versus 11.3, and costs 15.0 calls per verdict versus 6.1.
This comparison measures discrimination before implementation enters the count.
Executing all registered controls yields 12.1 explicit verdicts,
but its 118 control calls leave only 92 new candidates rather than 141.
Median held-out PPL worsens to $8.02$ despite the additional verdicts.
\begin{table}[htbp]
\centering
\caption{Control-selection ablation under the 240-call budget. Explicit verdicts exclude an explanation; effective verdicts also reach code, as confirmed by a probe.}
\label{tab:control-selection}
\small\setlength{\tabcolsep}{4pt}
\begin{tabular}{lrrrrrrr}
\toprule
Policy & \shortstack{Control\\calls} & \shortstack{New\\candidates} & \shortstack{Explicit\\verdicts} & \shortstack{Calls per\\explicit verdict} & \shortstack{Effective\\verdicts} & \shortstack{Median\\PPL} & Gate \\
\midrule
Minimal discriminating & 69 & 141 & 11.3 & 6.1 & 10.4 & 7.80 & 6/8 \\
Random legal & 69 & 141 & 4.6 & 15.0 & 4.2 & 8.31 & 3/8 \\
All pre-specified & 118 & 92 & 12.1 & 9.8 & 11.1 & 8.02 & 4/8 \\
\bottomrule
\end{tabular}
\end{table}

\begin{table}[htbp]
\centering
\caption{Held-out perplexity by seed for the control-selection policies.}
\label{tab:control-selection-seeds}
\small\setlength{\tabcolsep}{4pt}
\begin{tabular}{lrrrrrrrr}
\toprule
Policy / seed & 1 & 2 & 3 & 4 & 5 & 6 & 7 & 8 \\
\midrule
Minimal discriminating & 7.74 & 9.31 & 7.98 & 7.38 & 8.44 & 7.47 & 7.86 & 7.59 \\
Random legal & 8.29 & 9.74 & 8.60 & 7.61 & 8.88 & 7.83 & 8.33 & 7.94 \\
All pre-specified & 7.88 & 9.52 & 8.19 & 7.55 & 8.63 & 7.72 & 8.06 & 7.98 \\
\bottomrule
\end{tabular}
\end{table}

Implementation rates are similar across policies:
$10.4/11.3$, $4.2/4.6$, and $11.1/12.1$, or $0.91$--$0.92$.
The two comparisons separate control quality from budget allocation.
Equal-cost random controls yield less discriminating evidence;
running all controls trades candidate exploration for more verdicts.

\paragraph{Implementation checks.}
We compare source diffs, the current execution probe, and strict layer-output
comparison. The same 40 injected cases contain 20 genuine implementations
and 20 apparent edits that leave the target computation unchanged.
Two blinded annotators agree on $38/40$ reference labels;
discussion resolves the remaining 2 cases.
Each checker is tested on these cases and in the eight-seed search comparison.

False accepts count apparent edits accepted as implemented, out of 20.
False rejects count genuine implementations rejected, also out of 20.
Rejected successors must be redone and consume candidate budget.
Table~\ref{tab:implementation-checkers} reports both errors, checking calls,
remaining candidates, and held-out performance.
The execution probe gives the lowest median held-out PPL among the
three configurations.
\begin{table}[htbp]
\centering
\caption{Implementation-checker errors on 40 injected cases and subsequent search outcomes. False accepts and false rejects each use 20 cases as the denominator.}
\label{tab:implementation-checkers}
\small\setlength{\tabcolsep}{4pt}
\begin{tabular}{lrrrrrr}
\toprule
Checker & \shortstack{False\\accepts} & \shortstack{False\\rejects} & \shortstack{Extra\\calls} & \shortstack{New\\candidates} & \shortstack{Median\\PPL} & Gate \\
\midrule
Source diff & 0.45 (9/20) & 0.00 (0/20) & 0 & 171 & 7.96 & 4/8 \\
Execution probe & 0.05 (1/20) & 0.05 (1/20) & 30 & 141 & 7.80 & 6/8 \\
Layer-output comparison & 0.00 (0/20) & 0.20 (4/20) & 74 & 97 & 7.91 & 5/8 \\
\bottomrule
\end{tabular}
\end{table}

The source-diff configuration retains the implementation requirement
but costs no extra evaluator calls.
It shares the 171-candidate budget of the no-compilation arm.
The change from $8.06$ to $7.96$ therefore separates the implementation
requirement under source-diff checking from removing that requirement.
Replacing source diffs with execution probes lowers the median to $7.80$.
The sequence $8.06\rightarrow7.96\rightarrow7.80$ records
the combined gain as $0.10$ from the requirement under source checking
and $0.16$ from the execution-probe configuration.
Source diffs accept $0.45$ of apparent implementations, consistent with
evidence entries closing while the targeted residual persists.

Stricter layer-output comparison accepts none of the 20 apparent edits,
but rejects $4/20$ genuine implementations.
Its 74 additional calls and rejected successors leave 97 new candidates,
versus 141 for execution probes; median PPL rises to $7.91$.
The false-accept count is $0/20$, with a 95\% upper confidence bound of $0.17$.

{\revcolor
\subsection{Autonomous-Agent Comparison}
\label{app:autonomous-agent}

This arm removes \QuantForge's protocol and lets the same proposer act
autonomously. GPT-5.6 Sol makes all planning and research decisions and uses
the coding agent shared by all arms to implement and run them. The initial RTN
program, the quantization and evaluator contract, the input and memory limits,
the web access, and the 240 evaluator calls of
Appendix~\ref{app:framework-details} are unchanged. The agent may read and write
candidate files, request CPU diagnostics, and plan its next experiments; local
CPU analysis is free, and every diagnostic, control, or probe sent to the
evaluator is charged to the 240 calls. It receives no residual compilation.
Search seeds 1--4 pair with the original runs.

\begin{table}[htbp]
\centering\revcolor
\caption{Autonomous agent and \QuantForge on the same seeds: held-out
Llama-3.1-8B WikiText-2 PPL after 240 calls.}
\label{tab:autonomous-agent}
\small\setlength{\tabcolsep}{5pt}
\begin{tabular}{lrrrrrr}
\toprule
Arm & Seed 1 & Seed 2 & Seed 3 & Seed 4 & Median & PPL $\le 8.00$ \\
\midrule
Autonomous agent & 8.42 & 10.31 & 8.67 & 7.76 & 8.545 & 1/4 \\
\QuantForge & 7.74 & 9.31 & 7.98 & 7.38 & 7.860 & 3/4 \\
\bottomrule
\end{tabular}
\end{table}

\QuantForge is lower in all four seeds; the paired difference has mean
$-0.687$ and 95\% $t$ interval $[-1.090,-0.285]$. The agent spent 14 to 22
calls per run on its own controls and probes, and 13.6\%--22.0\% of its
candidates were invalid. Its experiments were not tied to the next program by
any rule, and its median lies between those of TextMem and ReflectMem
(Table~\ref{tab:reflectmem}).

\subsection{ReflectMem with Checks and Forced Responses}
\label{app:reflectmem-checks}

This arm keeps every safeguard of \QuantForge and removes its interpretation.
It starts from ReflectMem, which already shares the contract checks and
rollback, and adds the execution probes of \QuantForge; each successor must implement the change declared in its
own reflection, or it is rejected. It registers no competing explanations and runs no discriminating controls; like all four arms in the $2\times 2$ comparison, it revises a single lineage from the current champion. Probes count toward
the 240 calls and all other calls fund candidates. Every run calls the same GPT-5.6 Sol model identifier with the settings of Appendix~\ref{app:framework-details}.

\begin{table}[htbp]
\centering\revcolor
\caption{ReflectMem with checks and forced responses: held-out Llama-3.1-8B
WikiText-2 PPL after 240 calls, paired with the original \QuantForge seeds.}
\label{tab:reflectmem-checks}
\small\setlength{\tabcolsep}{3.5pt}
\begin{tabular}{lrrrrrrrrrr}
\toprule
Arm & 1 & 2 & 3 & 4 & 5 & 6 & 7 & 8 & Median & $\le 8.00$ \\
\midrule
ReflectMem + checks & 7.96 & 9.61 & 8.14 & 7.49 & 8.72 & 7.45 & 8.05 & 7.79 & 8.005 & 4/8 \\
\QuantForge & 7.74 & 9.31 & 7.98 & 7.38 & 8.44 & 7.47 & 7.86 & 7.59 & 7.800 & 6/8 \\
\bottomrule
\end{tabular}
\end{table}

\QuantForge is lower in seven of eight seeds; the paired difference has mean
$-0.180$ and 95\% $t$ interval $[-0.265,-0.095]$. The checks work as intended:
across the eight runs, 225 of the 247 probes run confirm the probed change (0.911), close to the rate under
\QuantForge's verdict constraint in the same-evidence replay (Table~\ref{tab:verdict-replay}). The arm
implements what it proposes, but nothing makes it abandon a direction that
discriminating evidence has already ruled out. All arms share the same
contract checks and rollback, so invalid-candidate filtering is common to
every arm.

Table~\ref{tab:two-by-two} places this arm in a $2\times2$ design with
ReflectMem, the arm without compilation (Appendix~\ref{app:errata-ablations}),
and \QuantForge. Either ingredient alone changes the median by less than 0.1;
together they lower it by 0.29. Per seed, the interaction contrast
$\text{QF}-\text{(ReflectMem+checks)}-\text{(without compilation)}+\text{ReflectMem}$
has mean $-0.205$ and 95\% $t$ interval $[-0.286,-0.124]$, and it is negative
in all eight seeds. All four arms in Table~\ref{tab:two-by-two} share the same
single-lineage search, parent update rule, strict-MXFP4 checks, and rollback
policy, isolating competing explanations and local controls (rows) from forced,
runtime-checked code responses (columns).

\begin{table}[htbp]
\centering\revcolor
\caption{Median held-out PPL by competing explanations and forced, checked
responses. All four arms share the same single-lineage search and format-contract checks.}
\label{tab:two-by-two}
\small\setlength{\tabcolsep}{8pt}
\begin{tabular}{lcc}
\toprule
& No forced response & Forced response \\
\midrule
No competing explanations & ReflectMem: 8.09 & ReflectMem + checks: 8.005 \\
Competing explanations and controls & Without compilation: 8.06 & \QuantForge: \textbf{7.80} \\
\bottomrule
\end{tabular}
\end{table}

\subsection{Proposer Models}
\label{app:second-proposer}

We repeat the comparison between ReflectMem and \QuantForge with two further
proposers from two other vendors, Claude Opus 5.5 and DeepSeek V4.1-Flash, on
search seeds 1--4. Within each proposer, both arms share the coding agent, the interface
configuration, generation settings, initial RTN program, frozen data, and the
240-call budget; the input limit stays at 65,536 tokens with at most 16,384 for
memory. Each proposer keeps its own internal reasoning budget.

\begin{table}[htbp]
\centering\revcolor
\caption{\QuantForge against ReflectMem across proposers: median held-out
Llama-3.1-8B WikiText-2 PPL after 240 calls, and the paired difference
(\QuantForge minus ReflectMem) with its 95\% $t$ interval.}
\label{tab:proposer-models}
\small\setlength{\tabcolsep}{5pt}
\begin{tabular}{lrrrr}
\toprule
Proposer & ReflectMem & \QuantForge & Paired difference & \QuantForge lower \\
\midrule
GPT-5.6 Sol (8 seeds) & 8.09 & 7.80 & $-0.25$ $[-0.39,-0.11]$ & 7/8 \\
Claude Opus 5.5 (4 seeds) & 8.16 & 7.76 & $-0.32$ $[-0.48,-0.15]$ & 4/4 \\
DeepSeek V4.1-Flash (4 seeds) & 8.70 & 8.43 & $-0.19$ $[-0.35,-0.03]$ & 4/4 \\
\bottomrule
\end{tabular}
\end{table}

\begin{table}[htbp]
\centering\revcolor
\caption{Per-seed held-out PPL for the two additional proposers.}
\label{tab:proposer-seeds}
\small\setlength{\tabcolsep}{5pt}
\begin{tabular}{llrrrr}
\toprule
Proposer & Arm & Seed 1 & Seed 2 & Seed 3 & Seed 4 \\
\midrule
Claude Opus 5.5 & ReflectMem & 8.14 & 9.76 & 8.17 & 7.54 \\
 & \QuantForge & 7.69 & 9.55 & 7.83 & 7.27 \\
DeepSeek V4.1-Flash & ReflectMem & 8.62 & 10.41 & 8.78 & 8.03 \\
 & \QuantForge & 8.39 & 10.33 & 8.47 & 7.89 \\
\bottomrule
\end{tabular}
\end{table}

\QuantForge is lower than ReflectMem for every proposer, with paired gaps from
$0.19$ to $0.32$. The proposer still sets the level that both arms reach.
Deriving the candidate formulas involves residual expansions, error cross-terms,
curvature approximations, and stability bounds. Table~\ref{tab:proposer-mechanisms}
counts the \QuantForge runs that implement each mechanism of
Table~\ref{tab:first-appearance}. The first three mechanisms appear in most
runs under every proposer; the last stage, remeasuring the MLP after attention,
varies with the proposer, with three of four runs for Opus~5.5, four of eight
for GPT-5.6 Sol, and one of four for DeepSeek V4.1-Flash. Opus~5.5 also reaches
the lowest median of the three proposers. DeepSeek V4.1-Flash implements the
earlier mechanisms about as often as the others but realizes them less well and spends
more of its budget on controls and execution checks. ReflectMem reads the same
diagnostics (Appendix~\ref{app:framework-details}) and reaches each mechanism in
fewer runs (Table~\ref{tab:first-appearance}). The comparison spans candidate specification and
control choice as well as mathematical reasoning.

\begin{table}[htbp]
\centering\revcolor
\caption{\QuantForge runs (out of four) that implement each mechanism, by
proposer.}
\label{tab:proposer-mechanisms}
\small\setlength{\tabcolsep}{5pt}
\begin{tabular}{lrrrr}
\toprule
Proposer & $K$-type corr. & Refit after $DT$ & Mismatch rule & MLP after attn. \\
\midrule
Claude Opus 5.5 & 4/4 & 4/4 & 4/4 & 3/4 \\
DeepSeek V4.1-Flash & 4/4 & 4/4 & 3/4 & 1/4 \\
\bottomrule
\end{tabular}
\end{table}

\subsection{Estimated A100 GPU-Hour Cost of Discovery}
\label{app:gpu-cost}

In the 240-call comparison of Appendix~\ref{app:framework-details}, baselines spend all 240 calls evaluating full Qwen3-4B candidates, whereas \QuantForge spends 99 of its 240 calls on cheap local twin controls (69) and runtime checks (30). To compare actual compute cost alongside call count, we convert all 7,680 recorded search calls (4 arms $\times$ 8 seeds $\times$ 240 calls) and the 12 held-out Llama-3.1-8B checks per run into estimated single-A100-80GB GPU-hours. Because the search logs recorded call types and active quantizer stages rather than per-call wall-clock timestamps, we estimate each call's runtime from a per-stage timing model anchored to single-A100 measurements of 4B baselines (WUSH, GPTQ, MR-GPTQ, WAET/PVSD), paired fixture packages, runtime checks, and 8B builds (excluding LLM API latency and shared startup). Across all 7,680 calls, full 4B candidates take a median of $30.2$\,min (range $5.1$--$54.7$\,min depending on active stages), early-stopped rejections $0.8$\,min, local twin controls $2.0$\,min, and runtime checks $4.3$\,s.

\begin{figure}[htbp]
  \centering
  \includegraphics[width=0.98\textwidth]{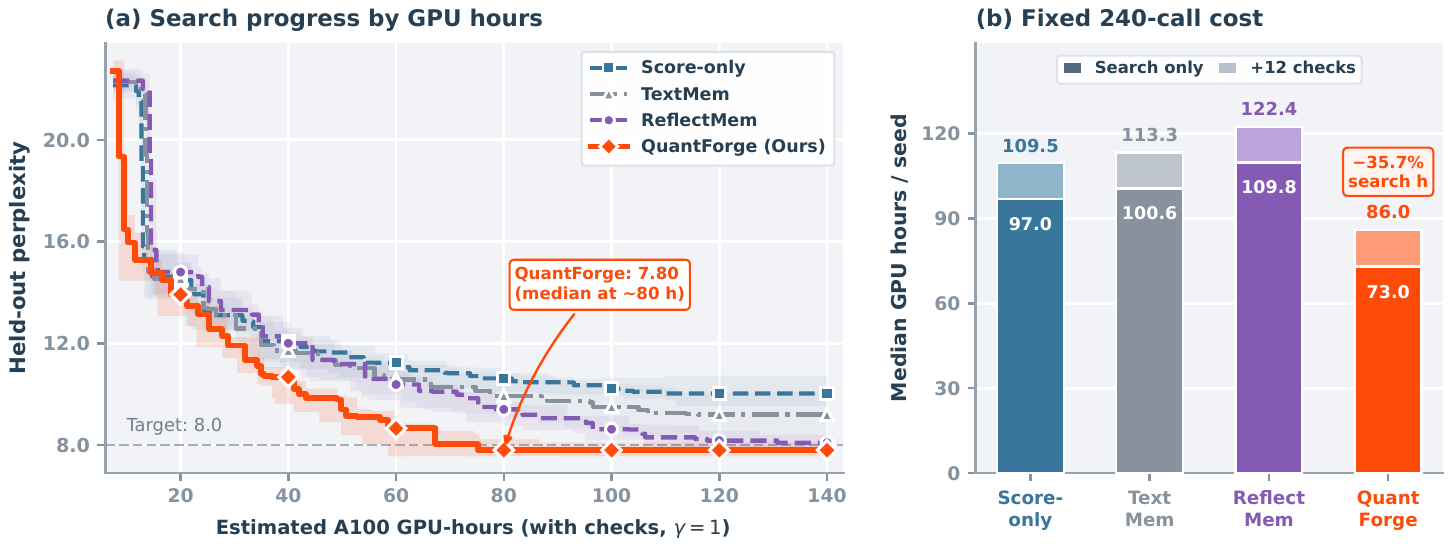}
  \caption{\textbf{Estimated A100 GPU-hour discovery cost across eight paired RTN-seeded runs ($\gamma=1$).} \textbf{(a)} Eight-seed median held-out Llama-3.1-8B WT2 PPL (shaded interquartile range) versus cumulative A100 GPU-hours including held-out checks; curves hold their final score once the 240-call budget finishes. \textbf{(b)} Fixed 240-call median GPU-hours per seed, split into search (solid) and $+12$ external held-out checks (pastel cap). \QuantForge cuts search GPU-hours by a paired median of $35.7\%$ relative to ReflectMem while reaching $7.80$ vs.\ $8.09$ PPL.}
  \label{fig:gpu-cost-progress}
\end{figure}

Figure~\ref{fig:gpu-cost-progress} and Table~\ref{tab:gpu-cost-summary} show that \QuantForge uses a median of $73.0$ search GPU-hours per seed ($86.0$\,h including the 12 held-out checks), compared with $109.8$\,h ($122.4$\,h) for ReflectMem---a paired median ratio of $0.64$ ($35.7\%$ lower search compute; $0.69$ with external checks). Among the three seeds where both \QuantForge and ReflectMem reach $\mathrm{PPL} \le 8.00$ (seeds 4, 6, and 8), \QuantForge reaches the target $1.91\times$ faster in median search GPU-hours ($38.2 / 45.3 / 49.7$\,h vs.\ $97.0 / 81.2 / 95.1$\,h; $1.82\times$ with external checks).

\begin{table}[htbp]
\centering
\caption{Estimated A100 GPU-hours for the 240-call runs (median across 8 paired seeds) and median time to reach $\mathrm{PPL}\le 8.00$ among passing runs.}
\label{tab:gpu-cost-summary}
\small\setlength{\tabcolsep}{5pt}
\begin{tabular}{lrrrrrr}
\toprule
Arm & Final PPL$\downarrow$ & Pass & Search (h)$\downarrow$ & +12 Checks (h)$\downarrow$ & Pass Search (h) & Pass +Checks (h) \\
\midrule
Score-only & 10.03 & 1/8 & 97.0 & 109.5 & 107.2 (1/8) & 119.6 (1/8) \\
TextMem & 9.20 & 3/8 & 100.6 & 113.3 & 99.4 (3/8) & 110.8 (3/8) \\
ReflectMem & 8.09 & 3/8 & 109.8 & 122.4 & 95.1 (3/8) & 106.0 (3/8) \\
\textbf{\QuantForge} & \textbf{7.80} & \textbf{6/8} & \textbf{73.0} & \textbf{86.0} & \textbf{53.2 (6/8)} & \textbf{62.7 (6/8)} \\
\bottomrule
\end{tabular}
\end{table}

\begin{table}[htbp]
\centering
\caption{Per-seed estimated A100 search GPU-hours (240 calls) and break-even candidate slowdown $\gamma$ at which \QuantForge matches ReflectMem's compute cost ($\gamma=1.0$ already accounts for stage-level complexity).}
\label{tab:gpu-cost-seeds}
\scriptsize\setlength{\tabcolsep}{4pt}
\begin{tabular}{rrrrrrrr}
\toprule
Seed & Score-only (h) & TextMem (h) & ReflectMem (h) & \QuantForge (h) & QF / Refl.\ & \shortstack[r]{Break-even $\gamma$\\(search)} & \shortstack[r]{Break-even $\gamma$\\(+checks)} \\
\midrule
1 & 99.4 & 102.6 & 117.7 & 77.4 & 0.66 & 1.54 & 1.52 \\
2 & 67.9 & 69.9 & 94.2 & 60.6 & 0.64 & 1.58 & 1.57 \\
3 & 96.8 & 99.9 & 114.0 & 72.6 & 0.64 & 1.59 & 1.58 \\
4 & 107.2 & 109.0 & 120.1 & 76.9 & 0.64 & 1.58 & 1.57 \\
5 & 73.4 & 96.6 & 98.7 & 61.0 & 0.62 & 1.65 & 1.63 \\
6 & 97.8 & 110.9 & 119.8 & 76.9 & 0.64 & 1.58 & 1.56 \\
7 & 97.2 & 99.4 & 102.7 & 71.9 & 0.70 & 1.44 & 1.45 \\
8 & 91.8 & 101.4 & 105.6 & 73.3 & 0.69 & 1.46 & 1.45 \\
\midrule
Median & 97.0 & 100.6 & 109.8 & \textbf{73.0} & \textbf{0.64} & 1.58 & 1.57 \\
\bottomrule
\end{tabular}
\end{table}

\paragraph{Sensitivity to timing assumptions.}
Table~\ref{tab:gpu-cost-sensitivity} checks how robust the \QuantForge/ReflectMem cost ratio is to four timing assumptions: (i) charging early-stopped rejections anywhere from $0\times$ to $1\times$ a full candidate keeps the search ratio within $0.63$--$0.64$; (ii) adding an extra overhead multiplier $\gamma \in [1.0, 2.0]$ to \QuantForge's full candidates shows \QuantForge stays faster up to $\gamma = 1.44$--$1.65$ across all eight seeds (Table~\ref{tab:gpu-cost-seeds}); (iii) scaling local-control cost by $0.5\times$--$1.5\times$ moves the search ratio only from $0.63$ to $0.66$; and (iv) scaling 8B external checks by $0.8\times$--$1.5\times$ keeps the total (+checks) ratio within $0.68$--$0.70$.

\begin{table}[htbp]
\centering
\caption{Sensitivity of the \QuantForge/ReflectMem GPU-hour ratio (8-seed median) to four timing assumptions.}
\label{tab:gpu-cost-sensitivity}
\scriptsize\setlength{\tabcolsep}{4.5pt}
\begin{tabular}{lrr@{\hspace{14pt}}lrr}
\toprule
Assumption Varied & Search Ratio & +Checks & Assumption Varied & Search Ratio & +Checks \\
\midrule
Early-stop $= 0.0\times$ full & 0.64 & 0.69 & Extra QF candidate cost $\gamma=1.00$ & 0.64 & 0.69 \\
Early-stop $=$ nominal & 0.64 & 0.69 & Extra QF candidate cost $\gamma=1.25$ & 0.80 & 0.82 \\
Early-stop $= 0.5\times$ full & 0.63 & 0.67 & Extra QF candidate cost $\gamma=1.50$ & 0.95 & 0.96 \\
Early-stop $= 1.0\times$ full & 0.63 & 0.67 & Extra QF candidate cost $\gamma=1.75$ & 1.11 & 1.10 \\
\cmidrule(r){1-3}
Local control $\times 0.50$ / $1.50$ & 0.63 / 0.66 & 0.68 / 0.70 & Extra QF candidate cost $\gamma=2.00$ & 1.26 & 1.24 \\
8B check scale $0.80\times$ / $1.50\times$ & 0.64 / 0.64 & 0.68 / 0.70 & & & \\
\bottomrule
\end{tabular}
\end{table}

\subsection{Black-Box PPL-Control Ablation}
\label{app:blackbox-controls}

To separate the benefit of \emph{paired BF16--MXFP4 local diagnostics} from the control-and-check workflow itself, we run a 4-seed ablation (seeds 1--4, matching Appendix~\ref{app:autonomous-agent}) that keeps the identical proposer, coding agent, RTN seed, hypothesis registration, enforced updates, runtime checks, and 240-call split (141 candidates, 69 controls, 30 checks), but restricts every control to return only scalar Qwen3-4B WikiText-2 PPL instead of paired layerwise error signatures. Runtime checks remain active ($26$--$28/30$ passed per seed), and held-out Llama-3.1-8B evaluation stays strictly isolated.

Table~\ref{tab:blackbox-controls} shows that black-box PPL controls fall midway between the autonomous agent ($8.55$ median, $1/4$ passing) and full paired-twin \QuantForge ($7.86$ median, $3/4$ passing). Full paired-twin \QuantForge beats black-box controls on all $4/4$ seeds (mean gain $0.45$ PPL, per-seed gains $0.32$--$0.62$) while using $31\%$ fewer estimated search GPU-hours ($74.8$ vs.\ $108.9$\,h median across seeds 1--4; Black-box/Full ratio $1.46\times$, or $1.39\times$ with external checks). Even under an idealized zero-build lower bound where all 69 black-box controls reuse weights and pay only setup plus evaluation, black-box search still requires $75.1$\,h ($1.00\times$ full \QuantForge; $1.23\times$ at $50\%$ reuse), and scaling the full-call workload estimate by $0.75\times$--$1.50\times$ keeps the Black-box/Full ratio within $1.10\times$--$2.18\times$. Table~\ref{tab:blackbox-traj} lists the 20-call held-out PPL trajectory for each black-box seed.

\begin{table}[htbp]
\centering
\caption{Isolating paired-twin local diagnostics from black-box PPL controls on held-out Llama-3.1-8B WT2 PPL (seeds 1--4, 240 calls). Full paired-twin \QuantForge outperforms black-box controls on all 4/4 seeds while lowering search compute.}
\label{tab:blackbox-controls}
\scriptsize\setlength{\tabcolsep}{4.5pt}
\begin{tabular}{rrrrrrr}
\toprule
Seed & Autonomous Agent$\downarrow$ & Black-Box Controls$\downarrow$ & Full Paired-Twin$\downarrow$ & Black-Box Search (h) & Black-Box +Checks (h) & Check Pass \\
\midrule
1 & 8.42 & 8.06 & \textbf{7.74} & 115.0 & 129.3 & 27/30 \\
2 & 10.31 & 9.93 & \textbf{9.31} & 86.9 & 99.6 & 26/30 \\
3 & 8.67 & 8.42 & \textbf{7.98} & 107.7 & 121.6 & 27/30 \\
4 & 7.76 & 7.80 & \textbf{7.38} & 110.1 & 124.0 & 28/30 \\
\midrule
Median & 8.55 & 8.24 & \textbf{7.86} & 108.9 & 122.8 & -- \\
$\mathrm{PPL}\le 8.00$ & 1/4 & 1/4 & \textbf{3/4} & -- & -- & -- \\
\bottomrule
\end{tabular}
\end{table}

\begin{table}[htbp]
\centering
\caption{Held-out Llama-3.1-8B WT2 PPL every 20 search calls for the black-box PPL-control ablation (seeds 1--4).}
\label{tab:blackbox-traj}
\scriptsize\setlength{\tabcolsep}{3.8pt}
\begin{tabular}{lrrrrrrrrrrrr}
\toprule
Call & 20 & 40 & 60 & 80 & 100 & 120 & 140 & 160 & 180 & 200 & 220 & 240 \\
\midrule
Seed 1 & 22.16 & 20.10 & 19.12 & 18.39 & 16.06 & 14.50 & 11.62 & 10.37 & 10.17 & 9.81 & 8.71 & 8.06 \\
Seed 2 & 23.62 & 20.92 & 19.83 & 16.88 & 16.16 & 15.26 & 12.86 & 11.64 & 11.20 & 10.75 & 10.14 & 9.93 \\
Seed 3 & 24.06 & 20.70 & 20.21 & 18.15 & 17.54 & 15.85 & 14.75 & 13.31 & 12.87 & 10.02 & 9.47 & 8.42 \\
Seed 4 & 23.10 & 20.62 & 20.98 & 18.99 & 18.40 & 15.87 & 12.88 & 11.20 & 10.34 & 9.09 & 8.56 & \textbf{7.80} \\
\bottomrule
\end{tabular}
\end{table}

}

\end{document}